\documentclass{article}
\usepackage{iclr2027_conference,times}

\usepackage{amsmath}
\usepackage{amssymb}
\usepackage{graphicx}
\usepackage{xcolor}
\usepackage{booktabs}
\usepackage{multirow}
\usepackage{hyperref}

\definecolor{linkInk}{HTML}{35618F}
\hypersetup{colorlinks=true, linkcolor=linkInk, citecolor=linkInk,
            urlcolor=linkInk, filecolor=linkInk}
\usepackage{url}
\usepackage{tikz}
\usetikzlibrary{positioning,arrows.meta,fit,backgrounds,calc}
\usepackage{listings}
\usepackage[most]{tcolorbox}
\definecolor{caseOK}{HTML}{3F7D62}     
\definecolor{caseBad}{HTML}{9E2B25}    
\definecolor{caseNull}{HTML}{6E6E6E}   
\definecolor{caseProb}{HTML}{35618F}   
\definecolor{caseHi}{HTML}{F2D680}     

\newcommand{\hlk}[1]{%
  \begingroup\setlength{\fboxsep}{0.9pt}%
  \colorbox{caseHi}{\strut\scriptsize\ttfamily #1}\endgroup}

\lstdefinestyle{transcript}{%
  basicstyle=\scriptsize\ttfamily,
  breaklines=true,
  breakatwhitespace=false,
  columns=fullflexible,
  keepspaces=true,
  showstringspaces=false,
  upquote=true,
  extendedchars=false,
  postbreak=\mbox{\textcolor{caseNull}{$\hookrightarrow$}\space},
  aboveskip=0pt,
  belowskip=0pt,
  escapeinside={(@}{@)},
}

\tcbset{
  casebase/.style={
    listing only,
    breakable,
    enhanced,
    listing style=transcript,
    boxrule=0.5pt,
    arc=1.6pt,
    left=4pt, right=4pt, top=3.5pt, bottom=3.5pt,
    colback=black!2,
    fonttitle=\scriptsize\sffamily,
    coltitle=white,
    toptitle=1pt, bottomtitle=1pt,
    before skip=6pt, after skip=6pt,
  },
}

\newtcblisting{outputok}[1]{casebase, colframe=caseOK,   colbacktitle=caseOK,   title={#1}}
\newtcblisting{outputbad}[1]{casebase, colframe=caseBad,  colbacktitle=caseBad,  title={#1}}
\newtcblisting{outputnull}[1]{casebase, colframe=caseNull, colbacktitle=caseNull, title={#1}}

\newtcolorbox{problemcard}[2]{%
  enhanced, breakable,
  colback=caseProb!4, colframe=caseProb, boxrule=0.5pt, arc=1.6pt,
  left=5pt, right=5pt, top=3.5pt, bottom=3.5pt,
  fonttitle=\scriptsize\sffamily, coltitle=white, colbacktitle=caseProb,
  toptitle=1pt, bottomtitle=1pt,
  before skip=6pt, after skip=6pt,
  title={\textsc{problem}\ \ \texttt{#1}\hfill reference answer\ \ \texttt{#2}},
  fontupper=\small,
}

\usepackage{amsmath,amsfonts,bm}

\def\eqref#1{equation~\ref{#1}}

\def\1{\bm{1}}

\DeclareMathAlphabet{\mathsfit}{\encodingdefault}{\sfdefault}{m}{sl}
\SetMathAlphabet{\mathsfit}{bold}{\encodingdefault}{\sfdefault}{bx}{n}

\newcommand{\clr}{\textsc{clr}}
\newcommand{\er}{\mathrm{ER}}

\definecolor{dUp}{HTML}{3F7D62}
\definecolor{dDown}{HTML}{9E2B25}
\newcommand{\dgain}[1]{\textcolor{dUp}{$\uparrow$\,#1}}
\newcommand{\dloss}[1]{\textcolor{dDown}{$\downarrow$\,#1}}

\title{Phantom Gains: Auditing Self-Improvement \\ Against a Measured Null}

\author{
  Cheng Xu$^1$ \quad Nan Yan$^2$ \quad Liming Chen$^3$ \quad M-Tahar Kechadi$^1$ \\
  $^1$University College Dublin 
  $^2$Georgia Institute of Technology 
  $^3$Dalian University of Technology \\
  \texttt{cheng.xu1@ucdconnect.ie}, \texttt{nan.yan@gatech.edu}, \texttt{tahar.kechadi@ucd.ie}
}

\iclrfinalcopy

\makeatletter
\g@addto@macro\@maketitle{\lhead{Preprint. Under review.}}
\makeatother

\begin{document}

\maketitle

\begin{abstract}
Whether a language model has improved itself is increasingly judged not by mean accuracy but by which individual problems it gains and loses. Tracking these transitions means differencing two noisy estimates, leaving them vulnerable to measurement artifacts. Auditing three rounds of rank-$32$ LoRA self-training on Qwen3-8B against a frozen control pushed through the identical pipeline, we identify seven measurement failures, each of which inverts a reported finding when its control is absent. Several are standard practice. A ledger built on a single greedy decode manufactures capability changes on an untrained model, largely an artifact of inference batching; the expansion statistic separating acquisition from sharpening assigns that same model a rate of $0.280$. The natural threshold repair does not survive replication: estimated across the frozen comparisons such a design already contains, its null stays non-zero. We replace it with a per-problem exact test against a pooled baseline under false-discovery-rate control, which detects nothing on any held-out replicate and is unchanged under the multiple-testing rule, error rate and pool size. Applied to a ladder of arms matched in stream, volume and evaluation, the audit finds that external distillation improves problems the base model rarely reaches while three forms of self-training do not; a regression rejects this asymmetry as a by-product of distillation's larger overall gain ($p < 10^{-8}$). On the far smaller set of problems the base model never reaches, the evidence is inconclusive, while self-training corrupts problems solved at baseline at rates well above the measured floor. Transition-level auditing therefore requires a separately measured null for every statistic it reports: nulls that cost no new experiments, built from baseline replicates a multi-arm study already owns, though not from as few as most possess. Code and evaluation artifacts are available at \url{https://github.com/chengxuphd/phantom-gains}.
\end{abstract}

\section{Introduction}
\label{sec:intro}

A language model that improves itself without new supervision would be a considerable thing, and a growing family of methods reports exactly that \citep{fang2025comprehensive,tao2024survey}. STaR fine-tunes on its own correct rationales \citep{zelikman2022star}; TTRL replaces labels with a majority vote over its own samples \citep{zuo2025ttrl}; more elaborate schemes generate their own curricula \citep{zhao2025absolutezero,chen2025selfquestioning,huang2026rzero} or learn to edit their own weights \citep{zweiger2025self}.

The evidence offered used to be a single number: accuracy before, accuracy after \citep{atreja2025alas}. That is now widely accepted to be insufficient, since an aggregate cannot distinguish a problem the model newly learned to solve from one it could already solve on a good day \citep{yu2026skilladaptor,yue2025does}, nor show what was lost while the mean rose. A second generation of analyses has therefore moved to individual problems, tracking which enter and leave a model's reachable set \citep{yuan2026diversity,wu2026invisibleleash,mayilvahanan2026math} and counting those destroyed under self-training \citep{lin2026detecting,yu2026self}.

This is the right move, and it creates a new problem. A transition is a difference between two estimates, each carrying sampling error, and the events of interest sit exactly where that error bites hardest. Auditing self-training on Qwen3-8B across three evaluation sets, we isolate \emph{seven} measurement failures, each of which produces a confident and wrong conclusion wherever its control is absent (Figure~\ref{fig:design}). Two are standard practice, and one lies inside the correction for another.

\textbf{A single decode is not a state.} Temperature-zero decoding is not deterministic under batching \citep{he2025nondeterminism}, so a ledger built on one greedy sample per problem gives an untrained model, re-evaluated against itself, $6$ apparent learnings and $9$ apparent corruptions. That is a corruption-to-learning ratio of $1.5$: the shape of the alarming result this literature reports, manufactured out of nothing. Serializing the requests removes three-quarters of those flips and confirms the mechanism, but not all of them: even one request at a time, a frozen model changes $2\%$ of its greedy verdicts.

\textbf{An expansion statistic needs its own null, and one measurement of a null is not a null.} Following \citet{mayilvahanan2026math}, a problem counts as \emph{expanded} when the base model produced no correct sample in $k$ draws and the trained model produces at least one. That rule compares two noisy binomial estimates through asymmetric thresholds, and what it returns when nothing has changed has not been measured. It returns $0.280$: a frozen Qwen3-8B evaluated twice at $k{=}128$ ``expands'' $7$ of the $25$ AIME problems it had not reached. All seven are single lucky samples, which appears to license requiring $m \geq 2$ successes and so to drive the null to zero. It does not: that conclusion rests on one frozen comparison, and over the $110$ our experiments contain the null is $0.058\,[0.038, 0.078]$, larger than the value it was supposed to certify. The repair is to stop thresholding: a per-problem exact test against a baseline pooled over every independent evaluation of the untrained model gives all arms a common denominator, returns no detection on any held-out replicate, and holds under every multiple-testing rule we try.

Applying the controlled audit, with distillation from a stronger external teacher as a positive control matched in stream, retained volume and evaluation, we find a dissociation. Distillation improves $8$--$11$ of the $22$ problems the base model reaches at most five times in $1{,}408$ draws, against $0$--$2$ for three forms of self-training, and a logistic model rejects the reading that this merely follows from the teacher's larger overall gain ($\beta = 1.91$, $p < 10^{-8}$). We are careful about what this licenses: on the ten problems the base model never reaches at all the comparison is not significant, so the claim is about rarely-reached problems and not about expansion. It aligns with the sharpening account of \citet{huang2025self,yue2025does} and with \citet{strozzi2026teacherfree}, on a different method family and with a positive control neither runs. We also quantify what is destroyed: both methods corrupt $88$--$106$ of $1{,}163$ band problems against a design-matched floor of $8$, over half of those events larger than any solve-rate change a frozen model produced across twenty independent comparisons.

\paragraph{Contributions.} (i) Seven measurement failure modes for transition-level audits, each demonstrated on a run where it reverses a conclusion; the last arises inside the correction for the second (Section~\ref{sec:failures}). (ii) A threshold-free replacement for the expansion statistic, whose null we measure rather than assume and whose comparisons survive the multiple-testing rule, the error rate and the pool size (Section~\ref{sec:f2}). (iii) The observation that makes measured nulls nearly free, namely that every arm's checkpoint $0$ is an independent evaluation of the untrained model, together with the replicate count at which one becomes trustworthy, which is more than a three-arm study owns. (iv) A transition ledger with a benchmark-specific noise floor and effect-size accounting that separates threshold flicker from real capability loss, and a policy-gradient arm on which three seeds disagree about when the model collapses but not that it does (Section~\ref{sec:results}).


\definecolor{fAime}{HTML}{35618F}
\definecolor{fBand}{HTML}{7A6E9B}
\definecolor{fMath}{HTML}{A9552F}
\definecolor{fTeach}{HTML}{3F7D62}
\definecolor{fRule}{HTML}{9A9A9A}
\definecolor{fInk}{HTML}{2B2B2B}
\definecolor{fWarn}{HTML}{9E2B25}

\begin{figure}[t]
\centering
\begin{tikzpicture}[
  x=1cm, y=1cm, font=\scriptsize,
  box/.style   ={draw=fInk, line width=0.5pt, rounded corners=1.2pt, fill=white,
                 inner sep=2.4pt, align=center, minimum height=8.2mm, anchor=center},
  ar/.style    ={-{Latex[length=1.4mm,width=1.1mm]}, fInk, line width=0.45pt},
  arg/.style   ={-{Latex[length=1.4mm,width=1.1mm]}, fRule, line width=0.45pt},
  fm/.style    ={fWarn, font=\scriptsize\bfseries, inner sep=1pt},
  lbl/.style   ={fRule, font=\scriptsize, inner sep=1.5pt}]

\node[box, minimum width=17mm] (stream) at (0.95,3.05) {unlabeled\\stream};
\node[box, minimum width=18mm] (gen)    at (3.20,3.05) {sample $n$\\rationales};
\node[box, minimum width=30mm, draw=fTeach] (filt) at (6.10,3.05)
      {\textbf{filter}: answer $\mid$ vote $\mid$ teacher};
\node[box, minimum width=14mm] (train)  at (8.85,3.05) {LoRA\\step};
\node[box, minimum width=21mm, draw=fMath, fill=fMath!7] (ck) at (11.55,3.05)
      {checkpoints\\$\theta_0\!\to\!\theta_3$};

\draw[ar] (stream) -- (gen);
\draw[ar] (gen) -- (filt);
\draw[ar] (filt) -- (train);
\draw[ar] (train) -- (ck);
\draw[ar] (train.north) -- ++(0,0.42) -| (gen.north);
\node[lbl] at (6.05,3.72) {$\times\,3$ rounds};

\node[box, minimum width=108mm, minimum height=5.2mm, draw=fRule, fill=black!4]
     (frozen) at (5.62,2.10)
     {\textbf{frozen control}\, $\theta_0$ evaluated twice, no training in between
      $\;\Rightarrow\;$ a measured null for \emph{every} statistic reported below};
\draw[arg] (stream.south) -- (stream.south |- frozen.north);

\node[box, minimum width=26mm] (eval) at (1.55,0.62)
     {evaluate $\mathcal{P}$\\$k{=}128\times4$ ckpts};
\node[box, minimum width=30mm] (est)  at (4.90,0.62)
     {$\hat\pi_t(p)=\frac1k\sum\mathbf{1}[\text{correct}]$\\
      band $\frac12\pm z\sqrt{1/4k}$};
\node[box, minimum width=26mm] (ledg) at (8.15,0.62)
     {seven-state ledger\\{\scriptsize learned $\mid$ corrupted $\mid \dots$}};
\node[box, minimum width=26mm, draw=fTeach, fill=fTeach!7] (met) at (11.45,0.62)
     {$\textsc{clr},\ \mathrm{ER}_m$\\{\scriptsize each against its floor}};

\draw[ar] (eval) -- (est);
\draw[ar] (est) -- (ledg);
\draw[ar] (ledg) -- (met);

\draw[ar] (ck.south) -- (11.55,1.50) -- (1.55,1.50) -- (eval.north);
\draw[arg] (5.62,1.845) -- (5.62,1.50);

\node[fm] at (9.55,3.72) {F5};
\node[fm] at (2.35,1.32) {F3};
\node[fm] at (4.90,1.32) {F1};
\node[fm] at (7.75,1.32) {F4};
\node[fm] at (8.60,1.32) {F6};
\node[fm] at (11.45,1.32) {F2};
\node[fm] at (-0.30,2.10) {F7};

\node[lbl, anchor=north, align=left, text width=132mm] at (6.45,0.02) {%
\textcolor{fWarn}{\textbf{F1}} a single decode in place of the estimator: the frozen model
scores $\textsc{clr}{=}1.5$. \;
\textcolor{fWarn}{\textbf{F2}} a one-success expansion threshold: the frozen model
``expands'' $0.280$ of its unreached problems. \;
\textcolor{fWarn}{\textbf{F3}} a fixed token cap under style drift: the most effective arm
is scored the most destructive.\;
\textcolor{fWarn}{\textbf{F4}} transition metrics carry ${\sim}10\times$ less power than
accuracy: a $112$-problem pilot cannot resolve the comparison. \;
\textcolor{fWarn}{\textbf{F5}} training-seed variance: $\textsc{clr}$ spans $0.55$--$1.53$
for one method. \;
\textcolor{fWarn}{\textbf{F6}} an underpowered probe: a phantom $10$-point safety drop. \;
\textcolor{fWarn}{\textbf{F7}} a null measured once: the ``corrected'' expansion threshold
has a null of $0.058$, not $0$.};

\end{tikzpicture}

\caption{\textbf{The audit pipeline, and the seven stages at which a measurement failure
enters it.} \emph{Top:} an evolution arm draws candidates from an unlabeled stream, filters
them by its own criterion (ground-truth answer for STaR, majority vote for the self-training arm, an
external teacher for the distillation control) and takes a LoRA step, for three rounds.
\emph{Middle:} a frozen copy of $\theta_0$ is pushed through the identical evaluation path,
so every statistic is read against a null measured on the same problems with a model that
did not train. \emph{Bottom:} evaluation draws $k{=}128$ samples per problem per checkpoint,
converts them to a solve rate rather than a single decode, discretizes with hysteresis, and
reads off a seven-state ledger. \textcolor{fWarn}{F7} sits on the control path itself: the
null each statistic is read against is an estimate, and measuring it once is the same error
one level up. Every mode is measured here, inverts a reported conclusion when uncontrolled,
and is caught by a control at the stage marked (Section~\ref{sec:failures}).}
\label{fig:design}
\end{figure}

\section{Related Work}
\label{sec:related}

\paragraph{What self-improvement can add.} \citet{huang2025self} model self-improvement as \emph{sharpening}: the model acts as its own verifier and concentrates mass on sequences it already assigns high likelihood, improving first-sample accuracy without introducing anything new; \citet{song2025mind} formalize the same limit through a generation--verification gap. Empirically, \citet{yue2025does} report that RLVR-trained models are outperformed by their base models at large $k$, \citet{wu2026invisibleleash} argue such training cannot place mass outside the base support, and \citet{yuan2026diversity} trace the effect to diversity collapse, while \citet{liu2025prorl} argue the opposite for prolonged training. MATH-Beyond \citep{mayilvahanan2026math} turns the debate into a measurement and supplies the expansion statistic we audit in Section~\ref{sec:f2}; \citet{qi2026generalization} benchmark closed-loop self-evolution against oracle supervision and find a persistent gap.

\paragraph{Concurrent work on transition-level analysis.} \citet{strozzi2026teacherfree} ask our headline question, whether teacher-free self-training acquires capability or only expresses existing capability, and report a pass@$K$ crossover on a DSL domain with a free verifier. We read our result as a replication in a different domain and supervision regime, with a positive control they do not run. \citet{yuan2026diversity} track per-problem boundary entry and exit across training, structurally our ledger, under an \emph{analytic} binomial null. Ours is measured, and we can now say what that buys: for the estimator we recommend the two agree closely, and they part company where the noise is not binomial, which is how we found the expansion statistic's null to be non-zero at any threshold below $m{=}5$ (Appendix~\ref{app:concurrent}).

\paragraph{Everything else.} Appendix~\ref{app:related} places this work against the self-evolution methods it audits, the degradation and model-collapse literature that motivates the corruption half, and the statistical-evaluation work our controls build on.

\section{A transition-level audit, and what it must control for}
\label{sec:audit}

\paragraph{The ledger.} Fix an evaluation set $\mathcal{P}$ and checkpoints $t = 0, \dots, T$, with $t=0$ the untrained base model. For each problem $p$ and checkpoint we draw $k$ samples and record the solve rate $\hat{\pi}_t(p)$. Discretizing into a binary state $s_t(p)$ gives a trajectory per problem, and every problem falls into exactly one of seven categories: \emph{stable-correct}, \emph{stable-incorrect}, \emph{learned}, \emph{corrupted}, \emph{recovered}, \emph{transient}, \emph{oscillating}. These partition $\mathcal{P}$, so counts need no normalization. The headline quantity is the \textbf{corruption-to-learning ratio},
\begin{equation}
\clr = \frac{|\{p : s_0(p) = \textsf{solved},\; s_T(p) = \textsf{unsolved}\}|}
            {|\{p : s_0(p) = \textsf{unsolved},\; s_T(p) = \textsf{solved}\}|},
\label{eq:clr}
\end{equation}
problems destroyed per problem gained; $\clr > 1$ removes more capability than it adds, whatever happens to mean accuracy. Equation~\ref{eq:clr} is undefined when nothing is learned, as on AIME, where we report the counts instead. We define it on endpoints, and report trajectory-based counts alongside (Appendix~\ref{app:counts}); they agree to within three problems.

\paragraph{Correctness as an estimator, not a draw.} The obvious definition of $s_t(p)$ is a single greedy sample, which modern inference does not support (Section~\ref{sec:f1}). We instead treat $\hat\pi_t(p)$ as an estimator of the model's pass@1 probability and discretize with hysteresis:
\begin{equation}
s_t(p) = \begin{cases}
\textsf{solved} & \hat\pi_t(p) \geq \theta_{\mathrm{hi}}, \\
\textsf{unsolved} & \hat\pi_t(p) < \theta_{\mathrm{lo}}, \\
s_{t-1}(p) & \text{otherwise,}
\end{cases}
\qquad \theta_{\mathrm{hi/lo}} = \tfrac{1}{2} \pm z\sqrt{\tfrac{1}{4k}},
\label{eq:band}
\end{equation}
with $z = 2$, $k = 128$ throughout, giving a band of $[0.41, 0.59]$. The half-width is the worst-case standard error of a binomial proportion at $k$ samples, so a problem must move further than ordinary jitter before it is recorded as changing state. The band degenerates at $k \leq z^2$, where it covers all of $[0,1]$ and no problem can change state.

\textbf{The baseline state carries no hysteresis.} Equation~\ref{eq:band} is undefined at $t=0$, where there is no previous state to hold; we assign $s_0(p) = \textsf{solved}$ iff $\hat\pi_0(p) \geq \tfrac12$. A problem whose baseline lies inside the band is therefore unprotected at that one checkpoint, which is why the floors of Table~\ref{tab:floors} are not zero on a set whose mean solve rate is $0.552$ (Appendix~\ref{app:s0}).

\paragraph{Expansion, with an explicit evidence threshold.} Whether a newly solved problem represents new capability is a separate question that pass@1 cannot answer. Following \citet{mayilvahanan2026math}, a problem that becomes pass@1-correct while the base model already solved it at least once in $k$ samples has been \emph{sharpened}; one the base model never reached has been \emph{expanded}. We make the threshold $m$ explicit:
\begin{equation}
\er_m = \frac{|\{p : \hat\pi_0(p) = 0 \;\wedge\; k\hat\pi_T(p) \geq m\}|}
             {|\{p : \hat\pi_0(p) = 0\}|}.
\label{eq:er}
\end{equation}
Prior work uses $m = 1$ implicitly; Section~\ref{sec:f2} shows that no threshold repairs this statistic and replaces it. Note what $\hat\pi_0(p) = 0$ does not mean: no successes in $128$ draws bounds the base pass@1 probability at $q \leq 0.023$ with $95\%$ confidence, it does not establish $q = 0$. We write ``base-unreached at $k$'' rather than ``unsolvable'' throughout, and every claim below is about what a stated sampling budget reaches, not about the support of the base distribution.

\paragraph{A control for every statistic.} Transitions are differences between noisy measurements, so some appear when nothing has changed. We measure how many by re-evaluating a \emph{frozen} model at every checkpoint and computing every statistic exactly as for a trained one; each transition it reports is an artifact by construction. No claim is admitted unless its counts exceed this floor. That floor is measured separately per statistic and per benchmark, and, as Section~\ref{sec:f7} establishes, with an interval and a matched number of checkpoints.

\paragraph{Benchmarks.} Corruption is observable only on problems the model already solves, expansion only on problems it does not reach; the two populations are disjoint, so we use three sets (Figure~\ref{fig:pops}). \textbf{MATH-500} \citep{lightman2024verify}, subsampled to $200$ problems by a fixed stratified draw over (subject, level), is thoroughly contaminated and hosts corruption rather than expansion. \textbf{AIME 2025 and 2026} ($60$ problems) are the least contaminated sets available to us, and AIME 2026 post-dates the backbone's March 2025 cutoff \citep{yang2025qwen3} outright, so we report that half separately wherever a claim rests on it. The base model reaches $0.175$--$0.180$ of AIME, leaving $22$ problems it reaches at most five times in $1{,}408$ pooled draws (only $10$ of them never) and $8$ that can be corrupted. A \textbf{difficulty band} of $1{,}163$ problems, profiled from $10{,}999$ held-out MATH problems \citep{hendrycks2021measuring} at $k{=}8$, is enriched near the decision boundary so that corruption is observable at all; its counts are therefore not a forgetting rate for a natural distribution, and it contains no base-unreached problems by construction (Appendix~\ref{app:band}).

We audit \textbf{STaR} \citep{zelikman2022star} (filter by ground-truth answer), \textbf{majority-vote self-training} in the manner of TTRL \citep{zuo2025ttrl} (filter by majority vote over $8$ samples, pooled by mathematical equivalence so $1/2$ and $0.5$ do not split a vote), and \textbf{distillation} from \texttt{gpt-oss-120b} \citep{openai2025gptoss} as a positive control filtered by exactly STaR's criterion. All arms share the backbone (Qwen3-8B, \citealp{yang2025qwen3}), rank-32 LoRA, three rounds of $256$ problems from one stream, and evaluation at $k{=}128$, $T{=}0.8$, top-$p$ $0.95$ plus one greedy sample per problem per checkpoint. The stream is verified disjoint from every evaluation set by identifier and by normalized text hash, and answers are compared by symbolic equivalence \citep{mathverify2025}. Appendices~\ref{app:protocol} and~\ref{app:grading} give the configuration and the grading rules in full.

\textbf{Two deviations from TTRL as published.} TTRL is reinforcement learning with a majority-vote reward, whereas the arm labeled \emph{majority-vote SFT} below is cross-entropy fine-tuning on the rationales that agree with the vote. The two differ in what happens to the samples that lose the vote, pushed down under the first and discarded under the second. We therefore also run a genuine policy-gradient arm (Section~\ref{sec:results}) and confine claims about a second method family to it. TTRL is also normally applied to the unlabeled target test set, which trains on the problems we score. That on-set configuration is reported throughout as an unmatched baseline, and every claim rests on the matched ladder of Section~\ref{sec:results}, whose training stream is disjoint from the evaluation set.

\section{Seven ways to mismeasure self-improvement}
\label{sec:failures}

Each of the following is demonstrated on a run from this study and would, uncorrected, support a confident and incorrect conclusion; Table~\ref{tab:six} summarizes them. The seventh arises \emph{inside the correction for the second}, which is why we state it as a failure mode in its own right rather than folding it into the discussion.

\begin{table}[t]
\centering
\caption{\textbf{Seven measurement failures, the conclusion each would have supported, and the control that caught it.} Every row is measured on a run from this study rather than posited, and F7 is the failure that arises inside the correction for F2. Cost is the marginal spend on the control; three of the seven cost nothing at all, being re-analyses of records the study already holds, and two more would cost nothing in any study that reports more than one arm.}
\label{tab:six}
\footnotesize
\setlength{\tabcolsep}{3.2pt}
\begin{tabular}{llll r}
\toprule
 & \textbf{failure} & \textbf{would have concluded} & \textbf{control that catches it} & \textbf{cost} \\
\midrule
\textbf{F1} & single-decode ledger & ``a frozen model has $\clr = 1.5$'' & frozen floor $+$ estimator & \$27 \\
\textbf{F2} & one-success $\er$ threshold & ``maj.-vote expands $0.143$ of AIME'' & frozen floor for $\er$ & \$0 \\
\textbf{F3} & fixed token cap & ``distillation is the most destructive'' & log truncation per ckpt & \$0 \\
\textbf{F4} & underpowered ledger & ``the two methods do not differ'' & prior power analysis & \$88 \\
\textbf{F5} & single training seed & ``maj.-vote corrupts less than STaR'' & $\geq 3$ seeds & \$282 \\
\textbf{F6} & underpowered probe & ``self-training erodes refusal by $10$ pts'' & size the probe first & \$2 \\
\textbf{F7} & a null measured once & ``$\er_2$ has a null of zero'' & every baseline is a replicate & \$0 \\
\bottomrule
\end{tabular}
\end{table}

\subsection{F1. Without a floor, a frozen model appears to corrupt}
\label{sec:f1}

Under a single greedy sample, an untrained model re-evaluated twice on MATH-500 shows $9$ corruptions against $6$ learnings, a $\clr$ of $1.5$. The solve-rate estimator on the same samples reports $1$ and $0$. The ratio is two single-digit counts, so we report the rate alongside it: $7.5\%\,[4.6, 12.0]$ of problems change verdict, itself unstable across three independent executions of the same two-pass comparison at $7.5\%$, $8.0\%$ and $4.5\%$.

\textbf{The mechanism, tested rather than cited.} These flips are conventionally attributed to batching \citep{he2025nondeterminism} without a direct test. One is available: sample the same $200$ problems greedily twice over, once with $64$ requests in flight and once strictly serialized. Serializing cuts verdict flips from $16/200$ to $4/200$ ($8.0\%\,[5.0, 12.6]$ against $2.0\%\,[0.8, 5.0]$, Fisher $p = 0.005$) and extracted-answer changes from $25$ to $5$ ($p < 10^{-4}$). Batching is therefore the dominant cause, but not the only one. A fully serialized frozen model still flips $2\%$ of its greedy verdicts, enough to fabricate a $\clr$ from nothing on a small evaluation set. No care with our own request pattern removes it: whatever remains is server-side batching across tenants or nondeterminism below the API. A single-decode ledger cannot be repaired by decoding more carefully; it has to be replaced by an estimator. On the $1{,}163$-problem band the gap is wider still, $18.9\%$ of problems changing state under greedy decoding against $0.8\%$ under the estimator (Figure~\ref{fig:measure}, Table~\ref{tab:floors}).

\subsection{F2. No threshold repairs the expansion statistic}
\label{sec:f2}

Equation~\ref{eq:er} at $m = 1$ requires the base model to produce \emph{zero} correct samples and the evolved model \emph{at least one}. That is a comparison of two binomial estimates through asymmetric thresholds, and its null is not zero. On the frozen AIME control it gives $\er_1 = 7/25 = \mathbf{0.280}$: a model that did not train ``expands'' more than a quarter of the problems it had not reached, and pooled over all $110$ frozen comparisons the rate is $0.176$. On the matched ladder of Section~\ref{sec:results}, STaR scores $\er_1 = 0.320$ and majority-vote self-training $0.364$, $0.167$ and $0.192$ across three seeds. Those values straddle the pooled null and lie inside the $0.000$--$0.364$ range a frozen model produces across its own evaluations. The correct reading is not that they are nil, but that $\er_1$ overstates them by about a factor of two and cannot separate them from a model that never trained. It does separate distillation, at $0.545$--$0.667$.

\textbf{The obvious repair does not work.} All seven frozen ``expansions'' are exactly one correct sample in $128$, so $m = 2$ removes every one and its null on that comparison is exactly zero. The appearance rests on a single realization over $25$ problems. Every arm's checkpoint $0$ is an independent $k{=}128$ evaluation of the same untrained model, and with the ladder in place AIME carries \emph{eleven} of them, giving $110$ ordered pairs at no cost beyond experiments already run. Across those pairs $\er_2$ pools to $\mathbf{0.058}\,[0.038, 0.078]$ and reaches $0.136$ on the worst pair. The value we measure for majority-vote self-training, $1/21 = 0.048$, is therefore \emph{indistinguishable} from its own null (Wilson $[0.008, 0.23]$) rather than below it. Raising the threshold does not rescue the statistic: $\er_3$ has a null of $0.023\,[0.009, 0.038]$ and only $\er_5$ approaches zero, at $0.003\,[0.000, 0.009]$; on MATH-500 the $m{=}2$ null is $0.128$. The threshold that appeared to fix the statistic was fitted to the noise it was fitted on, and $m$ is benchmark- and $k$-specific in exactly the way the statistic it corrects is.

\begin{figure}[t]
\centering
\includegraphics[width=\textwidth]{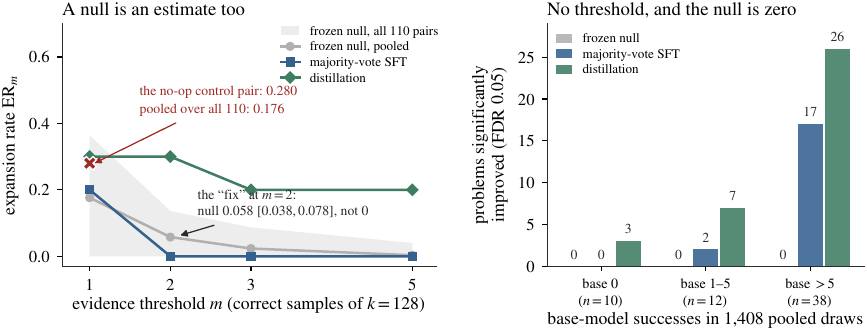}
\caption{\textbf{The expansion statistic needs a null, and one measurement of a null is not one.} \emph{Left:} $\er_m$ against the evidence threshold. At $m{=}1$, the criterion in use, a single frozen comparison sits at $0.280$, above majority-vote self-training. The shaded band is the range over all $110$ frozen comparisons the records contain, and it does not approach zero until $m{=}5$: at $m{=}2$, the natural repair, the pooled null is $0.058$ and the value it was meant to certify as null is $0.048$. \emph{Right:} the threshold-free test, by pooled base rate. Detections against a shared $1{,}408$-draw baseline under FDR control; the frozen null is zero in every stratum on all eleven held-out replicates.}
\label{fig:expansion}
\end{figure}

\textbf{What does work is to stop thresholding.} Pool every independent evaluation of the untrained model into one baseline: eleven on AIME, $1{,}408$ draws per problem, one denominator shared by all arms rather than a noise-selected set per arm. Then test each problem's endpoint count against it with a one-sided Fisher exact test under FDR control. The evidence threshold disappears and the arms become comparable. Four choices remain, the error rate, the one-sidedness, the pool size and the strata, and Appendix~\ref{app:testrobust} varies each (Table~\ref{tab:testsensitivity}). One belongs here, because a false-discovery-rate procedure is adaptive: an arm containing many true effects raises its own admissible cutoff, so detection counts need not be comparable across arms. Here they are: the cutoffs actually attained are $0.0095$ for distillation's $36$ detections and $0.0094$ for majority-vote self-training's $19$, and every comparison below is unchanged under Bonferroni, under one joint procedure across arms and problems, and at fixed thresholds. The null is measured the same way as everything else here: hold out one base evaluation, call it the ``after'', and run the identical test. It returns \textbf{zero} detections on all eleven held-out replicates, at every base rate (Figure~\ref{fig:expansion}, Table~\ref{tab:perproblem}); zero is itself an estimate, and the Clopper--Pearson bounds are $0.24$ per replicate and $0.005$ per test.

Under that test the dissociation is sharper than any thresholded version. On the matched ladder, distillation detects $36$ of $60$ problems and majority-vote self-training $19$; of the $21$ they disagree about, $19$ favor distillation, a paired exact $p = 2.2 \times 10^{-4}$ against the $p = 0.0015$ an unpaired test on mismatched denominators returns on the same data. Restricted to the $22$ problems whose pooled base rate is at most $5/1{,}408$, distillation detects $8$--$11$ per seed and the self-training arms $0$--$2$, the frozen control $0$ on all eleven replicates. We combine the per-seed comparisons rather than pooling detections across seeds, which is the operation F5 says three seeds cannot support: $p = 2.0\times10^{-6}$ against majority-vote SFT, $0.0078$ against STaR and $0.0019$ against the policy-gradient arm. Those $22$ are not all expansion candidates, however. Only $10$ have a pooled base rate of exactly zero, and there the union over seeds is $5$ against $2$, which is not significant ($p = 0.35$). Neither is the comparison on the ten low-base problems of the uncontaminated $2026$ half ($6$ against $2$, $p = 0.17$). Section~\ref{sec:results} keeps the two populations apart.

The small non-zero counts are stated rather than rounded away: applied to a stream of problems it can solve, self-training does occasionally raise a rarely-reached problem above the noise, at roughly a fifth of the teacher's rate. The comparison is in any case conservative in the teacher's disfavor, since the distillation arms truncate $36$--$46\%$ of completions even at an $8{,}192$-token cap (Appendix~\ref{app:testrobust}).

\subsection{The remaining failures}
\label{sec:f7}

Three of the remaining five are stated in summary and developed in Appendices~\ref{app:failures} and~\ref{app:collateral}. \textbf{F3}: a fixed token cap scores the most effective arm as the most destructive, because the distillation student adopts its teacher's verbose style and truncation rises from $16.5\%$ to $65.3\%$; that arm's corruption counts are excluded from the capability analysis, since severe generation-length drift under a fixed cap confounds capability loss with stylistic verbosity. \textbf{F4}: transition metrics carry an order of magnitude less power than accuracy, and a $112$-problem pilot yields $24$ events against the ${\sim}249$ per arm the comparison needs. \textbf{F6}: a $10$-prompt refusal probe yields a $10$-point safety finding that $50$ prompts dissolve.

\paragraph{F5. The comparison between methods is not stable across training seeds.} Two further seeds on a matched $175$-problem subsample invert the single-seed comparison (Table~\ref{tab:seeds}). STaR is stable ($\clr$ of $0.70$, $0.74$, $0.45$, accuracy up $+5.6$ to $+6.9$ points on every seed) and majority-vote self-training is not: $0.55$, $1.53$, $1.33$, with accuracy changes of $+4.0$, $-2.8$ and $-1.0$. A powered comparison at seed $0$ finds no difference ($p = 0.406$); pooling the two matched replicates for a Fisher test gives $p = 0.0076$, and we do not report it, because pooling seeds as a fixed effect is what the same paragraph says three seeds cannot support. The defensible reading is narrower and more useful: \emph{a majority-vote method's outcome is not determined by its design alone}. Same method, same data, same budget, run twice, landing on either side of $\clr = 1$.

\paragraph{F7. A null measured once is a point estimate, not a parameter.} Everything above compares a measured quantity to a measured null, and a null estimated once carries the same sampling error one level up. This design admits three instances: $\er_2$'s null of $0/25$ on a single frozen pair becomes $0.058\,[0.038, 0.078]$ over $110$ frozen comparisons, the band's corruption floor of $5$ becomes a median of $8$ once the control is given as many checkpoints as the arms, and the $0.188$ envelope on frozen solve-rate movement becomes $0.234$ over twenty pairs. None moves a headline, but the first removes the correction F2 would otherwise license. The remedy needs no additional experiments, and generalizes: \emph{every arm's checkpoint $0$ is an independent evaluation of the untrained model}. It needs a number attached, and ours is that four is not enough: over every subset of our eleven baselines the $\er_2$ null still ranges $0.022$--$0.098$ at $G = 4$, enough to certify a null that is not one, and only at $G = 9$ does it hold within $\pm 0.02$ of the value eleven give.

\section{What the controlled audit shows}
\label{sec:results}

\paragraph{A matched comparison.} Compared under their native protocols, self-training and distillation differ in more than the correctness filter: TTRL as published is applied to the $60$ AIME problems themselves and retains $50$ examples per round, where distillation draws from a disjoint stream and retains $247$, a $4.9\times$ volume gap on top of a stream mismatch. To eliminate the volume discrepancy and isolate the filter, we evaluate a ladder on one shared stream at matched volume, three seeds, so that only the correctness filter varies: the model's own majority vote, the ground-truth answer, an external teacher (Table~\ref{tab:ladder}), together with a policy-gradient arm using the same vote as a \emph{reward}, as TTRL specifies.

\begin{table}[t]
\centering
\caption{\textbf{The matched AIME ladder: one stream, one retained volume, one evaluation, and only the source of the correctness filter changing.} Detections are problems whose solve rate rises significantly against a pooled $1{,}408$-draw baseline of the untrained model, pooled over 11 independent evaluations (per-problem exact test, FDR $0.05$), split by how often that baseline reaches the problem at all. The frozen null is zero in every column. Expansion is only interpretable in the two low-base columns: there the self-training arms reach $0$--$2$ problems and the external teacher $8$--$11$, on every seed. Majority-vote accuracy also moves $+3.3$, $-1.5$ and $+0.7$ points across its three seeds while the teacher's moves $+13$ to $+16$: the instability of F5, on a second benchmark and under the matched protocol.}
\label{tab:ladder}
\footnotesize
\setlength{\tabcolsep}{3.4pt}
\begin{tabular}{llcrlrrrr}
\toprule
\textbf{filter} & \textbf{update} & \textbf{seed} & \textbf{accuracy} & \textbf{$\Delta$ (pts)} & \textbf{retained} & \textbf{base $0$} & \textbf{base $1$--$5$} & \textbf{base $>5$} \\
\midrule
majority vote & SFT & 0 & $0.180 \to 0.213$ & \dgain{$3.3$} & 250 & 0 & 2 & 17 \\
 &  & 1 & $0.178 \to 0.163$ & \dloss{$1.5$} & 251 & 1 & 0 & 3 \\
 &  & 2 & $0.179 \to 0.186$ & \dgain{$0.7$} & 251 & 0 & 0 & 4 \\
ground truth & SFT & 0 & $0.177 \to 0.195$ & \dgain{$1.8$} & 220 & 1 & 1 & 9 \\
external teacher$^{\dagger}$ & SFT & 0 & $0.180 \to 0.310$ & \dgain{$13.0$} & 247 & 3 & 7 & 26 \\
 &  & 1 & $0.177 \to 0.324$ & \dgain{$14.7$} & 248 & 4 & 7 & 26 \\
 &  & 2 & $0.177 \to 0.335$ & \dgain{$15.8$} & 246 & 2 & 6 & 25 \\
majority vote & policy grad.$^{\ddagger}$ & 0 & $0.176 \to 0.192$ & \dgain{$1.6$} & 1984 & 0 & 1 & 11 \\
\midrule
frozen control & none & --- & --- & --- & --- & $\mathbf{0}$ & $\mathbf{0}$ & $\mathbf{0}$ \\
\bottomrule
\end{tabular}

\vspace{0.3em}
{\small $^{\dagger}$positive control, not a self-evolution method. The frozen null is the same test with one base evaluation held out as the ``after'', over all $11$ such replicates. $^{\ddagger}$this arm collapsed into unbounded repetition on all three seeds and is $76.4\%$ truncated at the checkpoint shown; its detections are retained as a floor, since truncation can only suppress them, and its transition counts are excluded (Appendix~\ref{app:rl}). \emph{Retained} is the first round's count (Table~\ref{tab:protocol} gives all three): training examples for the SFT arms, rollouts for the policy-gradient arm, of which $135$, $143$ and $217$ of ${\sim}250$ carried any gradient.}
\end{table}

The ladder yields three results the unmatched comparison cannot. First, the dissociation survives matching: on the low-base pool the teacher reaches $8$--$11$ problems per seed and every self-training arm $0$--$2$. On the ten problems of that pool the base model never reaches at all, however, the split over seeds is $5$ against $2$ and is not significant, so what we claim is a low-base result and not an expansion result. Second, the dissociation is not an artifact of the teacher's larger gain. A logistic model on the per-problem counts, with terms for arm, stratum and their interaction, gives the teacher a further $\beta = 1.91\,[1.25, 2.56]$ on low-base problems beyond its overall lift ($p < 10^{-8}$), where majority-vote SFT gets $0.21\,[-0.60, 1.03]$ and the policy-gradient arm $-0.49$. Uniform multiplicative improvement accounts for the self-training arms and not for the teacher (Appendix~\ref{app:testrobust}). Third, the seed instability of F5 replicates on a second benchmark: majority-vote accuracy moves $+3.3$, $-1.5$ and $+0.7$ points across seeds, one of three net negative, while the teacher's moves $+13$ to $+16$ on all three.

\paragraph{Sharpening, at its true evidential weight.} Across every self-training arm, all problems newly solved at pass@1 were already reachable at pass@$k$; the informative evidence is nine problems rather than four hundred, since the band excludes base-unreached problems by construction (Appendix~\ref{app:counts}). The threshold-free test carries that claim at scale, and its null result is a bound rather than an absence. With eleven pooled baselines the test has $80\%$ power against a true post-training rate of $0.024$, and no self-training arm produced more than one correct sample in $128$ on any base-$0$ problem. None lifted such a problem above roughly a $2.4\%$ pass rate.

\paragraph{Reinforcement rather than filtering, and an arm outside the capability analysis.} Both self-training arms above are rejection-sampling SFT, whereas TTRL as published pushes the losing samples down. A policy-gradient arm on the band gives $19$ learned against $9$ corrupted ($\clr = 0.47$) and behaves; on AIME it does not survive at any of the three seeds F5 demands. The runs see identical data in identical order and agree at round one to within a percentage point: the vote is $77.0$--$77.9\%$ correct, and $135$--$136$ groups of roughly $250$ carry any gradient. They then diverge completely, collapsing into unbounded repetition after one update, two, and three, at $100\%$, $95.6\%$ and $76.4\%$ truncation. Every seed ends there, so the collapse belongs to the configuration and the round at which it arrives does not; the arm's transition counts are excluded under F3 exactly as distillation's are. What it is reported for is its own diagnostic. On seed~$2$, between the update that damaged the model and the one that finished it, the majority vote became \emph{more} accurate and \emph{more} unanimous ($77.9$ to $85.1\%$, share $0.78$ to $0.86$, over $248$ intact majorities) while the evaluation went from $47\%$ truncated to $96\%$. A supervision signal that reads healthy while the model it supervises is dying is the failure this paper is about (Table~\ref{tab:rlseeds}, Appendix~\ref{app:rl}).

\paragraph{What is removed is large, not a threshold flicker.} On the band both methods destroy problems solved at baseline, $106$ for STaR and $88$ for majority-vote SFT, while STaR's mean accuracy rises by $5.6$ points on that same run. Aggregate improvement and substantial destruction are the same run here, not alternatives. The floor needs the treatment the expansion null needed. The conventional two-checkpoint control gives $4$ learned and $5$ corrupted, where an arm has four checkpoints. A design-matched four-checkpoint no-op, built from the band's five independent evaluations, gives a median of $10$ and $8$ over $120$ orderings (Figure~\ref{fig:transitions}), so the arms clear it by $11$--$13\times$ rather than $18$--$21\times$. Nor is what they destroy a threshold flicker. Measured over all $20$ frozen pairs, a model that never trained never moved a solve rate by more than $0.234$. Yet $58$ of STaR's $106$ corruptions and $60$ of majority-vote SFT's $88$ exceed that: more than half, and more still under a stated quantile in place of the maximum, which is the less conservative choice (Table~\ref{tab:effects}, Figure~\ref{fig:effects}).

\paragraph{Two things we can only bound.} The discretization rule moves the counts without moving the conclusion, but under a pure endpoint rule the two self-training methods coincide exactly and at $z = 0$ their order reverses, so we rest nothing on the between-method comparison (Appendix~\ref{app:sensitivity}). And the refusal probe cannot resolve what it was built to measure: three replicates read $-0.015$, $-0.115$ and $-0.005$, the middle one significant and replicated by neither neighbor, so we report a bound (any change here is smaller than about eleven points) rather than a null result (Appendix~\ref{app:collateral}).

\section{Discussion}
\label{sec:discussion}

\paragraph{A reporting standard.} Seven practices follow, each of which caught an inverted result here: report a no-op control for \emph{every} statistic and \emph{with an interval}; match the control's design to the arm's; define correctness through a solve-rate estimator rather than a single decode; prefer a per-problem test against a pooled baseline to any thresholded expansion statistic; log generation length per checkpoint; size transition comparisons from a prior power analysis; and report at least three training seeds, scored on the same problems. As our own policy-gradient arm shows, the third seed can be the one that changes the finding. Five need no new experiments, being re-analyses of records a careful study already holds, but none of them on four frozen replicates, which is not enough for the null they would be read against. We report roughly a dozen $p$-values across seven failure modes and correct within each family rather than across them, which is stated here rather than left to be noticed (Appendix~\ref{app:testrobust}).

\paragraph{Limitations.} Three rounds and roughly $270$ optimizer steps of rank-$32$ LoRA, far short of the schedules we position against: what we establish is that \emph{this} regime does not expand, not that self-training cannot. Expansion rests on $22$ low-base problems on one benchmark, and on the $10$ the base model never reaches the comparison is not significant. Corruption rests on a band drawn from MATH \emph{training} problems almost certainly in the backbone's pretraining data and built to make corruption observable. One backbone family carries every claim (Appendix~\ref{app:limitations}).

\paragraph{Conclusion.} Transition-level auditing is the right response to the fact that accuracy conceals what a self-improving model gains and what it destroys. It is also a measurement problem in its own right: we isolate seven ways to get it wrong, two of them standard practice and one inside the correction for another. The expansion statistic in current use assigns a model that never trained a rate of $0.280$; the natural threshold repair takes that null to $0.058$ rather than zero, and calculating it would have returned $0.044$ and certified the same false conclusion. The replicates that bound a null already exist in any multi-arm study, provided there are enough of them.

\newpage

\subsubsection*{Ethics Statement}

This study evaluates language models on public mathematics benchmarks and releases only model-generated text about those problems; it involves no human subjects and no personal data. Two aspects warrant comment. The first is the out-of-domain refusal probe. Its prompts are deliberately mild and non-operational, asking for the shape of a harmful action rather than the means to carry one out, because their purpose is to elicit a refusal and not anything usable. The probe is released so that its result can be re-scored under a rule other than ours. What that result supports is a bound and not a clearance: we decline to claim that self-training causes no safety erosion here, and the data are consistent both with no effect and with one of practical size (Appendix~\ref{app:collateral}). Nothing in this paper should be read as evidence that a self-training method is safe to deploy without supervision. The second is that this paper reports, for every failure mode it documents, both the conclusion an uncontrolled analysis supports and the marginal cost of the control that rules it out. Presenting both is deliberate: a literature in which nulls are never re-measured is precisely the failure this paper is about. The distillation control uses an openly released teacher within the terms of its license, and Appendix~\ref{app:compute} reports the study's compute and monetary cost in full.

\subsubsection*{Reproducibility Statement}

Records for every run are released at \url{https://github.com/chengxuphd/phantom-gains}, reduced to per-problem counts: for each checkpoint, sampling mode and problem, the samples drawn, the samples correct, the completions truncated, and the tokens and cost they consumed. Every number and figure in this paper is a function of those counts and is recomputed from them by released code; none is transcribed by hand. The per-sample records they are reduced from --- one JSON line per generation, carrying the generated text and the extracted answer --- total roughly $15$\,GB, are impractical to distribute and matter only for re-grading under a different verifier; they are available on request, and the reduction that produced the released files is released with them. To support re-scoring under alternative refusal rules, the out-of-domain refusal probe similarly records individual outputs per generation; those records are what establish that the probe's reading does not reproduce across replicates (Appendix~\ref{app:collateral}). The evaluation subsets are released with the code, including the constructed difficulty band and the selection profile used to build it, together with the noise-floor runs for each benchmark. Section~\ref{sec:audit} and Appendix~\ref{app:protocol} state the training and sampling configuration in full, including temperature, top-$p$, token caps, learning rate, adapter rank and per-round training-example counts. The audited methods are reimplementations against Tinker\footnote{\url{https://thinkingmachines.ai/tinker/}}, a hosted LoRA training and sampling service, and are released; the policy-gradient arm uses that service's importance-sampling objective on group-centered advantages, with the group mean as baseline, no clipping and no off-policy correction beyond a single epoch per round.

\subsubsection*{Use of Large Language Models}

An LLM-based coding assistant was used to implement the evaluation harness and to run and monitor experiments, and assist with language editing and proofreading. All experimental design decisions, all interpretation of results, and all claims in this paper are the authors' own. Every number reported is computed by released code from released per-sample records; none was produced or transcribed by a language model.

\bibliography{iclr2027_conference}

@inproceedings{zelikman2022star,
 author = {Zelikman, Eric and Wu, Yuhuai and Mu, Jesse and Goodman, Noah},
 booktitle = {Advances in Neural Information Processing Systems},
 doi = {10.52202/068431-1126},
 editor = {S. Koyejo and S. Mohamed and A. Agarwal and D. Belgrave and K. Cho and A. Oh},
 pages = {15476--15488},
 publisher = {Curran Associates, Inc.},
 title = {STaR: Bootstrapping Reasoning With Reasoning},
 url = {https://proceedings.neurips.cc/paper_files/paper/2022/file/639a9a172c044fbb64175b5fad42e9a5-Paper-Conference.pdf},
 volume = {35},
 year = {2022}
}

@misc{xu2024benchmark,
      title={Benchmark Data Contamination of Large Language Models: A Survey}, 
      author={Cheng Xu and Shuhao Guan and Derek Greene and M-Tahar Kechadi},
      year={2024},
      eprint={2406.04244},
      archivePrefix={arXiv},
      primaryClass={cs.CL},
      url={https://arxiv.org/abs/2406.04244}, 
}

@inproceedings{xu2025dcr,
    title = "{DCR}: Quantifying Data Contamination in {LLM}s Evaluation",
    author = "Xu, Cheng  and
      Yan, Nan  and
      Guan, Shuhao  and
      Jin, Changhong  and
      Mei, Yuke  and
      Guo, Yibing  and
      Kechadi, Tahar",
    editor = "Christodoulopoulos, Christos  and
      Chakraborty, Tanmoy  and
      Rose, Carolyn  and
      Peng, Violet",
    booktitle = "Proceedings of the 2025 Conference on Empirical Methods in Natural Language Processing",
    month = nov,
    year = "2025",
    address = "Suzhou, China",
    publisher = "Association for Computational Linguistics",
    url = "https://aclanthology.org/2025.emnlp-main.1173/",
    doi = "10.18653/v1/2025.emnlp-main.1173",
    pages = "23013--23031",
    ISBN = "979-8-89176-332-6"
}

@inproceedings{sun2025the,
title={The Emperor's New Clothes in Benchmarking? A Rigorous Examination of Mitigation Strategies for {LLM} Benchmark Data Contamination},
author={Yifan Sun and Han Wang and Dongbai Li and Gang Wang and Huan Zhang},
booktitle={Forty-second International Conference on Machine Learning},
year={2025},
url={https://openreview.net/forum?id=TuvDxubEfE}
}

@inproceedings{zuo2025ttrl,
 author = {Zuo, Yuxin and Zhang, Kaiyan and Sheng, Li and Qu, Shang and Cui, Ganqu and Zhu, Xuekai and Li, Haozhan and zhang, yuchen and Long, Xinwei and Hua, Ermo and Qi, Biqing and Sun, Youbang and Ma, Zhiyuan and Yuan, Lifan and Ding, Ning and Zhou, Bowen},
 booktitle = {Advances in Neural Information Processing Systems},
 doi = {10.52202/085713-4376},
 editor = {D. Belgrave and C. Zhang and H. Lin and R. Pascanu and P. Koniusz and M. Ghassemi and N. Chen},
 pages = {131459--131483},
 publisher = {Curran Associates, Inc.},
 title = {TTRL: Test-Time Reinforcement Learning},
 url = {https://proceedings.neurips.cc/paper_files/paper/2025/file/be690ea16f005c174f6c4102a5970e67-Paper-Conference.pdf},
 volume = {38},
 year = {2025}
}

@inproceedings{zhao2025absolutezero,
 author = {Zhao, Andrew and Wu, Yiran and Wu, Tong and Xu, Quentin and Yue, Yang and Lin, Matthieu and Wang, Shenzhi and Wu, Qingyun and Zheng, Zilong and Huang, Gao},
 booktitle = {Advances in Neural Information Processing Systems},
 doi = {10.52202/085713-3534},
 editor = {D. Belgrave and C. Zhang and H. Lin and R. Pascanu and P. Koniusz and M. Ghassemi and N. Chen},
 pages = {105816--105879},
 publisher = {Curran Associates, Inc.},
 title = {Absolute Zero: Reinforced Self-play Reasoning with Zero Data},
 url = {https://proceedings.neurips.cc/paper_files/paper/2025/file/9837dc00ff67d176373268ed48042d49-Paper-Conference.pdf},
 volume = {38},
 year = {2025}
}

@inproceedings{huang2026rzero,
 author = {Huang, Chengsong and Yu, Wenhao and Wang, Xiaoyang and Zhang, Hongming and Li, Zongxia and Li, Ruosen and Huang, Jiaxin and Mi, Haitao and Yu, Dong},
 booktitle = {International Conference on Learning Representations},
 editor = {C. Vondrick and B. Hariharan and C. Raffel and L. Pinto and D. Yang and A. Faust},
 pages = {130770--130790},
 title = {R-Zero: Self-Evolving Reasoning LLM from Zero Data},
 url = {https://proceedings.iclr.cc/paper_files/paper/2026/file/d49b9aacebda61051166335af6fd3061-Paper-Conference.pdf},
 volume = {2026},
 year = {2026}
}

@inproceedings{zweiger2025self,
 author = {Zweiger, Adam and Pari, Jyo and Guo, Han and Kim, Yoon and Agrawal, Pulkit},
 booktitle = {Advances in Neural Information Processing Systems},
 doi = {10.52202/085713-2483},
 editor = {D. Belgrave and C. Zhang and H. Lin and R. Pascanu and P. Koniusz and M. Ghassemi and N. Chen},
 pages = {74084--74115},
 publisher = {Curran Associates, Inc.},
 title = {Self-Adapting Language Models},
 url = {https://proceedings.neurips.cc/paper_files/paper/2025/file/6b41e04c41726e2a60e456d0a2b961ab-Paper-Conference.pdf},
 volume = {38},
 year = {2025}
}

@inproceedings{yue2025does,
 author = {Yue, Yang and Chen, Zhiqi and Lu, Rui and Zhao, Andrew and Wang, Zhaokai and Yue, Yang and Song, Shiji and Huang, Gao},
 booktitle = {Advances in Neural Information Processing Systems},
 doi = {10.52202/085713-1933},
 editor = {D. Belgrave and C. Zhang and H. Lin and R. Pascanu and P. Koniusz and M. Ghassemi and N. Chen},
 pages = {57654--57689},
 publisher = {Curran Associates, Inc.},
 title = {Does Reinforcement Learning Really Incentivize Reasoning Capacity in LLMs Beyond the Base Model?},
 url = {https://proceedings.neurips.cc/paper_files/paper/2025/file/537d5aa768c2d534016a4d06f87bc8fb-Paper-Conference.pdf},
 volume = {38},
 year = {2025}
}

@inproceedings{mayilvahanan2026math,
 author = {Mayilvahanan, Prasanna and Dominguez-Olmedo, Ricardo and Wiedemer, Thadd\"{a}us and Brendel, Wieland},
 booktitle = {International Conference on Learning Representations},
 editor = {C. Vondrick and B. Hariharan and C. Raffel and L. Pinto and D. Yang and A. Faust},
 pages = {68471--68489},
 title = {MATH-Beyond: A Benchmark for RL to Expand Beyond the Base Model},
 url = {https://proceedings.iclr.cc/paper_files/paper/2026/file/6fdf57c71bc1f1ee29014b8dc52e723f-Paper-Conference.pdf},
 volume = {2026},
 year = {2026}
}

@misc{lin2026detecting,
      title={Detecting and Mitigating the Correct-Answer Extinction Window in Test-Time Reinforcement Learning with Majority Voting}, 
      author={Hongxiang Lin and Zhirui Kuai and Erpeng Xue and Lei Wang},
      year={2026},
      eprint={2605.19444},
      archivePrefix={arXiv},
      primaryClass={cs.LG},
      url={https://arxiv.org/abs/2605.19444}, 
}

@misc{yu2026self,
      title={Do Self-Evolving Agents Forget? Capability Degradation and Preservation in Lifelong LLM Agent Adaptation}, 
      author={Ye Yu and Xiaopeng Yuan and Haibo Jin and Heming Liu and Yaoning Yu and Haohan Wang},
      year={2026},
      eprint={2605.09315},
      archivePrefix={arXiv},
      primaryClass={cs.AI},
      url={https://arxiv.org/abs/2605.09315}, 
}

@inproceedings{shao2026your,
 author = {Shao, Shuai and Ren, Qihan and Liu, Dongrui and Qian, Chen and Wei, Boyi and Guo, Dadi and Yang, Jingyi and Song, Xinhao and Zhang, Linfeng and Zhang, Weinan and Shao, Jing},
 booktitle = {International Conference on Learning Representations},
 editor = {C. Vondrick and B. Hariharan and C. Raffel and L. Pinto and D. Yang and A. Faust},
 pages = {99728--99793},
 title = {Your Agent May Misevolve: Emergent Risks in Self-evolving LLM Agents},
 url = {https://proceedings.iclr.cc/paper_files/paper/2026/file/a24cd16bc361afa78e57d31d34f3d936-Paper-Conference.pdf},
 volume = {2026},
 year = {2026}
}

@inproceedings{lightman2024verify,
 author = {Lightman, Hunter and Kosaraju, Vineet and Burda, Yuri and Edwards, Harrison and Baker, Bowen and Lee, Teddy and Leike, Jan and Schulman, John  and Sutskever, Ilya and Cobbe, Karl},
 booktitle = {International Conference on Learning Representations},
 editor = {B. Kim and Y. Yue and S. Chaudhuri and K. Fragkiadaki and M. Khan and Y. Sun},
 pages = {39578--39601},
 title = {Let\textquotesingle s Verify Step by Step},
 url = {https://proceedings.iclr.cc/paper_files/paper/2024/file/aca97732e30bcf1303bc22ac3924fd16-Paper-Conference.pdf},
 volume = {2024},
 year = {2024}
}

@inproceedings{hendrycks2021measuring,
 author = {Hendrycks, Dan and Burns, Collin and Kadavath, Saurav and Arora, Akul and Basart, Steven and Tang, Eric and Song, Dawn and Steinhardt, Jacob},
 booktitle = {Proceedings of the Neural Information Processing Systems Track on Datasets and Benchmarks},
 editor = {J. Vanschoren and S. Yeung},
 pages = {},
 title = {Measuring Mathematical Problem Solving With the MATH Dataset},
 url = {https://datasets-benchmarks-proceedings.neurips.cc/paper_files/paper/2021/file/be83ab3ecd0db773eb2dc1b0a17836a1-Paper-round2.pdf},
 volume = {1},
 year = {2021}
}

@misc{chen2021evaluating,
      title={Evaluating Large Language Models Trained on Code}, 
      author={Mark Chen and Jerry Tworek and Heewoo Jun and Qiming Yuan and Henrique Ponde de Oliveira Pinto and Jared Kaplan and Harri Edwards and Yuri Burda and Nicholas Joseph and Greg Brockman and Alex Ray and Raul Puri and Gretchen Krueger and Michael Petrov and Heidy Khlaaf and Girish Sastry and Pamela Mishkin and Brooke Chan and Scott Gray and Nick Ryder and Mikhail Pavlov and Alethea Power and Lukasz Kaiser and Mohammad Bavarian and Clemens Winter and Philippe Tillet and Felipe Petroski Such and Dave Cummings and Matthias Plappert and Fotios Chantzis and Elizabeth Barnes and Ariel Herbert-Voss and William Hebgen Guss and Alex Nichol and Alex Paino and Nikolas Tezak and Jie Tang and Igor Babuschkin and Suchir Balaji and Shantanu Jain and William Saunders and Christopher Hesse and Andrew N. Carr and Jan Leike and Josh Achiam and Vedant Misra and Evan Morikawa and Alec Radford and Matthew Knight and Miles Brundage and Mira Murati and Katie Mayer and Peter Welinder and Bob McGrew and Dario Amodei and Sam McCandlish and Ilya Sutskever and Wojciech Zaremba},
      year={2021},
      eprint={2107.03374},
      archivePrefix={arXiv},
      primaryClass={cs.LG},
      url={https://arxiv.org/abs/2107.03374}, 
}

@inproceedings{hu2022lora,
title={Lo{RA}: Low-Rank Adaptation of Large Language Models},
author={Edward J Hu and yelong shen and Phillip Wallis and Zeyuan Allen-Zhu and Yuanzhi Li and Shean Wang and Lu Wang and Weizhu Chen},
booktitle={International Conference on Learning Representations},
year={2022},
url={https://openreview.net/forum?id=nZeVKeeFYf9}
}

@inproceedings{wang2023selfconsistency,
title={Self-Consistency Improves Chain of Thought Reasoning in Language Models},
author={Xuezhi Wang and Jason Wei and Dale Schuurmans and Quoc V Le and Ed H. Chi and Sharan Narang and Aakanksha Chowdhery and Denny Zhou},
booktitle={The Eleventh International Conference on Learning Representations },
year={2023},
url={https://openreview.net/forum?id=1PL1NIMMrw}
}

@misc{fang2025comprehensive,
      title={A Comprehensive Survey of Self-Evolving AI Agents: A New Paradigm Bridging Foundation Models and Lifelong Agentic Systems}, 
      author={Jinyuan Fang and Yanwen Peng and Xi Zhang and Yingxu Wang and Xinhao Yi and Guibin Zhang and Yi Xu and Bin Wu and Siwei Liu and Zihao Li and Zhaochun Ren and Nikos Aletras and Xi Wang and Han Zhou and Zaiqiao Meng},
      year={2025},
      eprint={2508.07407},
      archivePrefix={arXiv},
      primaryClass={cs.AI},
      url={https://arxiv.org/abs/2508.07407}, 
}

@misc{tao2024survey,
      title={A Survey on Self-Evolution of Large Language Models}, 
      author={Zhengwei Tao and Ting-En Lin and Xiancai Chen and Hangyu Li and Yuchuan Wu and Yongbin Li and Zhi Jin and Fei Huang and Dacheng Tao and Jingren Zhou},
      year={2024},
      eprint={2404.14387},
      archivePrefix={arXiv},
      primaryClass={cs.CL},
      url={https://arxiv.org/abs/2404.14387}, 
}

@misc{yuan2026diversity,
      title={Understanding Diversity Collapse in RLVR via the Lens of Overtraining}, 
      author={Suqin Yuan and Jinkun Chen and Jiyang Zheng and Muyang Li and Lei Feng and Dadong Wang and Tao Xiang and Tongliang Liu and Bo An},
      year={2026},
      eprint={2606.15455},
      archivePrefix={arXiv},
      primaryClass={cs.LG},
      url={https://arxiv.org/abs/2606.15455}, 
}

@misc{wu2026invisibleleash,
      title={The Invisible Leash: Why RLVR May or May Not Escape Its Origin}, 
      author={Fang Wu and Weihao Xuan and Ximing Lu and Mingjie Liu and Yi Dong and Zaid Harchaoui and Yejin Choi},
      year={2026},
      eprint={2507.14843},
      archivePrefix={arXiv},
      primaryClass={cs.LG},
      url={https://arxiv.org/abs/2507.14843}, 
}

@article{he2025nondeterminism,
  author = {Horace He and Thinking Machines Lab},
  title = {Defeating Nondeterminism in LLM Inference},
  journal = {Thinking Machines Lab: Connectionism},
  year = {2025},
  url = {https://thinkingmachines.ai/blog/defeating-nondeterminism-in-llm-inference/},
  doi = {10.64434/tml.20250910},
}

@misc{strozzi2026teacherfree,
      title={Teacher-Free Self-Training Amplifies but Does Not Compound: A Pass@$K$ Crossover on a Free-Verifier Domain}, 
      author={Igor Lima Strozzi},
      year={2026},
      eprint={2606.07856},
      archivePrefix={arXiv},
      primaryClass={cs.LG},
      url={https://arxiv.org/abs/2606.07856}, 
}

@inproceedings{huang2025self,
 author = {Huang, Audrey and Block, Adam and Foster, Dylan and Rohatgi, Dhruv and Zhang, Cyril and Simchowitz, Max and Ash, Jordan and Krishnamurthy, Akshay},
 booktitle = {International Conference on Learning Representations},
 editor = {Y. Yue and A. Garg and N. Peng and F. Sha and R. Yu},
 pages = {76687--76739},
 title = {Self-Improvement in Language Models: The Sharpening Mechanism},
 url = {https://proceedings.iclr.cc/paper_files/paper/2025/file/bee8c2bc757f6bbc3efd7cf1b979f0c9-Paper-Conference.pdf},
 volume = {2025},
 year = {2025}
}

@misc{yuan2023scaling,
      title={Scaling Relationship on Learning Mathematical Reasoning with Large Language Models}, 
      author={Zheng Yuan and Hongyi Yuan and Chengpeng Li and Guanting Dong and Keming Lu and Chuanqi Tan and Chang Zhou and Jingren Zhou},
      year={2023},
      eprint={2308.01825},
      archivePrefix={arXiv},
      primaryClass={cs.CL},
      url={https://arxiv.org/abs/2308.01825}, 
}

@article{
singh2024beyond,
title={Beyond Human Data: Scaling Self-Training for Problem-Solving with Language Models},
author={Avi Singh and John D Co-Reyes and Rishabh Agarwal and Ankesh Anand and Piyush Patil and Xavier Garcia and Peter J Liu and James Harrison and Jaehoon Lee and Kelvin Xu and Aaron T Parisi and Abhishek Kumar and Alexander A Alemi and Alex Rizkowsky and Azade Nova and Ben Adlam and Bernd Bohnet and Gamaleldin Fathy Elsayed and Hanie Sedghi and Igor Mordatch and Isabelle Simpson and Izzeddin Gur and Jasper Snoek and Jeffrey Pennington and Jiri Hron and Kathleen Kenealy and Kevin Swersky and Kshiteej Mahajan and Laura A Culp and Lechao Xiao and Maxwell Bileschi and Noah Constant and Roman Novak and Rosanne Liu and Tris Warkentin and Yamini Bansal and Ethan Dyer and Behnam Neyshabur and Jascha Sohl-Dickstein and Noah Fiedel},
journal={Transactions on Machine Learning Research},
issn={2835-8856},
year={2024},
url={https://openreview.net/forum?id=lNAyUngGFK},
note={Expert Certification}
}

@inproceedings{wang2023selfinstruct,
    title = "Self-Instruct: Aligning Language Models with Self-Generated Instructions",
    author = "Wang, Yizhong  and
      Kordi, Yeganeh  and
      Mishra, Swaroop  and
      Liu, Alisa  and
      Smith, Noah A.  and
      Khashabi, Daniel  and
      Hajishirzi, Hannaneh",
    editor = "Rogers, Anna  and
      Boyd-Graber, Jordan  and
      Okazaki, Naoaki",
    booktitle = "Proceedings of the 61st Annual Meeting of the Association for Computational Linguistics (Volume 1: Long Papers)",
    month = jul,
    year = "2023",
    address = "Toronto, Canada",
    publisher = "Association for Computational Linguistics",
    url = "https://aclanthology.org/2023.acl-long.754/",
    doi = "10.18653/v1/2023.acl-long.754",
    pages = "13484--13508"
}

@inproceedings{huang2023large,
    title = "Large Language Models Can Self-Improve",
    author = "Huang, Jiaxin  and
      Gu, Shixiang  and
      Hou, Le  and
      Wu, Yuexin  and
      Wang, Xuezhi  and
      Yu, Hongkun  and
      Han, Jiawei",
    editor = "Bouamor, Houda  and
      Pino, Juan  and
      Bali, Kalika",
    booktitle = "Proceedings of the 2023 Conference on Empirical Methods in Natural Language Processing",
    month = dec,
    year = "2023",
    address = "Singapore",
    publisher = "Association for Computational Linguistics",
    url = "https://aclanthology.org/2023.emnlp-main.67/",
    doi = "10.18653/v1/2023.emnlp-main.67",
    pages = "1051--1068"
}

@misc{zhou2026evolving,
      title={Evolving Language Models without Labels: Majority Drives Selection, Novelty Promotes Variation}, 
      author={Yujun Zhou and Zhenwen Liang and Haolin Liu and Wenhao Yu and Kishan Panaganti and Linfeng Song and Dian Yu and Xiangliang Zhang and Haitao Mi and Dong Yu},
      year={2026},
      eprint={2509.15194},
      archivePrefix={arXiv},
      primaryClass={cs.LG},
      url={https://arxiv.org/abs/2509.15194}, 
}

@inproceedings{moradi2025continuous,
title={Continuous Self-Improvement of Large Language Models by Test-time Training with Verifier-Driven Sample Selection},
author={Mohammad Mahdi Moradi and Hossam Amer and Sudhir Mudur and Weiwei Zhang and Yang Liu and Walid Ahmed},
booktitle={AI That Keeps Up: NeurIPS 2025 Workshop on Continual and Compatible Foundation Model Updates},
year={2025},
url={https://openreview.net/forum?id=6ahliSpvQ0}
}

@misc{acikgoz2025selfimproving,
      title={Self-Improving LLM Agents at Test-Time}, 
      author={Emre Can Acikgoz and Cheng Qian and Heng Ji and Dilek Hakkani-Tür and Gokhan Tur},
      year={2025},
      eprint={2510.07841},
      archivePrefix={arXiv},
      primaryClass={cs.LG},
      url={https://arxiv.org/abs/2510.07841}, 
}

@inproceedings{
hosseini2024vstar,
title={V-{ST}aR: Training Verifiers for Self-Taught Reasoners},
author={Arian Hosseini and Xingdi Yuan and Nikolay Malkin and Aaron Courville and Alessandro Sordoni and Rishabh Agarwal},
booktitle={First Conference on Language Modeling},
year={2024},
url={https://openreview.net/forum?id=stmqBSW2dV}
}

@misc{
chen2026enhancing,
title={Enhancing Reasoning in Large Language Models via Entropy-Aware Self-Evolution},
author={Lai Chen and Yao Zhu and Xin Ye and Peng Lin and Xiangyang Ji and Xiang Tian},
year={2026},
url={https://openreview.net/forum?id=nXENWUSRMw}
}

@InProceedings{yuan2024selfrewarding,
  title = 	 {Self-Rewarding Language Models},
  author =       {Yuan, Weizhe and Pang, Richard Yuanzhe and Cho, Kyunghyun and Li, Xian and Sukhbaatar, Sainbayar and Xu, Jing and Weston, Jason E},
  booktitle = 	 {Proceedings of the 41st International Conference on Machine Learning},
  pages = 	 {57905--57923},
  year = 	 {2024},
  editor = 	 {Salakhutdinov, Ruslan and Kolter, Zico and Heller, Katherine and Weller, Adrian and Oliver, Nuria and Scarlett, Jonathan and Berkenkamp, Felix},
  volume = 	 {235},
  series = 	 {Proceedings of Machine Learning Research},
  month = 	 {21--27 Jul},
  publisher =    {PMLR},
  url = 	 {https://proceedings.mlr.press/v235/yuan24d.html}
}

@InProceedings{chen2024self,
  title = 	 {Self-Play Fine-Tuning Converts Weak Language Models to Strong Language Models},
  author =       {Chen, Zixiang and Deng, Yihe and Yuan, Huizhuo and Ji, Kaixuan and Gu, Quanquan},
  booktitle = 	 {Proceedings of the 41st International Conference on Machine Learning},
  pages = 	 {6621--6642},
  year = 	 {2024},
  editor = 	 {Salakhutdinov, Ruslan and Kolter, Zico and Heller, Katherine and Weller, Adrian and Oliver, Nuria and Scarlett, Jonathan and Berkenkamp, Felix},
  volume = 	 {235},
  series = 	 {Proceedings of Machine Learning Research},
  month = 	 {21--27 Jul},
  publisher =    {PMLR},
  url = 	 {https://proceedings.mlr.press/v235/chen24j.html}
}

@inproceedings{srivastava2025debate,
    title = "{DEBATE}, {TRAIN}, {EVOLVE}: {S}elf{-}{E}volution of Language Model Reasoning",
    author = "Srivastava, Gaurav  and
      Bi, Zhenyu  and
      Lu, Meng  and
      Wang, Xuan",
    editor = "Christodoulopoulos, Christos  and
      Chakraborty, Tanmoy  and
      Rose, Carolyn  and
      Peng, Violet",
    booktitle = "Proceedings of the 2025 Conference on Empirical Methods in Natural Language Processing",
    month = nov,
    year = "2025",
    address = "Suzhou, China",
    publisher = "Association for Computational Linguistics",
    url = "https://aclanthology.org/2025.emnlp-main.1666/",
    doi = "10.18653/v1/2025.emnlp-main.1666",
    pages = "32764--32810",
    ISBN = "979-8-89176-332-6"
}

@misc{lu2024self,
      title={SELF: Self-Evolution with Language Feedback}, 
      author={Jianqiao Lu and Wanjun Zhong and Wenyong Huang and Yufei Wang and Qi Zhu and Fei Mi and Baojun Wang and Weichao Wang and Xingshan Zeng and Lifeng Shang and Xin Jiang and Qun Liu},
      year={2024},
      eprint={2310.00533},
      archivePrefix={arXiv},
      primaryClass={cs.CL},
      url={https://arxiv.org/abs/2310.00533}, 
}

@inproceedings{
zelikman2024quietstar,
title={Quiet-{ST}aR: Language Models Can Teach Themselves to Think Before Speaking},
author={Eric Zelikman and Georges Raif Harik and Yijia Shao and Varuna Jayasiri and Nick Haber and Noah Goodman},
booktitle={First Conference on Language Modeling},
year={2024},
url={https://openreview.net/forum?id=oRXPiSOGH9}
}

@article{
gao2026survey,
title={A Survey of Self-Evolving Agents: What, When, How, and Where to Evolve on the Path to Artificial Super Intelligence},
author={Huan-ang Gao and Jiayi Geng and Wenyue Hua and Mengkang Hu and Xinzhe Juan and Hongzhang Liu and Shilong Liu and Jiahao Qiu and Xuan Qi and Qihan Ren and Yiran Wu and Hongru WANG and Han Xiao and Yuhang Zhou and Shaokun Zhang and Jiayi Zhang and Jinyu Xiang and Yixiong Fang and Qiwen Zhao and Dongrui Liu and Cheng Qian and Zhenhailong Wang and Minda Hu and Huazheng Wang and Qingyun Wu and Heng Ji and Mengdi Wang},
journal={Transactions on Machine Learning Research},
issn={2835-8856},
year={2026},
url={https://openreview.net/forum?id=CTr3bovS5F},
note={Survey Certification}
}

@inproceedings{madaan2023selfrefine,
 author = {Madaan, Aman and Tandon, Niket and Gupta, Prakhar and Hallinan, Skyler and Gao, Luyu and Wiegreffe, Sarah and Alon, Uri and Dziri, Nouha and Prabhumoye, Shrimai and Yang, Yiming and Gupta, Shashank and Majumder, Bodhisattwa Prasad and Hermann, Katherine and Welleck, Sean and Yazdanbakhsh, Amir and Clark, Peter},
 booktitle = {Advances in Neural Information Processing Systems},
 doi = {10.52202/075280-2019},
 editor = {A. Oh and T. Naumann and A. Globerson and K. Saenko and M. Hardt and S. Levine},
 pages = {46534--46594},
 publisher = {Curran Associates, Inc.},
 title = {Self-Refine: Iterative Refinement with Self-Feedback},
 url = {https://proceedings.neurips.cc/paper_files/paper/2023/file/91edff07232fb1b55a505a9e9f6c0ff3-Paper-Conference.pdf},
 volume = {36},
 year = {2023}
}

@inproceedings{huang2024large,
 author = {Huang, Jie and Chen, Xinyun and Mishra, Swaroop and Zheng, Huaixiu Steven and Yu, Adams and Song, Xinying and Zhou, Denny},
 booktitle = {International Conference on Learning Representations},
 editor = {B. Kim and Y. Yue and S. Chaudhuri and K. Fragkiadaki and M. Khan and Y. Sun},
 pages = {32808--32824},
 title = {Large Language Models Cannot Self-Correct Reasoning Yet},
 url = {https://proceedings.iclr.cc/paper_files/paper/2024/file/8b4add8b0aa8749d80a34ca5d941c355-Paper-Conference.pdf},
 volume = {2024},
 year = {2024}
}

@inproceedings{song2025mind,
 author = {Song, Yuda and Zhang, Hanlin and Eisenach, Carson and Kakade, Sham and Foster, Dean and Ghai, Udaya},
 booktitle = {International Conference on Learning Representations},
 editor = {Y. Yue and A. Garg and N. Peng and F. Sha and R. Yu},
 pages = {39894--39931},
 title = {Mind the Gap: Examining the Self-Improvement Capabilities of Large Language Models},
 url = {https://proceedings.iclr.cc/paper_files/paper/2025/file/63943ee9fe347f3d95892cf87d9a42e6-Paper-Conference.pdf},
 volume = {2025},
 year = {2025}
}

@inproceedings{liu2025prorl,
 author = {Liu, Mingjie and Diao, Shizhe and Lu, Ximing and Hu, Jian and Dong, Xin and Choi, Yejin and Kautz, Jan and Dong, Yi},
 booktitle = {Advances in Neural Information Processing Systems},
 doi = {10.52202/085713-0608},
 editor = {D. Belgrave and C. Zhang and H. Lin and R. Pascanu and P. Koniusz and M. Ghassemi and N. Chen},
 pages = {17998--18031},
 publisher = {Curran Associates, Inc.},
 title = {ProRL: Prolonged Reinforcement Learning Expands Reasoning Boundaries in Large Language Models},
 url = {https://proceedings.neurips.cc/paper_files/paper/2025/file/1a22b912945fb7c0bdd079e792b31b6f-Paper-Conference.pdf},
 volume = {38},
 year = {2025}
}

@inproceedings{
qi2026generalization,
title={On the Generalization Gap in Self-Evolving Language Model Reasoning},
author={Zhenting Qi and Susanna Maria Baby and Stefanie Anna Baby and Kan Yuan and Andrew Tomkins and Tu Vu and Da-Cheng Juan and Cyrus Rashtchian},
booktitle={Forty-third International Conference on Machine Learning},
year={2026},
url={https://openreview.net/forum?id=mnUidYi5qO}
}

@inproceedings{mukherjee2025reinforcement,
 author = {Mukherjee, Sagnik and Yuan, Lifan and Hakkani-Tur, Dilek and Peng, Hao},
 booktitle = {Advances in Neural Information Processing Systems},
 doi = {10.52202/085713-4399},
 editor = {D. Belgrave and C. Zhang and H. Lin and R. Pascanu and P. Koniusz and M. Ghassemi and N. Chen},
 pages = {132119--132138},
 publisher = {Curran Associates, Inc.},
 title = {Reinforcement Learning Finetunes Small Subnetworks in Large Language Models},
 url = {https://proceedings.neurips.cc/paper_files/paper/2025/file/bf235a1d6780afd979f2f81676f43413-Paper-Conference.pdf},
 volume = {38},
 year = {2025}
}

@misc{cui2025entropy,
      title={The Entropy Mechanism of Reinforcement Learning for Reasoning Language Models}, 
      author={Ganqu Cui and Yuchen Zhang and Jiacheng Chen and Lifan Yuan and Zhi Wang and Yuxin Zuo and Haozhan Li and Yuchen Fan and Huayu Chen and Weize Chen and Zhiyuan Liu and Hao Peng and Lei Bai and Wanli Ouyang and Yu Cheng and Bowen Zhou and Ning Ding},
      year={2025},
      eprint={2505.22617},
      archivePrefix={arXiv},
      primaryClass={cs.LG},
      url={https://arxiv.org/abs/2505.22617}, 
}

@inproceedings{
shao2026spurious,
title={Spurious Rewards: Rethinking Training Signals in {RLVR}},
author={Rulin Shao and Shuyue Stella Li and Rui Xin and Scott Geng and Yiping Wang and Sewoong Oh and Simon Shaolei Du and Nathan Lambert and Sewon Min and Ranjay Krishna and Yulia Tsvetkov and Hannaneh Hajishirzi and Pang Wei Koh and Luke Zettlemoyer},
booktitle={Forty-third International Conference on Machine Learning},
year={2026},
url={https://openreview.net/forum?id=tqTNOpkP5j}
}

@inproceedings{wang2025reinforcement,
 author = {Wang, Yiping and Yang, Qing and Zeng, Zhiyuan and Ren, Liliang and Liu, Liyuan and Peng, Baolin and Cheng, Hao and He, Xuehai and Wang, Kuan and Gao, Jianfeng and Chen, Weizhu and Wang, Shuohang and Du, Simon and shen, yelong},
 booktitle = {Advances in Neural Information Processing Systems},
 doi = {10.52202/085713-4093},
 editor = {D. Belgrave and C. Zhang and H. Lin and R. Pascanu and P. Koniusz and M. Ghassemi and N. Chen},
 pages = {122721--122764},
 publisher = {Curran Associates, Inc.},
 title = {Reinforcement Learning for Reasoning in Large Language Models with One Training Example},
 url = {https://proceedings.neurips.cc/paper_files/paper/2025/file/b1ea3f93167c55f3fc2999b68a170ff3-Paper-Conference.pdf},
 volume = {38},
 year = {2025}
}

@misc{shao2024deepseekmath,
      title={DeepSeekMath: Pushing the Limits of Mathematical Reasoning in Open Language Models}, 
      author={Zhihong Shao and Peiyi Wang and Qihao Zhu and Runxin Xu and Junxiao Song and Xiao Bi and Haowei Zhang and Mingchuan Zhang and Y. K. Li and Y. Wu and Daya Guo},
      year={2024},
      eprint={2402.03300},
      archivePrefix={arXiv},
      primaryClass={cs.CL},
      url={https://arxiv.org/abs/2402.03300}, 
}

@article{deepseek2025r1,
   title={DeepSeek-R1 incentivizes reasoning in LLMs through reinforcement learning},
   volume={645},
   ISSN={1476-4687},
   url={http://dx.doi.org/10.1038/s41586-025-09422-z},
   DOI={10.1038/s41586-025-09422-z},
   number={8081},
   journal={Nature},
   publisher={Springer Science and Business Media LLC},
   author={Guo, Daya and Yang, Dejian and Zhang, Haowei and Song, Junxiao and Wang, Peiyi and Zhu, Qihao and Xu, Runxin and Zhang, Ruoyu and Ma, Shirong and Bi, Xiao and Zhang, Xiaokang and Yu, Xingkai and Wu, Yu and Wu, Z. F. and Gou, Zhibin and Shao, Zhihong and Li, Zhuoshu and Gao, Ziyi and Liu, Aixin and Xue, Bing and Wang, Bingxuan and Wu, Bochao and Feng, Bei and Lu, Chengda and Zhao, Chenggang and Deng, Chengqi and Ruan, Chong and Dai, Damai and Chen, Deli and Ji, Dongjie and Li, Erhang and Lin, Fangyun and Dai, Fucong and Luo, Fuli and Hao, Guangbo and Chen, Guanting and Li, Guowei and Zhang, H. and Xu, Hanwei and Ding, Honghui and Gao, Huazuo and Qu, Hui and Li, Hui and Guo, Jianzhong and Li, Jiashi and Chen, Jingchang and Yuan, Jingyang and Tu, Jinhao and Qiu, Junjie and Li, Junlong and Cai, J. L. and Ni, Jiaqi and Liang, Jian and Chen, Jin and Dong, Kai and Hu, Kai and You, Kaichao and Gao, Kaige and Guan, Kang and Huang, Kexin and Yu, Kuai and Wang, Lean and Zhang, Lecong and Zhao, Liang and Wang, Litong and Zhang, Liyue and Xu, Lei and Xia, Leyi and Zhang, Mingchuan and Zhang, Minghua and Tang, Minghui and Zhou, Mingxu and Li, Meng and Wang, Miaojun and Li, Mingming and Tian, Ning and Huang, Panpan and Zhang, Peng and Wang, Qiancheng and Chen, Qinyu and Du, Qiushi and Ge, Ruiqi and Zhang, Ruisong and Pan, Ruizhe and Wang, Runji and Chen, R. J. and Jin, R. L. and Chen, Ruyi and Lu, Shanghao and Zhou, Shangyan and Chen, Shanhuang and Ye, Shengfeng and Wang, Shiyu and Yu, Shuiping and Zhou, Shunfeng and Pan, Shuting and Li, S. S. and Zhou, Shuang and Wu, Shaoqing and Yun, Tao and Pei, Tian and Sun, Tianyu and Wang, T. and Zeng, Wangding and Liu, Wen and Liang, Wenfeng and Gao, Wenjun and Yu, Wenqin and Zhang, Wentao and Xiao, W. L. and An, Wei and Liu, Xiaodong and Wang, Xiaohan and Chen, Xiaokang and Nie, Xiaotao and Cheng, Xin and Liu, Xin and Xie, Xin and Liu, Xingchao and Yang, Xinyu and Li, Xinyuan and Su, Xuecheng and Lin, Xuheng and Li, X. Q. and Jin, Xiangyue and Shen, Xiaojin and Chen, Xiaosha and Sun, Xiaowen and Wang, Xiaoxiang and Song, Xinnan and Zhou, Xinyi and Wang, Xianzu and Shan, Xinxia and Li, Y. K. and Wang, Y. Q. and Wei, Y. X. and Zhang, Yang and Xu, Yanhong and Li, Yao and Zhao, Yao and Sun, Yaofeng and Wang, Yaohui and Yu, Yi and Zhang, Yichao and Shi, Yifan and Xiong, Yiliang and He, Ying and Piao, Yishi and Wang, Yisong and Tan, Yixuan and Ma, Yiyang and Liu, Yiyuan and Guo, Yongqiang and Ou, Yuan and Wang, Yuduan and Gong, Yue and Zou, Yuheng and He, Yujia and Xiong, Yunfan and Luo, Yuxiang and You, Yuxiang and Liu, Yuxuan and Zhou, Yuyang and Zhu, Y. X. and Huang, Yanping and Li, Yaohui and Zheng, Yi and Zhu, Yuchen and Ma, Yunxian and Tang, Ying and Zha, Yukun and Yan, Yuting and Ren, Z. Z. and Ren, Zehui and Sha, Zhangli and Fu, Zhe and Xu, Zhean and Xie, Zhenda and Zhang, Zhengyan and Hao, Zhewen and Ma, Zhicheng and Yan, Zhigang and Wu, Zhiyu and Gu, Zihui and Zhu, Zijia and Liu, Zijun and Li, Zilin and Xie, Ziwei and Song, Ziyang and Pan, Zizheng and Huang, Zhen and Xu, Zhipeng and Zhang, Zhongyu and Zhang, Zhen},
   year={2025},
   month=Sep, pages={633–638} }

@inproceedings{yu2025dapo,
 author = {Yu, Qiying and Zhang, Zheng and Zhu, Ruofei and Yuan, Yufeng and Zuo, Xiaochen and Yue, Yu and Dai, Weinan and Fan, Tiantian and Liu, Gaohong and liu, juncai and Liu, LingJun and Liu, Xin and Lin, Haibin and Lin, Zhiqi and Ma, Bole and Sheng, Guangming and Tong, Yuxuan and Zhang, Chi and Zhang, Mofan and Zhang, Ru and Zhang, Wang and Zhu, Hang and Zhu, Jinhua and Chen, Jiaze and Chen, Jiangjie and Wang, Chengyi and Yu, Hongli and Song, Yuxuan and Wei, Xiangpeng and Zhou, Hao and Liu, Jingjing and Ma, Wei-Ying and Zhang, Ya-Qin and Yan, Lin and Wu, Yonghui and Wang, Mingxuan},
 booktitle = {Advances in Neural Information Processing Systems},
 doi = {10.52202/085713-3775},
 editor = {D. Belgrave and C. Zhang and H. Lin and R. Pascanu and P. Koniusz and M. Ghassemi and N. Chen},
 pages = {113222--113244},
 publisher = {Curran Associates, Inc.},
 title = {DAPO: An Open-Source LLM Reinforcement Learning System at Scale},
 url = {https://proceedings.neurips.cc/paper_files/paper/2025/file/a4277440d50f1f15d2cb4c14f7e0c0d2-Paper-Conference.pdf},
 volume = {38},
 year = {2025}
}

@inproceedings{
liu2025understanding,
title={Understanding R1-Zero-Like Training: A Critical Perspective},
author={Zichen Liu and Changyu Chen and Wenjun Li and Penghui Qi and Tianyu Pang and Chao Du and Wee Sun Lee and Min Lin},
booktitle={Second Conference on Language Modeling},
year={2025},
url={https://openreview.net/forum?id=5PAF7PAY2Y}
}

@inproceedings{
zhao2026large,
title={Large Language Model Agents Are Not Always Faithful Self-Evolvers},
author={Weixiang Zhao and Yingshuo Wang and Yichen Zhang and Yang Deng and Yanyan Zhao and Wanxiang Che and Bing Qin and Ting Liu},
booktitle={Forty-third International Conference on Machine Learning},
year={2026},
url={https://openreview.net/forum?id=kTjSSqgqGf}
}

@article{kirkpatrick2017overcoming,
author = {James Kirkpatrick  and Razvan Pascanu  and Neil Rabinowitz  and Joel Veness  and Guillaume Desjardins  and Andrei A. Rusu  and Kieran Milan  and John Quan  and Tiago Ramalho  and Agnieszka Grabska-Barwinska  and Demis Hassabis  and Claudia Clopath  and Dharshan Kumaran  and Raia Hadsell },
title = {Overcoming catastrophic forgetting in neural networks},
journal = {Proceedings of the National Academy of Sciences},
volume = {114},
number = {13},
pages = {3521-3526},
year = {2017},
doi = {10.1073/pnas.1611835114},
URL = {https://www.pnas.org/doi/abs/10.1073/pnas.1611835114},
eprint = {https://www.pnas.org/doi/pdf/10.1073/pnas.1611835114}}

@ARTICLE{luo2025empirical,
  author={Luo, Yun and Yang, Zhen and Meng, Fandong and Li, Yafu and Zhou, Jie and Zhang, Yue},
  journal={IEEE Transactions on Audio, Speech and Language Processing}, 
  title={An Empirical Study of Catastrophic Forgetting in Large Language Models During Continual Fine-Tuning}, 
  year={2025},
  volume={33},
  number={},
  pages={3776-3786},
  doi={10.1109/TASLPRO.2025.3606231},
  url={https://doi.org/10.1109/TASLPRO.2025.3606231}
}

@article{
biderman2024lora,
title={Lo{RA} Learns Less and Forgets Less},
author={Dan Biderman and Jacob Portes and Jose Javier Gonzalez Ortiz and Mansheej Paul and Philip Greengard and Connor Jennings and Daniel King and Sam Havens and Vitaliy Chiley and Jonathan Frankle and Cody Blakeney and John Patrick Cunningham},
journal={Transactions on Machine Learning Research},
issn={2835-8856},
year={2024},
url={https://openreview.net/forum?id=aloEru2qCG},
note={Featured Certification}
}

@article{shumailov2024collapse,
  title   = {AI models collapse when trained on recursively generated data},
  author  = {Shumailov, Ilia and Shumaylov, Zakhar and Zhao, Yiren and Papernot, Nicolas and Anderson, Ross and Gal, Yarin},
  journal = {Nature},
  volume  = {631},
  pages   = {755--759},
  year    = {2024},
  doi     = {10.1038/s41586-024-07566-y},
  url     = {https://www.nature.com/articles/s41586-024-07566-y}
}

@inproceedings{alemohammad2024selfconsuming,
 author = {Alemohammad, Sina and Casco-Rodriguez, Josue and Luzi, Lorenzo and Humayun, Ahmed Imtiaz and Babaei, Hossein and LeJeune, Daniel and Siahkoohi, Ali and Baraniuk, Richard},
 booktitle = {International Conference on Learning Representations},
 editor = {B. Kim and Y. Yue and S. Chaudhuri and K. Fragkiadaki and M. Khan and Y. Sun},
 pages = {53581--53608},
 title = {Self-Consuming Generative Models Go MAD},
 url = {https://proceedings.iclr.cc/paper_files/paper/2024/file/ebc042e767de551803ccfcc45e2454f5-Paper-Conference.pdf},
 volume = {2024},
 year = {2024}
}

@inproceedings{
gerstgrasser2024is,
title={Is Model Collapse Inevitable? Breaking the Curse of Recursion by Accumulating Real and Synthetic Data},
author={Matthias Gerstgrasser and Rylan Schaeffer and Apratim Dey and Rafael Rafailov and Tomasz Korbak and Henry Sleight and Rajashree Agrawal and John Hughes and Dhruv Bhandarkar Pai and Andrey Gromov and Dan Roberts and Diyi Yang and David L. Donoho and Sanmi Koyejo},
booktitle={First Conference on Language Modeling},
year={2024},
url={https://openreview.net/forum?id=5B2K4LRgmz}
}

@inproceedings{qi2024finetuning,
 author = {Qi, Xiangyu and Zeng, Yi and Xie, Tinghao and Chen, Pin-Yu and Jia, Ruoxi and Mittal, Prateek and Henderson, Peter},
 booktitle = {International Conference on Learning Representations},
 editor = {B. Kim and Y. Yue and S. Chaudhuri and K. Fragkiadaki and M. Khan and Y. Sun},
 pages = {30988--31043},
 title = {Fine-tuning Aligned Language Models Compromises Safety, Even When Users Do Not Intend To!},
 url = {https://proceedings.iclr.cc/paper_files/paper/2024/file/83b7da3ed13f06c13ce82235c8eedf35-Paper-Conference.pdf},
 volume = {2024},
 year = {2024}
}

@misc{miller2024adding,
      title={Adding Error Bars to Evals: A Statistical Approach to Language Model Evaluations}, 
      author={Evan Miller},
      year={2024},
      eprint={2411.00640},
      archivePrefix={arXiv},
      primaryClass={stat.AP},
      url={https://arxiv.org/abs/2411.00640}, 
}

@inproceedings{hariri2026dont,
 author = {Hariri, Mohsen and Samandar, Amirhossein and Hinczewski, Michael and Chaudhary, Vipin},
 booktitle = {International Conference on Learning Representations},
 editor = {C. Vondrick and B. Hariharan and C. Raffel and L. Pinto and D. Yang and A. Faust},
 pages = {148539--148579},
 title = {Don’t Pass@k: A Bayesian Framework for Large Language Model Evaluation},
 url = {https://proceedings.iclr.cc/paper_files/paper/2026/file/f04edfc65463d020629673a4bc4c58e7-Paper-Conference.pdf},
 volume = {2026},
 year = {2026}
}

@misc{dodge2020finetuning,
      title={Fine-Tuning Pretrained Language Models: Weight Initializations, Data Orders, and Early Stopping}, 
      author={Jesse Dodge and Gabriel Ilharco and Roy Schwartz and Ali Farhadi and Hannaneh Hajishirzi and Noah Smith},
      year={2020},
      eprint={2002.06305},
      archivePrefix={arXiv},
      primaryClass={cs.CL},
      url={https://arxiv.org/abs/2002.06305}, 
}

@inproceedings{zhang2024careful,
 author = {Zhang, Hugh and Da, Jeff and Lee, Dean and Robinson, Vaughn and Wu, Catherine and Song, Will and Zhao, Tiffany and Raja, Pranav and Zhuang, Charlotte and Slack, Dylan and Lyu, Qin and Hendryx, Sean and Kaplan, Russell and Lunati, Michele and Yue, Summer},
 booktitle = {Advances in Neural Information Processing Systems},
 doi = {10.52202/079017-1485},
 editor = {A. Globerson and L. Mackey and D. Belgrave and A. Fan and U. Paquet and J. Tomczak and C. Zhang},
 pages = {46819--46836},
 publisher = {Curran Associates, Inc.},
 title = {A Careful Examination of Large Language Model Performance on Grade School Arithmetic},
 url = {https://proceedings.neurips.cc/paper_files/paper/2024/file/53384f2090c6a5cac952c598fd67992f-Paper-Datasets_and_Benchmarks_Track.pdf},
 volume = {37},
 year = {2024}
}

@misc{cobbe2021training,
      title={Training Verifiers to Solve Math Word Problems}, 
      author={Karl Cobbe and Vineet Kosaraju and Mohammad Bavarian and Mark Chen and Heewoo Jun and Lukasz Kaiser and Matthias Plappert and Jerry Tworek and Jacob Hilton and Reiichiro Nakano and Christopher Hesse and John Schulman},
      year={2021},
      eprint={2110.14168},
      archivePrefix={arXiv},
      primaryClass={cs.LG},
      url={https://arxiv.org/abs/2110.14168}, 
}

@inproceedings{gema2025done,
    title = "Are We Done with {MMLU}?",
    author = "Gema, Aryo Pradipta  and
      Leang, Joshua Ong Jun  and
      Hong, Giwon  and
      Devoto, Alessio  and
      Mancino, Alberto Carlo Maria  and
      Saxena, Rohit  and
      He, Xuanli  and
      Zhao, Yu  and
      Du, Xiaotang  and
      Ghasemi Madani, Mohammad Reza  and
      Barale, Claire  and
      McHardy, Robert  and
      Harris, Joshua  and
      Kaddour, Jean  and
      Van Krieken, Emile  and
      Minervini, Pasquale",
    editor = "Chiruzzo, Luis  and
      Ritter, Alan  and
      Wang, Lu",
    booktitle = "Proceedings of the 2025 Conference of the Nations of the Americas Chapter of the Association for Computational Linguistics: Human Language Technologies (Volume 1: Long Papers)",
    month = apr,
    year = "2025",
    address = "Albuquerque, New Mexico",
    publisher = "Association for Computational Linguistics",
    url = "https://aclanthology.org/2025.naacl-long.262/",
    doi = "10.18653/v1/2025.naacl-long.262",
    pages = "5069--5096",
    ISBN = "979-8-89176-189-6"
}

@article{wilson1927probable,
author = {Edwin B. Wilson},
title = {Probable Inference, the Law of Succession, and Statistical Inference},
journal = {Journal of the American Statistical Association},
volume = {22},
number = {158},
pages = {209--212},
year = {1927},
publisher = {Taylor \& Francis},
doi = {10.1080/01621459.1927.10502953},
URL = {https://www.tandfonline.com/doi/abs/10.1080/01621459.1927.10502953},
eprint = {https://www.tandfonline.com/doi/pdf/10.1080/01621459.1927.10502953}
}

@article{brown2001interval,
author = {Lawrence D. Brown and T. Tony Cai and Anirban DasGupta},
title = {{Interval Estimation for a Binomial Proportion}},
volume = {16},
journal = {Statistical Science},
number = {2},
publisher = {Institute of Mathematical Statistics},
pages = {101 -- 133},
year = {2001},
doi = {10.1214/ss/1009213286},
URL = {https://doi.org/10.1214/ss/1009213286}
}

@article{efron1979bootstrap,
author = {B. Efron},
title = {{Bootstrap Methods: Another Look at the Jackknife}},
volume = {7},
journal = {The Annals of Statistics},
number = {1},
publisher = {Institute of Mathematical Statistics},
pages = {1 -- 26},
year = {1979},
doi = {10.1214/aos/1176344552},
URL = {https://doi.org/10.1214/aos/1176344552}
}

@article{fisher1922on,
 ISSN = {09528385},
 URL = {http://www.jstor.org/stable/2340521},
 author = {R. A. Fisher},
 journal = {Journal of the Royal Statistical Society},
 number = {1},
 pages = {87--94},
 publisher = {[Wiley, Royal Statistical Society]},
 title = {On the Interpretation of {$\chi^2$} from Contingency Tables, and the Calculation of P},
 urldate = {2026-08-10},
 volume = {85},
 year = {1922}
}

@InProceedings{sun2025learning,
  title = 	 {Learning to ({L}earn at Test Time): {RNN}s with Expressive Hidden States},
  author =       {Sun, Yu and Li, Xinhao and Dalal, Karan and Xu, Jiarui and Vikram, Arjun and Zhang, Genghan and Dubois, Yann and Chen, Xinlei and Wang, Xiaolong and Koyejo, Sanmi and Hashimoto, Tatsunori and Guestrin, Carlos},
  booktitle = 	 {Proceedings of the 42nd International Conference on Machine Learning},
  pages = 	 {57503--57522},
  year = 	 {2025},
  editor = 	 {Singh, Aarti and Fazel, Maryam and Hsu, Daniel and Lacoste-Julien, Simon and Berkenkamp, Felix and Maharaj, Tegan and Wagstaff, Kiri and Zhu, Jerry},
  volume = 	 {267},
  series = 	 {Proceedings of Machine Learning Research},
  month = 	 {13--19 Jul},
  publisher =    {PMLR},
  url = 	 {https://proceedings.mlr.press/v267/sun25h.html}
}

@InProceedings{gozeten2025testtime,
  title = 	 {Test-Time Training Provably Improves Transformers as In-context Learners},
  author =       {Gozeten, Halil Alperen and Ildiz, Muhammed Emrullah and Zhang, Xuechen and Soltanolkotabi, Mahdi and Mondelli, Marco and Oymak, Samet},
  booktitle = 	 {Proceedings of the 42nd International Conference on Machine Learning},
  pages = 	 {20266--20295},
  year = 	 {2025},
  editor = 	 {Singh, Aarti and Fazel, Maryam and Hsu, Daniel and Lacoste-Julien, Simon and Berkenkamp, Felix and Maharaj, Tegan and Wagstaff, Kiri and Zhu, Jerry},
  volume = 	 {267},
  series = 	 {Proceedings of Machine Learning Research},
  month = 	 {13--19 Jul},
  publisher =    {PMLR},
  url = 	 {https://proceedings.mlr.press/v267/gozeten25a.html}
}

@misc{snell2024scaling,
      title={Scaling LLM Test-Time Compute Optimally can be More Effective than Scaling Model Parameters}, 
      author={Charlie Snell and Jaehoon Lee and Kelvin Xu and Aviral Kumar},
      year={2024},
      eprint={2408.03314},
      archivePrefix={arXiv},
      primaryClass={cs.LG},
      url={https://arxiv.org/abs/2408.03314}, 
}

@inproceedings{muennighoff2025s1,
    title = "s1: Simple test-time scaling",
    author = "Muennighoff, Niklas  and
      Yang, Zitong  and
      Shi, Weijia  and
      Li, Xiang Lisa  and
      Fei-Fei, Li  and
      Hajishirzi, Hannaneh  and
      Zettlemoyer, Luke  and
      Liang, Percy  and
      Cand{\`e}s, Emmanuel  and
      Hashimoto, Tatsunori",
    editor = "Christodoulopoulos, Christos  and
      Chakraborty, Tanmoy  and
      Rose, Carolyn  and
      Peng, Violet",
    booktitle = "Proceedings of the 2025 Conference on Empirical Methods in Natural Language Processing",
    month = nov,
    year = "2025",
    address = "Suzhou, China",
    publisher = "Association for Computational Linguistics",
    url = "https://aclanthology.org/2025.emnlp-main.1025/",
    doi = "10.18653/v1/2025.emnlp-main.1025",
    pages = "20275--20321",
    ISBN = "979-8-89176-332-6"
}

@misc{hinton2015distilling,
      title={Distilling the Knowledge in a Neural Network}, 
      author={Geoffrey Hinton and Oriol Vinyals and Jeff Dean},
      year={2015},
      eprint={1503.02531},
      archivePrefix={arXiv},
      primaryClass={stat.ML},
      url={https://arxiv.org/abs/1503.02531}, 
}

@inproceedings{hsieh2023distilling,
    title = "Distilling Step-by-Step! Outperforming Larger Language Models with Less Training Data and Smaller Model Sizes",
    author = "Hsieh, Cheng-Yu  and
      Li, Chun-Liang  and
      Yeh, Chih-kuan  and
      Nakhost, Hootan  and
      Fujii, Yasuhisa  and
      Ratner, Alex  and
      Krishna, Ranjay  and
      Lee, Chen-Yu  and
      Pfister, Tomas",
    editor = "Rogers, Anna  and
      Boyd-Graber, Jordan  and
      Okazaki, Naoaki",
    booktitle = "Findings of the Association for Computational Linguistics: ACL 2023",
    month = jul,
    year = "2023",
    address = "Toronto, Canada",
    publisher = "Association for Computational Linguistics",
    url = "https://aclanthology.org/2023.findings-acl.507/",
    doi = "10.18653/v1/2023.findings-acl.507",
    pages = "8003--8017"
}

@misc{openai2025gptoss,
      title={gpt-oss-120b \& gpt-oss-20b Model Card}, 
      author={OpenAI and : and Sandhini Agarwal and Lama Ahmad and Jason Ai and Sam Altman and Andy Applebaum and Edwin Arbus and Rahul K. Arora and Yu Bai and Bowen Baker and Haiming Bao and Boaz Barak and Ally Bennett and Tyler Bertao and Nivedita Brett and Eugene Brevdo and Greg Brockman and Sebastien Bubeck and Che Chang and Kai Chen and Mark Chen and Enoch Cheung and Aidan Clark and Dan Cook and Marat Dukhan and Casey Dvorak and Kevin Fives and Vlad Fomenko and Timur Garipov and Kristian Georgiev and Mia Glaese and Tarun Gogineni and Adam Goucher and Lukas Gross and Katia Gil Guzman and John Hallman and Jackie Hehir and Johannes Heidecke and Alec Helyar and Haitang Hu and Romain Huet and Jacob Huh and Saachi Jain and Zach Johnson and Chris Koch and Irina Kofman and Dominik Kundel and Jason Kwon and Volodymyr Kyrylov and Elaine Ya Le and Guillaume Leclerc and James Park Lennon and Scott Lessans and Mario Lezcano-Casado and Yuanzhi Li and Zhuohan Li and Ji Lin and Jordan Liss and Lily and Liu and Jiancheng Liu and Kevin Lu and Chris Lu and Zoran Martinovic and Lindsay McCallum and Josh McGrath and Scott McKinney and Aidan McLaughlin and Song Mei and Steve Mostovoy and Tong Mu and Gideon Myles and Alexander Neitz and Alex Nichol and Jakub Pachocki and Alex Paino and Dana Palmie and Ashley Pantuliano and Giambattista Parascandolo and Jongsoo Park and Leher Pathak and Carolina Paz and Ludovic Peran and Dmitry Pimenov and Michelle Pokrass and Elizabeth Proehl and Huida Qiu and Gaby Raila and Filippo Raso and Hongyu Ren and Kimmy Richardson and David Robinson and Bob Rotsted and Hadi Salman and Suvansh Sanjeev and Max Schwarzer and D. Sculley and Harshit Sikchi and Kendal Simon and Karan Singhal and Yang Song and Dane Stuckey and Zhiqing Sun and Philippe Tillet and Sam Toizer and Foivos Tsimpourlas and Nikhil Vyas and Eric Wallace and Xin Wang and Miles Wang and Olivia Watkins and Kevin Weil and Amy Wendling and Kevin Whinnery and Cedric Whitney and Hannah Wong and Lin Yang and Yu Yang and Michihiro Yasunaga and Kristen Ying and Wojciech Zaremba and Wenting Zhan and Cyril Zhang and Brian Zhang and Eddie Zhang and Shengjia Zhao},
      year={2025},
      eprint={2508.10925},
      archivePrefix={arXiv},
      primaryClass={cs.CL},
      url={https://arxiv.org/abs/2508.10925}, 
}

@misc{yang2025qwen3,
      title={Qwen3 Technical Report}, 
      author={An Yang and Anfeng Li and Baosong Yang and Beichen Zhang and Binyuan Hui and Bo Zheng and Bowen Yu and Chang Gao and Chengen Huang and Chenxu Lv and Chujie Zheng and Dayiheng Liu and Fan Zhou and Fei Huang and Feng Hu and Hao Ge and Haoran Wei and Huan Lin and Jialong Tang and Jian Yang and Jianhong Tu and Jianwei Zhang and Jianxin Yang and Jiaxi Yang and Jing Zhou and Jingren Zhou and Junyang Lin and Kai Dang and Keqin Bao and Kexin Yang and Le Yu and Lianghao Deng and Mei Li and Mingfeng Xue and Mingze Li and Pei Zhang and Peng Wang and Qin Zhu and Rui Men and Ruize Gao and Shixuan Liu and Shuang Luo and Tianhao Li and Tianyi Tang and Wenbiao Yin and Xingzhang Ren and Xinyu Wang and Xinyu Zhang and Xuancheng Ren and Yang Fan and Yang Su and Yichang Zhang and Yinger Zhang and Yu Wan and Yuqiong Liu and Zekun Wang and Zeyu Cui and Zhenru Zhang and Zhipeng Zhou and Zihan Qiu},
      year={2025},
      eprint={2505.09388},
      archivePrefix={arXiv},
      primaryClass={cs.CL},
      url={https://arxiv.org/abs/2505.09388}, 
}

@misc{mathverify2025,
  title  = {Math-Verify: Math Verification Library},
  author = {Kydlíček, Hynek},
  year   = {2025},
  howpublished = {GitHub repository},
  url    = {https://github.com/huggingface/Math-Verify}
}

@misc{yu2026skilladaptor,
      title={SkillAdaptor: Self-Adapting Skills for LLM Agents from Trajectories}, 
      author={Zhuoyun Yu and Xin Xie and Wuguannan Yao and Chenxi Wang and Lei Liang and Xiang Qi and Shumin Deng},
      year={2026},
      eprint={2606.01311},
      archivePrefix={arXiv},
      primaryClass={cs.CL},
      url={https://arxiv.org/abs/2606.01311}, 
}

@misc{atreja2025alas,
      title={ALAS: Autonomous Learning Agent for Self-Updating Language Models}, 
      author={Dhruv Atreja},
      year={2025},
      eprint={2508.15805},
      archivePrefix={arXiv},
      primaryClass={cs.CL},
      url={https://arxiv.org/abs/2508.15805}, 
}

@misc{chen2025selfquestioning,
      title={Self-Questioning Language Models}, 
      author={Lili Chen and Mihir Prabhudesai and Katerina Fragkiadaki and Hao Liu and Deepak Pathak},
      year={2025},
      eprint={2508.03682},
      archivePrefix={arXiv},
      primaryClass={cs.LG},
      url={https://arxiv.org/abs/2508.03682}, 
}
\bibliographystyle{iclr2027_conference}

\newpage

\appendix

\section{Experimental protocols and implementation details}

\label{app:protocols}

This appendix fixes what every arm and every generation in the study was run under: the training and sampling configuration (\S\ref{app:protocol}), the prompt and the grading rules a completion passed through (\S\ref{app:grading}), the policy-gradient objective and its three seeds (\S\ref{app:rl}), and the compute the study consumed (\S\ref{app:compute}).

\subsection{Training and sampling configuration}

\label{app:protocol}

\paragraph{The prompt.} Every generation in this study, for evaluation and for the evolution stream alike, uses one template with no few-shot examples and no system message:

\begin{quote}\footnotesize\ttfamily
Solve the following math problem. Reason step by step, then give your final answer inside \textbackslash boxed\{\}.\\[4pt] Problem: \{problem\}\\[4pt] Solution:
\end{quote}

\noindent The problem string is passed through unmodified, including any Asymptote figure source, which is why some MATH-500 geometry items carry a diagram listing into the context. Holding the prompt fixed across arms and checkpoints matters more here than choosing a good one: a transition is a difference between two evaluations, so anything that changes between them and is not the model is a confound.

Table~\ref{tab:protocol} gives the configuration of every arm and the per-round statistics of its training signal. Three points bear on the fairness of the comparison.

\textbf{Training volume is not matched on AIME by the obvious construction.} On the difficulty band and MATH-500 the retained examples per round are within $7\%$ across arms. On AIME the native protocols do not match: applied to the $60$-problem evaluation set itself, as TTRL specifies, the majority-vote arm retains $50$/$51$/$50$ against distillation's $247$/$246$/$247$ drawn from a disjoint stream of $256$ problems per round, a factor of $4.9$. Training volume is therefore confounded with the filter for precisely the arms that carry the expansion result, which is what the AIME ladder of Section~\ref{sec:results} removes: run on one shared stream of $256$ problems per round, majority-vote self-training retains $246$--$254$ examples per round across three seeds against distillation's $246$--$249$, within $2\%$. STaR retains $205$--$220$, because ground-truth filtering rejects more than a vote taken over the model's own samples can: a self-consistent vote almost always produces a majority, whether or not it is right, and on this stream $70$--$81\%$ of those majorities are correct. That residual gap is a property of the filter rather than a design choice, it runs \emph{against} the arm whose null we report, and we state it rather than tune it away.

\textbf{The vote is a different instrument on a different stream.} Pseudo-label accuracy, the fraction of majority votes that were in fact correct and a diagnostic that never influences training, is $91$--$95\%$ on the difficulty band, $70$--$81\%$ on the hard training stream the matched ladder uses, and $26$--$30\%$ when the vote is taken on AIME itself. The unmatched on-set arm is not simply smaller; it operates in a supervision regime the others never enter. That gap is the subject of Appendix~\ref{app:falsified}.

\textbf{Filtering and reinforcement are not the same update.} The arms labeled SFT keep the generations that agree with their filter and discard the rest; the policy-gradient arm keeps all of them and centers the reward within each group, so disagreeing samples are pushed down rather than dropped. Group size, stream, volume and evaluation are otherwise identical. Its retained count is reported as rollouts rather than examples, and only groups with reward spread contribute: $1{,}976$, $1{,}920$ and $2{,}016$ rollouts drawn per round on the band, of which $76$, $56$ and $59$ groups carried any gradient.

\begin{table}[h]
\centering
\caption{Arm configuration and training-signal statistics. Evaluation is identical across arms within a benchmark: $k{=}128$, $T{=}0.8$, top-$p$ $0.95$, plus one greedy sample, with independent sampling calls. All arms use Qwen3-8B with rank-32 LoRA, $\eta = 10^{-4}$, batch $8$, $3$ rounds of $256$ stream problems, and at most one training example per problem; SFT arms train $3$ epochs per round and the policy-gradient arm one, since its rollouts are off-policy after a single pass. The AIME block above the rule is the matched ladder; the arm below it is the unmatched baseline that trains on the evaluation set itself.}
\label{tab:protocol}
\scriptsize
\setlength{\tabcolsep}{7pt}
\begin{tabular}{llccclr}
\toprule
\textbf{arm} & \textbf{benchmark} & \textbf{gen.}\ $T$ & \textbf{max tokens} & \textbf{samples/prob} & \textbf{retained per round}
& \textbf{signal correct} \\
\midrule
STaR              & MATH-500 & 0.9 & 2{,}048 & 4 & 229 / 227 / 231 & $100\%$ (ground truth) \\
STaR              & band     & 0.9 & 2{,}048 & 4 & 232 / 229 / 230 & $100\%$ (ground truth) \\
maj.-vote SFT     & MATH-500 & 1.0 & 2{,}048 & 8 votes & 241 / 247 / 246
                  & $92.1 / 89.5 / 91.5\%$ \\
maj.-vote SFT     & band     & 1.0 & 2{,}048 & 8 votes & 247 / 246 / 240
                  & $92.7 / 91.1 / 94.6\%$ \\
Distillation      & band     & 0.7 & 2{,}048 & 4 & 232 / 239 / 234 & $100\%$ (filtered) \\
\midrule
\multicolumn{7}{l}{\emph{matched AIME ladder: one stream (\texttt{math-train-hard}), one volume, three seeds}} \\
maj.-vote SFT     & AIME & 1.0 & 8{,}192 & 8 votes & 250 / 250 / 246
                  & $78.0 / 77.6 / 80.5\%$ \\
STaR              & AIME & 0.9 & 8{,}192 & 4 & 220 / 205 / 205 & $100\%$ (ground truth) \\
Distillation      & AIME & 0.7 & 8{,}192 & 4 & 247 / 246 / 247 & $100\%$ (filtered) \\
maj.-vote RL      & AIME & 1.0 & 8{,}192 & 8 rollouts & $\dagger$ & $\dagger$ \\
maj.-vote RL      & band (175) & 1.0 & 2{,}048 & 8 rollouts & $\dagger$ & $\dagger$ \\
\midrule
\multicolumn{7}{l}{\emph{unmatched baseline (on-set training): unmatched in stream and volume, the reference for Section~\ref{sec:f2}}} \\
maj.-vote SFT     & AIME (on-set) & 1.0 & 8{,}192 & 8 votes & 50 / 51 / 50
                  & $26.0 / 27.5 / 30.0\%$ \\
\bottomrule
\end{tabular}

\vspace{0.3em}
{\small $\dagger$ final-round figures are filled from the released \texttt{audit.json} of each run, except the two policy-gradient seeds that were stopped after collapsing, whose per-round figures come from their logs (Table~\ref{tab:rlseeds}). The vote is $70$--$81\%$ correct on the hard training stream against $26$--$30\%$ when taken on AIME itself: the unmatched baseline differs from the ladder in supervision quality as well as in volume.}
\end{table}

\subsection{Prompting, grading and answer extraction}

\label{app:grading}

Answers are extracted from \verb|\boxed{}| where present and from an explicit ``the answer is'' construction otherwise, then compared by symbolic equivalence \citep{mathverify2025}. We deliberately reject a third fallback that is common in practice: taking the last number appearing in the completion. On truncated reasoning that heuristic is close to a coin flip, and its errors are \emph{random} rather than systematic. Random correctness is far more damaging to a transition metric than uniform strictness: a systematically strict grader lowers accuracy at every checkpoint and largely cancels in the difference, whereas a lottery moves problems between states for no reason and is counted as learning or corruption. On one pilot run the fallback supplied $78\%$ of all extracted answers.

Truncation interacts with this directly, which is F3. A completion that hits the token cap has no final answer and is graded incorrect, harmless if the rate is stable across checkpoints and severe if it is not (Table~\ref{tab:trunc}). We also track which extraction path produced each answer, since a method that stops emitting \verb|\boxed{}| would otherwise register as capability loss.

\begin{table}[h]
\centering
\caption{Truncation rate by checkpoint on the difficulty band, the diagnostic that excludes an arm's corruption counts (F3). A method that changes generation style will be scored as destroying capability under a fixed token budget, and nothing else in the ledger distinguishes the two.}
\label{tab:trunc}
\small
\begin{tabular}{lrrrrl}
\toprule
\textbf{arm} & $t{=}0$ & $t{=}1$ & $t{=}2$ & $t{=}3$ & \textbf{verdict} \\
\midrule
STaR         & $16.5\%$ & $15.9\%$ & $12.4\%$ & $14.7\%$ & stable; counts stand \\
maj.-vote SFT & $16.6\%$ & $14.6\%$ & $16.9\%$ & $16.0\%$ & stable; counts stand \\
Distillation & $16.5\%$ & $\mathbf{65.3\%}$ & $\mathbf{46.1\%}$ & $\mathbf{49.6\%}$
             & drift; corruption counts excluded \\
\bottomrule
\end{tabular}
\end{table}

\subsection{Policy gradient on a majority-vote reward}

\label{app:rl}

\paragraph{Why the arm exists.} Both self-training arms of the main audit are rejection-sampling SFT: they keep what their filter accepts and discard the rest, where TTRL as published pushes the losing samples down instead. An audit that positions itself against a reinforcement-learning literature and then measures only supervised fine-tuning is auditing something else, so we ran the genuine article: a group-relative policy gradient on the same stream and the same majority vote, under the training service's importance-sampling loss with group-centered advantages.

\paragraph{On the band it behaves; on AIME it does not survive.} The band subsample gives $19$ learned against $9$ corrupted ($\clr = 0.47$) with accuracy rising $0.530 \to 0.583$ and truncation flat at $15.5$, $24.4$, $7.6$ and $8.2\%$ across its four checkpoints. The AIME arm, on a harder stream at a four times larger token cap, collapses. Table~\ref{tab:rlseeds} gives all three of its seeds.

\begin{table}[h]
\centering
\caption{\textbf{Three seeds of the policy-gradient arm, on identical data.} The evolution stream is drawn under a fixed seed, so the three runs see the same $256$ problems in the same order each round; \texttt{-{}-seed} varies adapter initialization, training order and sampling. Round~$1$ is the same experiment three times over: the vote is $77.0$--$77.9\%$ correct and $135$--$136$ groups of ${\sim}250$ carry any gradient. The runs then diverge completely. Every seed ends in severe truncation, so the collapse belongs to the configuration; the round at which it arrives does not. \emph{Trained} is the number of problems that formed a majority, of $256$; \emph{spread} the groups carrying gradient; \emph{vote} the fraction of those majorities that were in fact correct, a diagnostic that never enters training.}
\label{tab:rlseeds}
\footnotesize
\setlength{\tabcolsep}{5pt}
\begin{tabular}{llrrrrrr}
\toprule
& & \multicolumn{2}{c}{\textbf{evaluation, AIME}} & \multicolumn{4}{c}{\textbf{training stream}} \\
\cmidrule(lr){3-4}\cmidrule(lr){5-8}
\textbf{seed} & \textbf{round} & \textbf{accuracy} & \textbf{truncated} & \textbf{trained} & \textbf{spread} & \textbf{share} & \textbf{vote} \\
\midrule
$\mathbf{0}$ & $t=0$ & $0.1764$ & 4.2\% & --- & --- & --- & --- \\
 & $t=1$ & $0.2082$ & 18.3\% & $248$ & $135$ & $0.78$ & $77.0\%$ \\
 & $t=2$ & $0.0643$ & 24.3\% & $251$ & $143$ & $0.77$ & $73.7\%$ \\
 & $t=3$ & $0.1921$ & $\mathbf{76.4}$\% & $243$ & $217$ & $0.58$ & $49.8\%$ \\
\midrule
$\mathbf{1}$ & $t=0$ & $0.1760$ & 4.4\% & --- & --- & --- & --- \\
 & $t=1$ & $0.0000$ & $\mathbf{100.0}$\% & $249$ & $135$ & $0.78$ & $77.5\%$ \\
 & $t=2$ & --- & --- & $34$ & $32$ & $0.51$ & $85.3\%$ \\
\midrule
$\mathbf{2}$ & $t=0$ & $0.1829$ & 4.5\% & --- & --- & --- & --- \\
 & $t=1$ & $0.2952$ & 47.1\% & $253$ & $136$ & $0.78$ & $77.9\%$ \\
 & $t=2$ & $0.0900$ & $\mathbf{95.6}$\% & $248$ & $105$ & $0.86$ & $85.1\%$ \\
\bottomrule
\end{tabular}

\vspace{0.3em}
{\small Seeds $1$ and $2$ were stopped after the collapse rather than sampled to $t{=}3$, since every further generation runs to the token cap. Seed $1$'s $t{=}2$ vote statistics are computed over the $34$ majorities that survived and are a survivorship artifact; seed $2$'s $t{=}2$ are computed over $248$ and are not. Training loss at $t{=}1$ was $169.9$, $8.5$ and $145.1$ across the three seeds, a twentyfold spread from adapter initialization alone.}
\end{table}

\paragraph{What three seeds show that one could not.} The evolution stream is drawn under a fixed seed, so the three runs see the same $256$ problems in the same order every round; \texttt{-{}-seed} varies adapter initialization, training order and sampling. Round one is therefore the same experiment three times, and it comes out the same three times: the vote is $77.0$, $77.5$ and $77.9\%$ correct and $135$, $135$ and $136$ groups of roughly $250$ carry any gradient. The runs then diverge to the point of one dying outright. Seed~$1$ truncates $100\%$ of its completions after a single update and scores $0.0000$; seed~$2$ reaches $95.6\%$ after two; seed~$0$ reaches $76.4\%$ after three. Two readings follow, and they point in opposite directions, which is why we give both. That every seed ends in severe truncation makes the collapse a property of this configuration rather than of one unlucky run; the answer to whether F3's drift was reproducible is yes. That the round at which it arrives varies from one update to three, on identical data, makes the arm's outcome undetermined by its design, which is F5 in a more extreme form than the supervised arms show. The training loss at round one was $169.9$, $8.5$ and $145.1$ across the three seeds: a twentyfold spread from adapter initialization alone.

\paragraph{The diagnostic reads healthy while the model dies.} A group contributes gradient exactly when its rollouts disagree, so the arm's supervision statistics are a live readout of how self-consistent the model is. On seed~$0$ they degrade as the model does. The vote falls $77.0 \to 73.7 \to 49.8\%$ and the share of rollouts agreeing with it $0.78 \to 0.58$, while the groups carrying gradient rise $135 \to 143 \to 217$: by round three the pseudo-label is close to a coin flip and the optimizer receives more gradient than ever, precisely because the model has stopped agreeing with itself. That is a real measurement ($243$ of $256$ problems still formed a majority), but it is one seed at one round, and it does not replicate. On seed~$2$ the same statistics move the other way at round two, over an equally intact denominator of $248$: the vote becomes \emph{more} accurate ($77.9 \to 85.1\%$) and \emph{more} unanimous (share $0.78 \to 0.86$), and fewer groups carry gradient ($136 \to 105$), while the evaluation goes from $47\%$ truncated to $96\%$. Seed~$1$'s round-two figures ($85.3\%$ correct) are a survivorship artifact and we mark them as such: only $34$ of $256$ problems formed a majority at all, because a degenerate generation yields no extractable answer to vote on. What survives across the seeds is not a direction but a warning: on two of three runs the arm's own supervision statistics looked stable or improving at the checkpoint after which the model was unusable, so they cannot be used to decide when to stop.

\paragraph{The scope of the claim, and the exclusion criterion.} The AIME arm's transition counts are excluded from the capability analysis. They are read off models truncating $76$ to $100\%$ of their completions, and F3 establishes that a token cap under generation-length drift scores style change as capability loss; the criterion applies to this arm exactly as it does to distillation. The detection counts of Table~\ref{tab:ladder} are retained because truncation moves them the other way: a truncated completion cannot be scored correct, so a collapsing arm \emph{understates} its detections and the count is a floor. We claim, for this configuration only, that three rounds of policy gradient on a majority-vote reward at rank-$32$ LoRA and $\eta = 10^{-4}$ collapsed the policy into unbounded repetition on every seed we ran. We do not claim that a majority-vote reward does this in general: seed~$2$ took two intact updates at the identical learning rate, so the rate is not uniformly fatal, and the band arm at a shorter cap on an easier stream never collapses at all. Separating the reward from the step size needs a learning-rate sweep we did not run.

\subsection{Compute footprint and cost}

\label{app:compute}

All experiments used Tinker, a hosted LoRA training and sampling service. Unless stated otherwise the backbone is Qwen3-8B with rank-32 adapters; the distillation teacher is \texttt{gpt-oss-120b} \citep{openai2025gptoss}, and one noise floor was measured on a second $20$B backbone. The study consumed $3.44$M sampled generations and $5.30$B generated tokens across $48$ runs, itemized in Table~\ref{tab:compute}.

\begin{table}[h]
\centering
\caption{Compute by experiment class. Cost is the metered price of sampling and training on the hosted service; token counts are generated tokens, which dominate cost.}
\label{tab:compute}
\small
\begin{tabular}{lrrrr}
\toprule
\textbf{experiment class} & \textbf{runs} & \textbf{samples (M)} & \textbf{tokens (M)} & \textbf{cost (USD)} \\
\midrule
Primary audits, difficulty band ($n{=}1{,}163$) & 3 & 1.80 & 2{,}441 & 1{,}465 \\
Primary audits, MATH-500 and AIME & 4 & 0.27 & 414 & 248 \\
Matched AIME ladder, 3 filters $\times$ seeds & 6 & 0.19 & 692 & 415 \\
Policy-gradient arms & 4 & 0.16 & 443 & 266 \\
Seed replication (2 methods $\times$ 2 seeds) & 4 & 0.36 & 470 & 282 \\
\midrule
Noise floors (4 benchmarks) & 7 & 0.42 & 589 & 340 \\
Benchmark construction and model profiling & 7 & 0.11 & 101 & 61 \\
Out-of-domain probes (3 replicates each) & 6 & 0.01 & 4 & 2 \\
Batch-invariance test (F1 mechanism) & 1 & $<0.01$ & 1 & $<1$ \\
\midrule
Pilot audits, 112-problem band & 2 & 0.12 & 147 & 88 \\
Method pilots and calibration & 4 & $<0.01$ & 4 & 2 \\
\midrule
\textbf{Total} & \textbf{48} & \textbf{3.44} & \textbf{5{,}303} & \textbf{3{,}171} \\
\bottomrule
\end{tabular}
\end{table}

Three ratios bear on the paper's recommendation. The measurement infrastructure (noise floors, benchmark construction and the out-of-domain probes) accounts for $13\%$ of total spend, and the small pilots that exposed four of the seven failure modes cost \$2 between them. Against that, the three primary audits alone cost \$$1{,}465$. And the controls the paper's conclusions rest on cost \emph{nothing}: the empirical null distributions, the design-matched floors, the threshold-free expansion test and the seed analysis are all re-computations over records the study already holds, because every arm's baseline is another evaluation of the untrained model. What money buys is the matched ladder (\$$415$), which removes a confound rather than measuring something new, and the policy-gradient arms (\$$266$), which audit a method family the SFT arms cannot stand in for and surface a failure mode none of the others can exhibit.

The dominant cost is sampling rather than training: evaluating $1{,}163$ problems at $k{=}128$ over four checkpoints is $600$K generations per arm, against roughly $270$ optimizer steps. This is a property of transition-level auditing generally: statistical power comes from problems near the decision boundary, and reaching them requires sampling breadth rather than longer training. An audit budget is therefore set almost entirely by the evaluation design, and can be estimated in advance from the benchmark size, $k$, the checkpoint count and the mean completion length, all known before any method is run.

\section{Benchmark suites and dataset construction}

\label{app:benchmarks}

Corruption and expansion are observable on disjoint populations of problems, so no single evaluation set can host both. This appendix gives the construction of the set built to make corruption observable (\S\ref{app:band}) and the characterization of all three sets against the regimes they can and cannot measure (\S\ref{app:benchsets}).

\subsection{Constructing the difficulty band}

\label{app:band}

Corruption is observable only on problems a model already solves, and expansion only on problems it does not reach. MATH-500 supplies the first population and almost none of the second; AIME the reverse (Figure~\ref{fig:pops}). Neither can host a well-powered corruption measurement. We therefore construct a third set: profile $10{,}999$ held-out MATH training problems at $k{=}8$ and keep the $1{,}611$ whose measured solve rate falls in $[0.25, 0.75]$, and retain the $1{,}163$ of them closest to the center of that interval. The retained set has a mean solve rate of $0.522$ under the $k{=}8$ selection profile and $0.552$ when re-measured at $k{=}128$, where it holds $711$ corruptible and $452$ learnable problems, the first balanced pools in the study.

\begin{figure}[h]
\centering
\includegraphics[width=\textwidth]{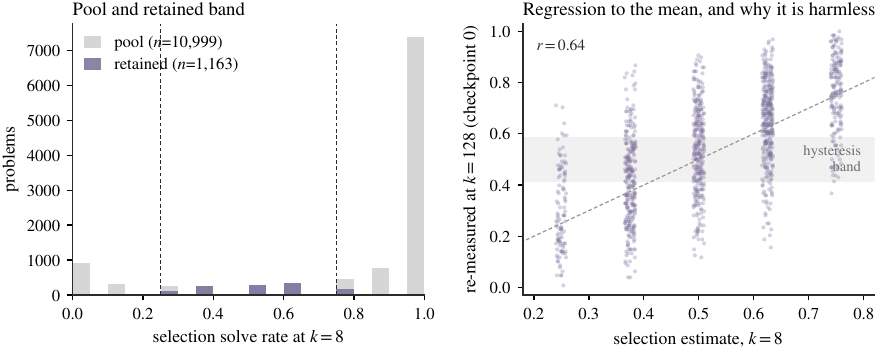}
\caption{\textbf{Construction of the difficulty band.} \emph{Left:} the held-out pool and the retained band; dashed lines mark the selection interval $[0.25, 0.75]$. \emph{Right:} selection estimate at $k{=}8$ against the independent re-measurement at $k{=}128$ that serves as checkpoint $0$. Regression to the mean widens the realized band, costing statistical power; because the two measurements are independent it cannot create transitions.}
\label{fig:band}
\end{figure}

Selecting problems by measured solve rate and then measuring them again invites an obvious objection: regression to the mean. Figure~\ref{fig:band} shows it plainly. This does not bias the transition counts, for a specific reason: selection uses an independent $k{=}8$ profile, the audit re-evaluates every problem from scratch at $k{=}128$, and \emph{that} measurement is checkpoint $0$, the baseline against which all transitions are computed. The noisy estimate used to choose never enters the ledger. What regression to the mean costs is realized band width: fewer problems sit near threshold at $k{=}128$ than the selection targeted, which reduces power but cannot manufacture a transition.

The band's construction removes every problem the base model cannot reach, so it has no expansion candidates and its expansion statistic is vacuous. This is the price of its corruption sensitivity. It also means the band's raw counts are not a forgetting rate for a naturally sampled problem distribution: the set is deliberately enriched for problems that can transition, and Table~\ref{tab:effects} rather than the raw count is the quantity that transfers.

\subsection{Benchmark characterization and solvability regimes}

\label{app:benchsets}

\label{app:falsified}

\begin{figure}[h]
\centering
\includegraphics[width=\textwidth]{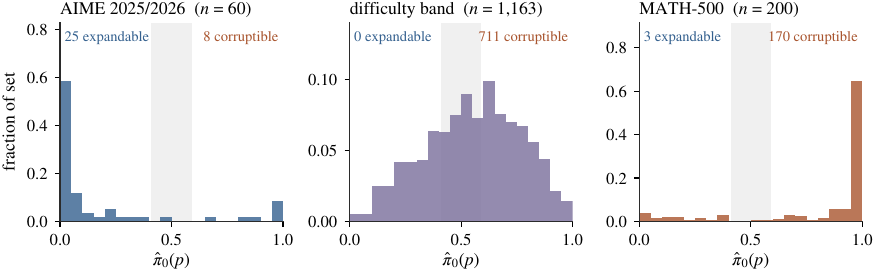}
\caption{\textbf{An evaluation set fixes which question it can answer, before any method is applied.} Measured distribution of the untrained backbone's per-problem solve rate at $k{=}128$, from the frozen-control records; shading marks the hysteresis band. Expansion is observable only in the spike at $\hat\pi_0 = 0$ and corruption only in the mass above $0.5$, so AIME can host an expansion measurement and not a corruption one, MATH-500 the reverse, and the constructed band neither expansion nor a natural forgetting rate.}
\label{fig:pops}
\end{figure}

\paragraph{Benchmarks, in full.} Corruption is observable only on problems the model already solves, expansion only on problems it does not reach. These populations are disjoint, and a benchmark rich in one is poor in the other (Figure~\ref{fig:pops}), so we use three sets. \textbf{MATH-500} \citep{lightman2024verify}, subsampled to $200$ problems by a fixed stratified draw over (subject, level) released with the code, gives comparability with prior work but is thoroughly contaminated ($168$--$170$ of $200$ problems already solved at $k{=}128$), hosting corruption but not expansion. \textbf{AIME 2025 and 2026} ($60$ problems) are the least contaminated sets available to us: AIME 2026 post-dates the backbone's March 2025 cutoff \citep{yang2025qwen3} outright, while AIME 2025 was administered in February 2025 and sits just inside it, so we report the $2026$ half separately wherever the expansion claim rests on it. The base model's mean solve rate on AIME is $0.175$--$0.180$ across its eleven baselines, leaving $22$ problems it reaches at most five times in $1{,}408$ pooled draws and only $8$ that can be corrupted, so AIME carries the expansion claims and has no corruption power. A \textbf{difficulty band} of $1{,}163$ problems, built by profiling $10{,}999$ held-out MATH problems \citep{hendrycks2021measuring} at $k{=}8$ and retaining those with solve rate in $[0.25, 0.75]$, is deliberately enriched for problems near the decision boundary so that corruption is observable at all. Its counts are therefore not a forgetting rate for a natural problem distribution, and by construction it contains no base-unreached problems (Appendix~\ref{app:band}). Because the fraction of problems near threshold differs across sets, $\clr$ is \emph{not} portable between them.

\paragraph{A falsified prediction about where corruption appears.} We pre-registered the prediction that majority-vote self-training would corrupt more where its pseudo-labels were less reliable, and it was falsified. On AIME its label accuracy falls to $26$--$30\%$, so the vote is wrong most of the time (Table~\ref{tab:protocol}), and corruption goes to \emph{zero}. The explanation is that AIME leaves only $8$ problems the model solves and could therefore lose; the smallest non-zero corruption rate observable there is $12.5\%$, against a rate of $1.2\%$ measured on MATH-500.

Wrong supervision on problems a model already fails destroys nothing. Corruption is bounded by what there is to lose, not by how wrong the teacher is, and an evaluation set can be too hard to detect it as easily as it can be too easy. We report the falsification because the account that replaces it, that corruption requires both unreliable supervision \emph{and} prior competence on the affected problems, is what motivates constructing the difficulty band, and because it is the reason the AIME arms carry expansion claims but not corruption ones.

\section{The transition ledger and its noise floor}

\label{app:ledger}

The ledger of Section~\ref{sec:audit} turns per-problem solve rates into transition counts. This appendix gives the rule in full and the counts it produces (\S\ref{app:counts}), the floors and the sensitivity of both to $z$ and $k$ (\S\ref{app:sensitivity}), and the three failure modes that attack the ledger itself (\S\ref{app:failures}).

\subsection{The baseline state, hysteresis, and the full ledger counts}

\label{app:counts}

\label{app:s0}

\textbf{The baseline state carries no hysteresis, and this matters.} Equation~\ref{eq:band} is undefined at $t=0$, where there is no previous state to hold; we assign $s_0(p) = \textsf{solved}$ iff $\hat\pi_0(p) \geq \tfrac12$. The consequence is load-bearing and easy to miss. A problem whose baseline sits inside the band is unprotected at that one checkpoint, so a frozen model can register a transition on a movement smaller than the band is wide. That is why the floors of Table~\ref{tab:floors} are not zero on a set whose mean solve rate is $0.552$. Section~\ref{sec:results} reports one alternative, a pure two-checkpoint rule with no carry. The other is symmetric: treat a baseline inside the band as \emph{indeterminate} and let a problem transition only once it has left the band. That rule drives the frozen floor to exactly $0$ learned and $0$ corrupted, with no artifact surviving it, while STaR gives $64/53$ and majority-vote SFT $75/55$. It buys that zero by setting aside $325$ of the $1{,}163$ problems, which is why we do not adopt it as the primary rule: a floor of zero over a set chosen for having no determinate starting state is a weaker achievement than a small floor over the whole set. The conclusion is unchanged under it, the arms now clearing the floor absolutely rather than by a ratio.

\begin{table}[t]
\centering
\caption{\textbf{What each method adds and removes, against a floor measured on the same problems and matched to the same number of checkpoints.} Counts are problems. The floor columns give the median over four-checkpoint orderings of the independent evaluations of the untrained model that benchmark carries, with the conventional two-checkpoint floor in parentheses.}
\label{tab:main}
\footnotesize
\setlength{\tabcolsep}{5pt}
\begin{tabular}{llrrccl}
\toprule
& & \multicolumn{2}{c}{\textbf{transitions}} & \multicolumn{2}{c}{\textbf{matched floor}} & \\
\cmidrule(lr){3-4}\cmidrule(lr){5-6}
\textbf{method} & \textbf{benchmark} & \textbf{learned} & \textbf{corrupted} & \textbf{L} & \textbf{C} & \textbf{accuracy} \\
\midrule
STaR                     & MATH-500 & 2 & 1 & 0 (0) & 1 (1) & $0.835 \to 0.850$ \\
maj.-vote SFT            & MATH-500 & 3 & 2 & 0 (0) & 1 (1) & $0.832 \to 0.830$ \\
maj.-vote SFT            & AIME     & 4 & 0 & 0 (0) & 0 (0) & $0.177 \to 0.202$ \\
Distillation$^{\dagger}$ & AIME     & 9 & 0 & 0 (0) & 0 (0) & $0.180 \to 0.310$ \\
\midrule
STaR                     & band & 149 & 106 & 10 (4) & 8 (5) & $0.550 \to 0.606$ \\
maj.-vote SFT            & band & 145 &  88 & 10 (4) & 8 (5) & $0.550 \to 0.590$ \\
Distillation$^{\dagger\ddagger}$ & band & 161 & \emph{177} & 10 (4) & 8 (5) & $0.550 \to 0.546$ \\
\bottomrule
\end{tabular}

\vspace{0.35em}
{\small $^{\dagger}$positive control. \quad $^{\ddagger}$corruption counts excluded: truncation drifts $16.5\% \to 65.3\%$ (F3). \quad The AIME rows here are the unmatched baselines (on-set training, unmatched volume); the matched ladder is Table~\ref{tab:ladder}. \par\vspace{0.15em} Sharpened share of newly solved problems on the sets where expansion is observable at all: $2/2$, $3/3$, $4/4$, so $9/9$, Wilson $[0.70, 1.00]$. The band arms ($149/149$, $145/145$, $161/161$) are vacuous by construction and are excluded from that proportion rather than added to it. \quad $\clr$ with percentile-bootstrap intervals over problems, band: STaR $0.711\,[0.557, 0.915]$, maj.-vote SFT $0.607\,[0.460, 0.785]$; seed variance is roughly three times this (Table~\ref{tab:seeds}).}
\end{table}

Table~\ref{tab:counts} gives the complete seven-category partition for every arm. The categories are mutually exclusive and sum to the problem count. \emph{Recovered} and \emph{transient} problems, those that change state and change back, are reported separately because they are invisible to an endpoint comparison but bear on whether a method's damage is permanent. Their scarcity (at most $5\%$ of problems in any arm) is why we report the endpoint definition of $\clr$ in the main text: on this data the two definitions coincide to within three problems. Movement is bidirectional in every arm, the band producing $88$--$175$ corruptions alongside $144$--$160$ learnings, so a rising mean accuracy is a net result and not a uniform lift.

\begin{figure}[h]
\centering
\includegraphics[width=\textwidth]{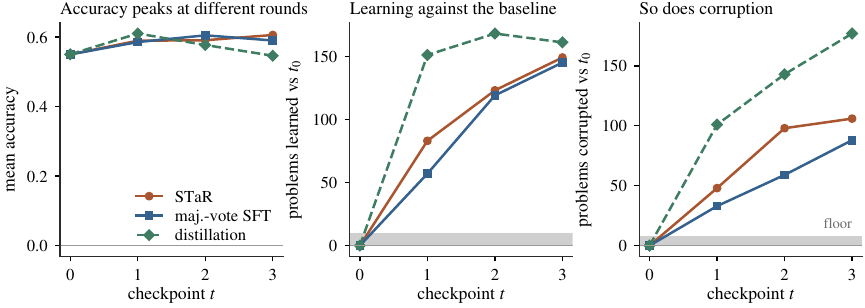}
\caption{\textbf{The path the endpoint comparison hides.} Mean accuracy and the learned and corrupted counts against checkpoint~$0$ at every checkpoint on the difficulty band, with the frozen-model floor shaded on the two count panels. Distillation peaks at round~$1$ and declines under its own truncation drift; majority-vote self-training's accuracy peaks at round~$2$ while its corruption count continues to climb.}
\label{fig:trajectories}
\end{figure}

\paragraph{The path to the endpoint.} Table~\ref{tab:trajectory} and Figure~\ref{fig:trajectories} give the same ledger evaluated at every checkpoint rather than only at the last. Two features are invisible in the endpoint comparison. Distillation on the band peaks at round~$1$ ($0.610$) and then falls to below its own baseline by round~$3$ ($0.546$), which is the truncation drift of F3 taking hold as the student's completions lengthen. Majority-vote self-training peaks at round~$2$ ($0.605$) and gives back $1.5$ points at round~$3$ while its corruption count rises from $59$ to $88$; a study that stopped at two rounds would have reported a better method. Neither observation changes a headline, and we report the endpoint comparison because it is the one prior work reports, but the trajectory is where a practitioner would choose when to stop.

\begin{figure}[h]
\centering
\includegraphics[width=\textwidth]{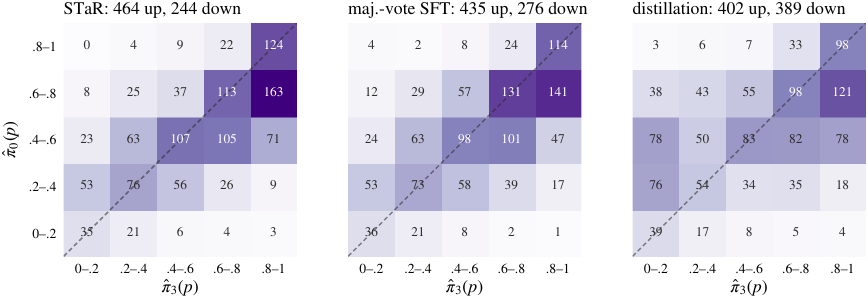}
\caption{\textbf{Migration matrices: where problems actually move.} Joint distribution of the per-problem solve rate before and after evolution, binned into fifths, for the three difficulty-band arms; the dashed line is no change and cell counts sum to $1{,}163$. ``Up'' and ``down'' count problems above and below the diagonal. Movement is substantial in both directions for every arm, which is the point that a mean accuracy cannot express.}
\label{fig:migration}
\end{figure}

\paragraph{Where the mass goes.} Figure~\ref{fig:migration} bins the solve rate into five bands and tabulates the joint distribution of before and after. The diagonal holds; the off-diagonal mass is large in both directions for every arm: $464$ problems up and $244$ down for STaR, $435$ and $276$ for majority-vote self-training, $402$ and $389$ for distillation. The bottom-left cell is the one that matters for the expansion claim, and it is where the band's construction bites: with no problems at $\hat\pi_0 = 0$ there is nothing in that cell to move. Appendix~\ref{app:shift} shows the same movement without binning.

\begin{table}[h]
\centering
\caption{\textbf{Per-checkpoint accuracy and transition counts.} Every headline number in the paper compares $t{=}0$ with $t{=}3$; this is the path between them. Each count compares that checkpoint with checkpoint $0$ under the ledger of Eq.~\ref{eq:band}; it is a running state, not a cumulative total, so it can fall when a problem re-crosses the band, and the final column of each block reproduces Table~\ref{tab:main}. Distillation on the band peaks at round $1$ and declines under its own truncation drift (F3); majority-vote SFT on the band peaks at round $2$ in accuracy while its corruption count keeps rising. The AIME policy-gradient row is reported for completeness only: that arm collapses into unbounded repetition and is $76.4\%$ truncated at $t{=}3$, so its transition counts are excluded from the capability analysis (Appendix~\ref{app:rl}).}
\label{tab:trajectory}
\footnotesize
\setlength{\tabcolsep}{4pt}
\begin{tabular}{llrrrrrrrrrrrr}
\toprule
& & \multicolumn{4}{c}{mean accuracy} & \multicolumn{4}{c}{learned vs $t_0$} & \multicolumn{4}{c}{corrupted vs $t_0$} \\
\cmidrule(lr){3-6}\cmidrule(lr){7-10}\cmidrule(lr){11-14}
arm & set & $t_0$ & $t_1$ & $t_2$ & $t_3$ & $t_0$ & $t_1$ & $t_2$ & $t_3$ & $t_0$ & $t_1$ & $t_2$ & $t_3$ \\
\midrule
STaR & band & 0.550 & 0.590 & 0.591 & 0.606 & 0 & 83 & 123 & 149 & 0 & 48 & 98 & 106 \\
maj.-vote SFT & band & 0.550 & 0.586 & 0.605 & 0.590 & 0 & 57 & 119 & 145 & 0 & 33 & 59 & 88 \\
Distillation$^\dagger$ & band & 0.550 & 0.610 & 0.577 & 0.546 & 0 & 151 & 168 & 161 & 0 & 101 & 143 & 177 \\
\midrule
maj.-vote RL & band (175) & 0.530 & 0.603 & 0.578 & 0.583 & 0 & 22 & 21 & 19 & 0 & 13 & 12 & 9 \\
STaR & MATH-500 & 0.835 & 0.841 & 0.851 & 0.850 & 0 & 1 & 4 & 2 & 0 & 1 & 1 & 1 \\
maj.-vote SFT & MATH-500 & 0.832 & 0.836 & 0.830 & 0.830 & 0 & 1 & 2 & 3 & 0 & 0 & 0 & 2 \\
maj.-vote SFT & AIME & 0.180 & 0.193 & 0.227 & 0.213 & 0 & 1 & 1 & 1 & 0 & 0 & 0 & 0 \\
STaR & AIME & 0.177 & 0.190 & 0.204 & 0.195 & 0 & 0 & 0 & 1 & 0 & 0 & 0 & 0 \\
Distillation$^\dagger$ & AIME & 0.180 & 0.347 & 0.308 & 0.310 & 0 & 12 & 11 & 9 & 0 & 0 & 0 & 0 \\
maj.-vote RL$^{\ddagger}$ & AIME & 0.176 & 0.208 & 0.064 & 0.192 & 0 & 3 & 0 & 4 & 0 & 0 & 6 & 2 \\
\bottomrule
\end{tabular}

\vspace{0.3em}
{\small $^{\dagger}$positive control, not a self-evolution method. $^{\ddagger}$counts excluded: generation collapse (Table~\ref{tab:rlseeds}).}
\end{table}

\begin{table}[h]
\centering
\caption{Ledger categories by arm, sampled ledger at $k{=}128$, $z{=}2$. \emph{Learned} and \emph{corrupted} here are the trajectory-based labels; the endpoint counts used for $\clr$ in the main text differ only where a problem oscillates.}
\label{tab:counts}
\footnotesize
\resizebox{1\textwidth}{!}{%
\begin{tabular}{llrrrrrrr}
\toprule
& & \multicolumn{2}{c}{\textbf{stable}} & \multicolumn{5}{c}{\textbf{changed}} \\
\cmidrule(lr){3-4}\cmidrule(lr){5-9}
\textbf{arm} & \textbf{benchmark} & \textbf{correct} & \textbf{incorrect} & \textbf{learned} & \textbf{corrupted} & \textbf{recovered} & \textbf{transient}
& \textbf{oscillating} \\
\midrule
STaR & MATH-500 & 167 & 25 & 2 & 1 & 2 & 3 & 0 \\
maj.-vote SFT & MATH-500 & 167 & 28 & 3 & 2 & 0 & 0 & 0 \\
maj.-vote SFT & AIME & 8 & 48 & 4 & 0 & 0 & 0 & 0 \\
Distillation$^{\dagger}$ & AIME & 8 & 40 & 9 & 0 & 0 & 3 & 0 \\
\midrule
STaR & band & 567 & 299 & 149 & 103 & 25 & 17 & 3 \\
maj.-vote SFT & band & 589 & 299 & 144 & 88 & 19 & 23 & 1 \\
Distillation$^{\dagger\ddagger}$ & band & 532 & 248 & 160 & 175 & 11 & 34 & 3 \\
\bottomrule
\end{tabular}
}
\vspace{0.3em}
{\small $^{\dagger}$positive control. \quad $^{\ddagger}$corruption counts excluded: truncation drift (F3).}
\end{table}

\subsection{Noise floors, and sensitivity to \texorpdfstring{$z$}{z} and \texorpdfstring{$k$}{k}}

\label{app:sensitivity}

\begin{table}[h]
\centering
\caption{\textbf{Noise floors: apparent transitions produced by an untrained model} (F1, F7). Estimator columns give the range over every ordered pair of independent frozen evaluations ($110$ pairs on AIME, $12$ on MATH-500 and $20$ on the band) and, in the last estimator column, the median over four-checkpoint orderings that are design-matched to an arm. A conventional single-pair floor is the low end of these ranges. The single-greedy protocol is what a per-example before/after comparison implicitly uses; its changed rate carries a Wilson interval since it is a proportion over problems.}
\label{tab:floors}
\scriptsize
\setlength{\tabcolsep}{6pt}
\begin{tabular}{llccc cr r}
\toprule
& & \multicolumn{3}{c}{\textbf{solve-rate estimator, $k{=}128$}}
  & \multicolumn{2}{c}{\textbf{single greedy sample}} & \\
\cmidrule(lr){3-5}\cmidrule(lr){6-7}
\textbf{benchmark} & \textbf{backbone} & \textbf{learned} & \textbf{corrupted}
& \textbf{matched (L/C)} & \textbf{L/C} & \textbf{changed [95\%]}
& \textbf{problems} \\
\midrule
MATH-500        & Qwen3-8B    & 0--1 & 0--1 & 0 / 1 & 6 / 9 & $7.5\%\,[4.6, 12.0]$ & 200 \\
AIME 2025/26    & Qwen3-8B    & 0--0 & 0--1 & 0 / 0 & 2 / 2 & $6.7\%\,[2.6, 15.9]$ & 60 \\
difficulty band & Qwen3-8B    & 0--7 & 1--9 & \textbf{10 / 8} & 100 / 120
                & $\mathbf{18.9\%}\,[16.8, 21.3]$ & 1{,}163 \\
band (pilot)    & Qwen3-8B    & 0 & 0 & --- & --- & --- & 112 \\
AIME 2025/26    & gpt-oss-20b & 0 & 0 & --- & 5 / 4 & $15.0\%\,[8.1, 26.1]$ & 60 \\
\bottomrule
\end{tabular}

\vspace{0.3em}
{\small The gpt-oss-20b row is the only measurement we report on a second backbone; its arms were not run because its capability profile places it in a different regime, leaving only $7$ base-unreached AIME problems. Pilot and second-backbone rows have a single frozen pair each and so have no range.}
\end{table}

Table~\ref{tab:greedyflips} names every MATH-500 problem whose greedy verdict changed between two evaluations of the frozen model, with the solve rate measured on the same samples. The point is the disagreement between the two: \texttt{num\_theory/598} is scored as corrupted while its solve rate is $0.898$ at both checkpoints, and \texttt{int\_algebra/1063} is scored as \emph{learned} while its solve rate \emph{falls} from $0.133$ to $0.023$. The single-decode verdict is not a noisy measurement of capability; on these problems it is close to independent of it. Nine of the fifteen are scored as corruptions and six as learnings, which is the $\textsc{clr} = 1.5$ of Section~\ref{sec:f1} in full.

\begin{table}[h]
\centering
\caption{\textbf{Every problem whose greedy verdict changed between two evaluations of the frozen model} (F1), MATH-500, $T{=}0$ with a fixed seed. $\checkmark$ and $\times$ are the greedy verdicts; $\hat\pi$ is the solve rate over $128$ samples at the same checkpoint. The point of the table is the last two columns of each block: the underlying solve rate barely moves, since it is a frozen model, while the single-decode verdict flips. Nine of these are scored as corruptions and six as learnings, giving the $\textsc{clr} = 1.5$ of Section~\ref{sec:f1}.}
\label{tab:greedyflips}
\footnotesize
\setlength{\tabcolsep}{3.4pt}
\begin{tabular}{lccrr@{\hskip 10pt}lccrr}
\toprule
problem & $t_0$ & $t_1$ & $\hat\pi_0$ & $\hat\pi_1$ & problem & $t_0$ & $t_1$ & $\hat\pi_0$ & $\hat\pi_1$ \\
\midrule
\texttt{counting\_prob/199} & $\times$ & $\checkmark$ & 0.633 & 0.672 & \texttt{num\_theory/183} & $\times$ & $\checkmark$ & 0.875 & 0.852 \\
\texttt{geometry/226} & $\checkmark$ & $\times$ & 0.719 & 0.750 & \texttt{num\_theory/357} & $\checkmark$ & $\times$ & 0.594 & 0.453 \\
\texttt{geometry/702} & $\times$ & $\checkmark$ & 0.078 & 0.117 & \texttt{num\_theory/516} & $\checkmark$ & $\times$ & 0.398 & 0.328 \\
\texttt{int\_algebra/1063} & $\times$ & $\checkmark$ & 0.133 & 0.023 & \texttt{num\_theory/598} & $\checkmark$ & $\times$ & 0.898 & 0.898 \\
\texttt{int\_algebra/1350} & $\checkmark$ & $\times$ & 0.367 & 0.367 & \texttt{prealgebra/1128} & $\checkmark$ & $\times$ & 0.688 & 0.758 \\
\texttt{int\_algebra/1797} & $\times$ & $\checkmark$ & 0.859 & 0.844 & \texttt{precalculus/1313} & $\checkmark$ & $\times$ & 0.602 & 0.695 \\
\texttt{int\_algebra/754} & $\times$ & $\checkmark$ & 0.547 & 0.367 & \texttt{precalculus/44} & $\checkmark$ & $\times$ & 0.922 & 0.953 \\
\texttt{int\_algebra/966} & $\checkmark$ & $\times$ & 0.391 & 0.344 & & & & &  \\
\bottomrule
\end{tabular}

\vspace{0.3em}
{\small Under the solve-rate estimator of Eq.~\ref{eq:band}, the same $200$ problems yield $0$ learnings and $1$ corruption.}
\end{table}

Table~\ref{tab:sensitivity} sweeps the hysteresis width $z$ and the sampling budget $k$ on the difficulty band. The arms' ratios are flat across both sweeps while the frozen-model floor is not, which is what the hysteresis band is for. Smaller $k$ is simulated by subsampling the stored $128$ draws without replacement, so the comparison is exact rather than a fresh sample.

\begin{table}[h]
\centering
\caption{Sensitivity of the ledger to $z$ and $k$: learned/corrupted counts with $\clr$ in parentheses, difficulty band.}
\label{tab:sensitivity}
\small
\setlength{\tabcolsep}{4.5pt}
\begin{tabular}{lccc c lccc}
\toprule
\multicolumn{4}{c}{\textbf{hysteresis width $z$ (at $k{=}128$)}} & &
\multicolumn{4}{c}{\textbf{sampling budget $k$ (at $z{=}2$)}} \\
\cmidrule(lr){1-4}\cmidrule(lr){6-9}
$z$ & \textbf{STaR} & \textbf{maj.-vote} & \textbf{frozen} & & $k$ & \textbf{STaR} & \textbf{maj.-vote} & \textbf{frozen} \\
\midrule
$0.0$ & 172/111 (0.65) & 167/117 (0.70) & 44/39 & & $16$ & 115/64 (0.56) & 125/69 (0.55)
      & 10/7 \\
$1.0$ & 165/111 (0.67) & 158/103 (0.65) & 12/22 & & $32$ & 103/83 (0.81) & 120/88 (0.73)
      & 6/8 \\
$1.5$ & 160/109 (0.68) & 151/97  (0.64) & 7/12  & & $64$ & 134/95 (0.71) & 143/79 (0.55)
      & 5/7 \\
$2.0$ & \textbf{149/106 (0.71)} & \textbf{145/88 (0.61)} & \textbf{4/5} & &
        $128$ & \textbf{149/106 (0.71)} & \textbf{145/88 (0.61)} & \textbf{4/5} \\
$2.5$ & 136/98  (0.72) & 131/85  (0.65) & 1/3   & & & & & \\
$3.0$ & 125/89  (0.71) & 121/76  (0.63) & 0/2   & & & & & \\
\bottomrule
\end{tabular}
\end{table}

\paragraph{What the measured floor buys, and what the rule does.} A binomial calculation at each problem's pooled base rate, the analytic unchanged-policy null of \citet{yuan2026diversity}, predicts a median of $3$ learned and $4$ corrupted on the band against our measured $4$ and $3$, and an envelope of $0.203$ against $0.234$. For the solve-rate estimator the analytic null is very nearly right, and we say so rather than claim an advantage we cannot demonstrate; where the two part company is the single-decode protocol, for which an analytic null predicts no transitions at all. Discretization choices likewise move the counts without moving the conclusion: under a pure two-checkpoint rule with no carry, STaR's counts fall from $149/106$ to $120/79$ and majority-vote SFT's from $145/88$ to $118/78$ while the floor stays at $4/5$; dropping the band entirely raises the arms to $172/111$ and $167/117$ but the floor to $44/39$. $\clr$ stays within $0.65$--$0.71$ for STaR across all three rules, and within $0.65$--$0.72$ over $z \in [0,3]$ and $0.56$--$0.81$ over $k \in \{16,32,64,128\}$ (Table~\ref{tab:sensitivity}). One caution bears on the main claim: under the pure endpoint rule the two methods coincide exactly ($0.66$ and $0.66$), and at $z = 0$ their order reverses, so the between-method comparison is sensitive to the rule as well as to the seed and we rest nothing on it.

\subsection{Failure modes F3, F4 and F5 in full}

\label{app:failures}

\paragraph{F3. A token cap scores style change as capability loss.} The distillation arm on the band records $177$ corruptions and $\clr = 1.10$, nominally the most destructive method tested. Its truncation rate meanwhile moves from $16.5\%$ to $65.3\%$ between checkpoints while both self-training arms stay flat at $12$--$17\%$: the student has adopted its $120$B teacher's verbose style and is hitting the token cap, losing the final answer. Conditioned on completions that finish, it improves from $0.656$ to $0.845$ against STaR's $0.655$ to $0.702$: the most effective method tested, not the most destructive. Its corruption counts are excluded from the capability analysis rather than caveated, since severe generation-length drift under a fixed token cap confounds capability degradation with stylistic verbosity (Appendix~\ref{app:grading}).

\paragraph{F4. Transition metrics need an order of magnitude more data than accuracy.} Only problems near the decision boundary can produce an event, so a benchmark's difficulty profile determines power directly. Detecting the difference in corrupted share we observe between STaR and majority-vote self-training at $80\%$ power requires roughly $249$ events per arm; a $112$-problem pilot yields $24$ and cannot resolve it at any significance level, which is why the primary comparison uses $1{,}163$ problems (Appendix~\ref{app:power}).

\begin{figure}[h]
\centering
\includegraphics[width=\textwidth]{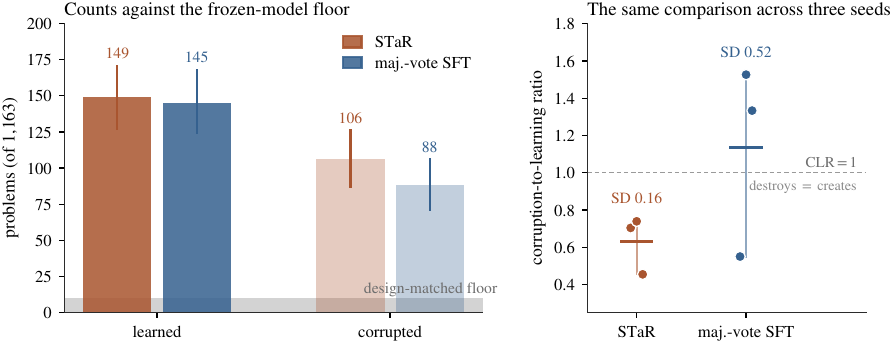}
\caption{\textbf{Transitions clear the noise floor by ${\sim}11\times$, but the comparison between methods is not stable across seeds.} \emph{Left:} learned and corrupted counts on the $1{,}163$-problem difficulty band with bootstrap intervals over problems; the shaded region is the frozen-model floor, design-matched to the arms' four checkpoints rather than the conventional two-checkpoint control, which understates it by roughly half. \emph{Right:} $\clr$ across three training seeds, every seed scored on the same $175$ problems (F5). STaR holds $0.45$--$0.74$; majority-vote self-training spans $0.55$ to $1.53$, crossing the threshold at which a method destroys more than it creates.}
\label{fig:transitions}
\end{figure}

\paragraph{F5. The comparison between methods is not stable across training seeds.} Table~\ref{tab:seeds} gives all three seeds of both band arms with every seed scored on the same $175$ problems, and Figure~\ref{fig:transitions} plots the resulting spread against the noise floor. Seed $0$ ran on the full band and seeds $1$--$2$ on a $175$-problem subsample, so the matched columns restrict seed $0$ to those same $175$ problems: the reversal is a seed effect and not a set effect. STaR holds $\clr$ within $0.45$--$0.74$ on every seed, while majority-vote self-training spans $0.55$ to $1.53$ and crosses the threshold at which a method destroys more than it creates. Section~\ref{sec:f7} states what that licenses and what it does not.

\begin{table}[h]
\centering
\caption{\textbf{Seed replication (F5), with every seed scored on the same problems.} Seed $0$ ran on the full band and seeds $1$--$2$ on a $175$-problem subsample, so an as-run comparison confounds seed with evaluation set. Restricting seed $0$ to the same $175$ problems costs no additional sampling and is given in the matched columns: the reversal is a seed effect, not a set effect.}
\label{tab:seeds}
\scriptsize
\setlength{\tabcolsep}{3pt}
\begin{tabular}{llrrrrrl}
\toprule
\textbf{method} & \textbf{seed} & $n$ (matched) & \textbf{learned} & \textbf{corrupted} & $\clr$ \textbf{matched} & $\clr$ \textbf{as run} & \\
\midrule
\multirow{3}{*}{STaR}
 & 0 & 175 & 27 & 19 & $0.704$ & $0.711$ & \multirow{3}{*}{\ range $0.45$--$0.74$; net
   positive} \\
 & 1 & 175 & 23 & 17 & $0.739$ & $0.739$ & \\
 & 2 & 175 & 22 & 10 & $0.455$ & $0.455$ & \\
\midrule
\multirow{3}{*}{maj.-vote}
 & 0 & 175 & 20 & 11 & $0.550$ & $0.607$ & \multirow{3}{*}{\ range $0.55$--$1.53$; $2/3$ net
   destructive} \\
 & 1 & 175 & 19 & 29 & $\mathbf{1.526}$ & $1.526$ & \\
 & 2 & 175 & 18 & 24 & $\mathbf{1.333}$ & $1.333$ & \\
\midrule
\multicolumn{8}{l}{\scriptsize seed $0$, full band: corrupted share $0.416$ vs $0.378$,
Fisher $p = 0.406$. Pooling seeds as a fixed effect is not defensible with three of them,} \\
\multicolumn{8}{l}{\scriptsize so we report the spread rather than a pooled $p$-value. $\Delta$ accuracy by seed:} \\
\multicolumn{8}{l}{\scriptsize STaR $+0.056 / {+}0.056 / {+}0.069$; maj.-vote $+0.040 / {-}0.028 / {-}0.010$.} \\
\bottomrule
\end{tabular}
\end{table}

\begin{figure}[h]
\centering
\includegraphics[width=\textwidth]{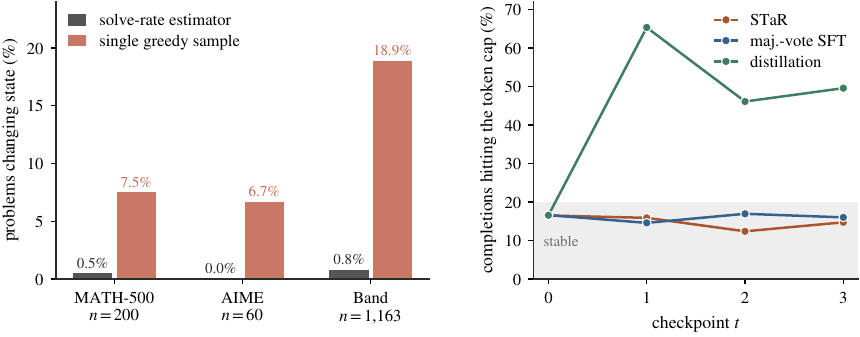}
\caption{\textbf{Two of the seven failures, on the runs where they occurred.} \emph{Left:} apparent transitions produced by a model that never trained, under two definitions of correctness (F1). A single greedy sample manufactures state changes on $6.7$--$18.9\%$ of problems; the solve-rate estimator with a hysteresis band reports $0.0$--$0.8\%$. \emph{Right:} truncation rate per checkpoint (F3). The distillation student adopted its teacher's verbose style and began hitting the token cap, losing final answers and registering as capability loss; the self-evolution arms are flat, which is why their counts survive.}
\label{fig:measure}
\end{figure}

\section{Expansion statistics, power and out-of-domain probes}

\label{app:expansion}

The expansion half of the audit rests on a per-problem exact test against a pooled baseline. This appendix varies every choice that test leaves open (\S\ref{app:testrobust}), itemizes the problems it is computed on (\S\ref{app:candidates}), gives the power behind its null result and the effect sizes behind the corruption counts (\S\ref{app:power}), and reports the two out-of-domain probes (\S\ref{app:collateral}).

\subsection{Robustness of the threshold-free exact test}

\label{app:testrobust}

The test of Section~\ref{sec:f2} replaces one explicit parameter with four implicit ones: the error rate, the one-sidedness, the number of baselines pooled and the base-rate strata. It also rests on an assumption, that the eleven baseline evaluations are exchangeable. This subsection varies each and checks the assumption. Nothing here required a new experiment.

\paragraph{Does the decision rule make the comparison?} A false-discovery-rate procedure is adaptive: it admits a larger per-hypothesis cutoff in an arm containing many true effects, so in principle an arm with a large global uplift can purchase a more permissive standard for its own marginal problems, and a signal-free control can be held to a stricter one. The concern is well founded in general and does not apply here, for a reason that is checkable rather than arguable: the cutoff a procedure \emph{admits} is $k\alpha/M$, but the cutoff it \emph{attains} is the $k$-th smallest $p$-value, and here those are $0.0095$ for distillation and $0.0094$ for majority-vote self-training. The two arms are held to the same standard to three decimal places. Table~\ref{tab:decisionrules} repeats the comparison under Bonferroni, at fixed thresholds, and under a single procedure applied jointly across every arm and problem at once; the low-base counts move by at most three and the ordering never changes. The one qualification is that the frozen control's zero is a property of a corrected rule: at an uncorrected $p \leq 0.05$ it returns $13$ detections over $660$ tests, which is the nominal rate and not a failure.

\paragraph{What the interaction shows.} The arms of the ladder are matched in stream, retained volume and evaluation, but not in how much they improve: the teacher moves accuracy $+13$ to $+16$ points and majority-vote self-training $-1.5$ to $+3.3$. A model in which every method lifts every problem by a common factor in log-odds would reproduce more detections for the teacher everywhere, and disproportionately more where the base rate is low, without any qualitative difference in kind. Table~\ref{tab:interaction} tests that model and rejects it: the teacher moves low-base problems by a further $1.91$ in log-odds beyond its own overall lift, where both self-training arms are indistinguishable from zero. The dissociation is a difference in what the methods reach, not only in how far they move.

\paragraph{An analytic null, and what measuring buys.} Appendix~\ref{app:sensitivity} reports that for the solve-rate ledger an analytic unchanged-policy null very nearly reproduces our measured floor, which invites the question of whether the expansion null could have been calculated rather than measured. It can be, and Table~\ref{tab:analyticer} does so. The calculation is optimistic at every threshold and increasingly so as $m$ rises, because the base-unreached set is selected on a noisy zero. A problem observed at $0/1{,}408$ has a plug-in success probability of exactly zero and is therefore assigned no chance of a later success, when in truth it has a small one. At $m{=}2$ the calculation returns $0.044$ against a measured $0.058$. An author who derived the null instead of measuring it would certify the threshold that F7 rules out.

\paragraph{How many replicates a null needs.} Our recommendation, that a multi-arm study already owns several frozen replicates, is only useful with a number attached to it. Table~\ref{tab:replicates} re-estimates the $\er_2$ null from every subset of our eleven baselines, exhaustively, since there are never more than $462$ of them. At four replicates, which is what a three-arm study with a no-op control possesses, the estimate still ranges from $0.022$ to $0.098$: such a study could certify a null of $0.022$ and reproduce F7 at a slightly larger sample size. The spread falls to $0.043$ at seven replicates and $0.024$ at nine, and it is nine before the estimate holds within $\pm 0.02$ of the eleven-replicate value. The recommendation should therefore be read as \emph{report the interval and the replicate count it came from}, and a null one intends to certify against needs rather more than a three-arm study owns.

\paragraph{Are the eleven baselines exchangeable?} They were run at different times against a hosted service whose batching we ourselves show to be non-deterministic (F1), so pooling them assumes something that has to be checked: that no run-level effect inflates the arm-versus-pool comparison while leaving the held-out null clean. Fitting the eleven per-problem counts against their pooled rate gives a Pearson dispersion of $0.966$ on $500$ degrees of freedom (a value of $1$ is exactly binomial), with a median per-problem ratio of $0.907$. The implied intraclass correlation is indistinguishable from zero and the design effect at $k{=}128$ is $1.000$. The draws are exchangeable across evaluations, which is what the exact test assumes and what the pooling requires.

\paragraph{Zero is an estimate too.} The held-out null returns no detection on any of the eleven replicates, which is a count and not a rate. At the replicate level the Clopper--Pearson bound is $0.24$; over the $660$ individual tests it is $0.0045$. We report both rather than repeat, one level up, the failure F7 describes.

\paragraph{The small non-zero counts, and the teacher's handicap.} We state the small non-zero counts rather than round them away. They are what an unmatched comparison hides: applied to a stream of problems it can actually solve, self-training does occasionally raise a rarely-reached problem above the noise. It does so at a fifth of the teacher's rate; the direction is unchanged, and the quantity that transfers is a rate rather than a categorical ``never''. The comparison is also conservative in the teacher's disfavor for a reason F3 predicts: the distillation arms truncate $36$--$46\%$ of completions even at an $8{,}192$-token cap, rising $4.2\%$ to $53.4\%$ across checkpoints on seed $0$, against $4$--$15\%$ for the self-training arms. F3 excludes an arm's \emph{corruption} counts because truncation inflates them; here it works the other way, since a truncated completion cannot be scored correct and so cannot produce a detection. The teacher's counts are a floor on what it did, and the dissociation is understated rather than manufactured. Sensitivity is arithmetic rather than something to be bought: against a base of $0/1{,}408$ an endpoint of $2/128$ already clears raw alpha, which is also the smallest count that survived FDR in the distillation arm.

\begin{table}[h]
\centering
\caption{\textbf{The comparison does not depend on the decision rule.} Low-base detections (pooled base rate $\leq 5/1{,}408$, $n = 22$) under four rules applied identically to every arm, and under one joint procedure over all arms and problems at once. A false-discovery-rate procedure is adaptive, so an arm with many true effects admits a larger cutoff; the cutoffs actually attained here are $0.0095$ for distillation's $36$ detections and $0.0094$ for majority-vote self-training's $19$, so the two arms are held to the same standard. The frozen row is the leave-one-out null over all eleven replicates.}
\label{tab:decisionrules}
\footnotesize
\setlength{\tabcolsep}{5pt}
\begin{tabular}{lccccc}
\toprule
\textbf{arm} & \textbf{BH, per arm} & \textbf{Bonferroni} & \textbf{$p \leq 0.01$} & \textbf{$p \leq 0.05$} & \textbf{joint BH} \\
\midrule
distillation, seed 0 & 10 & 7 & 10 & 10 & 9 \\
distillation, seed 1 & 11 & 6 & 10 & 11 & --- \\
distillation, seed 2 & 8 & 7 & 7 & 8 & --- \\
maj.-vote SFT, seed 0 & 2 & 0 & 2 & 2 & 0 \\
maj.-vote SFT, seed 1 & 1 & 1 & 1 & 1 & --- \\
maj.-vote SFT, seed 2 & 0 & 0 & 0 & 0 & --- \\
STaR & 2 & 2 & 2 & 2 & 2 \\
policy gradient & 1 & 0 & 1 & 1 & 1 \\
\midrule
\textbf{frozen}, 11 replicates & $\mathbf{0}$ & $\mathbf{0}$ & $\mathbf{0}$ & $\mathbf{0}$--$\mathbf{1}$ & $\mathbf{0}$ \\
\bottomrule
\end{tabular}

\vspace{0.3em}
{\small Joint BH is over $900$ hypotheses (four arms and eleven frozen replicates $\times\ 60$ problems), attained cutoff $0.00248$. At an uncorrected $p \leq 0.05$ the frozen control is no longer at zero: it returns $13$ detections over $660$ tests, close to the nominal rate.}
\end{table}

\begin{table}[h]
\centering
\caption{\textbf{The dissociation is not a by-product of the teacher's larger gain.} Logistic model on the per-problem counts, with terms for arm, base-rate stratum, before/after, and their interactions; the coefficient shown is the additional log-odds an arm moves a low-base problem \emph{beyond} its own overall lift. A model in which every method improves problems by a common multiplicative factor predicts zero here. It is zero for both self-training arms and is not for the teacher.}
\label{tab:interaction}
\footnotesize
\begin{tabular}{lccc}
\toprule
\textbf{term} & $\beta$ & \textbf{95\% CI} & $p$ \\
\midrule
low-base $\times$ post (reference arm, STaR) & $+1.17$ & $[+0.52, +1.82]$ & $10^{-3}$ \\
\textbf{distillation} $\times$ low-base $\times$ post & $+1.91$ & $[+1.25, +2.56]$ & $10^{-8}$ \\
majority-vote SFT $\times$ low-base $\times$ post & $+0.21$ & $[-0.60, +1.03]$ & $0.61$ \\
policy gradient $\times$ low-base $\times$ post & $-0.49$ & $[-1.51, +0.53]$ & $0.34$ \\
\bottomrule
\end{tabular}
\end{table}

\begin{table}[h]
\centering
\caption{\textbf{Sensitivity of the threshold-free test to the choices that remain.} Low-base detections ($n = 22$) as the number of pooled baselines and the error rate vary. Pooling fewer baselines costs sensitivity in the arm with signal and never manufactures it in the arms without.}
\label{tab:testsensitivity}
\footnotesize
\setlength{\tabcolsep}{5pt}
\begin{tabular}{lccccccc}
\toprule
& \multicolumn{4}{c}{\textbf{baselines pooled}} & \multicolumn{3}{c}{\textbf{FDR level}} \\
\cmidrule(lr){2-5}\cmidrule(lr){6-8}
\textbf{arm} & $3$ & $5$ & $7$ & $11$ & $0.01$ & $0.05$ & $0.10$ \\
\midrule
distillation, seed 0 & 8 & 10 & 10 & 10 & 9 & 10 & 10 \\
maj.-vote SFT, seed 0 & 0 & 1 & 1 & 2 & 0 & 2 & 2 \\
STaR & 1 & 2 & 2 & 2 & 2 & 2 & 2 \\
policy gradient & 1 & 1 & 1 & 1 & 0 & 1 & 1 \\
\bottomrule
\end{tabular}

\vspace{0.3em}
{\small Draws per problem at each pool size: $384$, $640$, $896$, $1{,}408$.}
\end{table}

\begin{table}[h]
\centering
\caption{\textbf{An analytic null for the expansion rate, and why measuring it still matters.} $\mathbb{E}[\er_m] = \mathbb{E}_q[(1-q)^k \Pr(X \geq m \mid k,q)] / \mathbb{E}_q[(1-q)^k]$, with the distribution of $q$ taken from the pooled $1{,}408$-draw baseline, as a plug-in and under an empirical-Bayes beta prior. The calculation is systematically optimistic and increasingly so as $m$ rises, because the base-unreached set is selected on a noisy zero: a problem observed at $0/1{,}408$ has a plug-in $q$ of exactly zero and is assigned no chance of a later success. At the threshold that appears to repair the statistic, an author trusting the calculation would certify $0.044$ and reach the same false conclusion.}
\label{tab:analyticer}
\footnotesize
\begin{tabular}{cccc}
\toprule
$m$ & \textbf{measured} & \textbf{analytic, plug-in} & \textbf{analytic, beta-binomial} \\
\midrule
$1$ & $0.176$ & $0.158$ & $0.163$ \\
$2$ & $\mathbf{0.058}$ & $\mathbf{0.044}$ & $0.049$ \\
$3$ & $0.023$ & $0.013$ & $0.015$ \\
$5$ & $0.003$ & $0.001$ & $0.002$ \\
\bottomrule
\end{tabular}
\end{table}

\begin{table}[h]
\centering
\caption{\textbf{How many frozen replicates a null needs before it settles.} The pooled $\er_2$ null re-estimated from \emph{every} subset of $G$ of our eleven baseline evaluations, exhaustively rather than sampled, since $\binom{11}{G}$ never exceeds $462$. Four replicates, which is what a three-arm study with a no-op control possesses for free, still admit estimates from $0.022$ to $0.098$, so such a study could certify a null of $0.022$ and reproduce F7. The spread halves by seven and again by nine; it takes nine before the estimate stays within $\pm 0.02$ of the $0.058$ that eleven give.}
\label{tab:replicates}
\footnotesize
\begin{tabular}{ccccccc}
\toprule
\textbf{replicates} $G$ & \textbf{subsets} & \textbf{ordered pairs} & \textbf{min} & \textbf{median} & \textbf{max} & \textbf{spread} \\
\midrule
$2$ & $55$ & $2$ & $0.000$ & $0.048$ & $0.120$ & $0.120$ \\
$3$ & $165$ & $6$ & $0.015$ & $0.056$ & $0.103$ & $0.088$ \\
$\mathbf{4}$ & $330$ & $12$ & $0.022$ & $0.058$ & $0.098$ & $0.077$ \\
$5$ & $462$ & $20$ & $0.028$ & $0.057$ & $0.091$ & $0.063$ \\
$7$ & $330$ & $42$ & $0.035$ & $0.058$ & $0.078$ & $0.043$ \\
$9$ & $55$ & $72$ & $0.043$ & $0.059$ & $0.066$ & $0.024$ \\
$11$ & $1$ & $110$ & $0.058$ & $0.058$ & $0.058$ & $0.000$ \\
\bottomrule
\end{tabular}
\end{table}

\begin{table}[b]
\centering
\caption{\textbf{The threshold-free test: per-problem exact tests against a pooled baseline, with FDR control.} Each arm's endpoint count ($x/128$) is tested against the untrained model's pooled count ($x/1{,}408$, eleven independent evaluations) with a one-sided Fisher exact test at FDR $0.05$ across all $60$ AIME problems. Strata are the pooled base rate. The null holds out one base evaluation and treats it as the ``after''; all eleven replicates return zero detections everywhere. Expansion lives in the two left columns, sharpening in the right one.}
\label{tab:perproblem}
\footnotesize
\setlength{\tabcolsep}{6pt}
\begin{tabular}{lrrrr}
\toprule
\textbf{arm} & \textbf{base $0$ (10)} & \textbf{base $1$--$5$ (12)} & \textbf{base $>5$ (38)} & \textbf{total} \\
\midrule
majority-vote SFT & 0 & 2 & 17 & 19 \\
distillation$^{\dagger}$ & 3 & 7 & 26 & 36 \\
\midrule
\textbf{frozen null}, 11 replicates & $\mathbf{0}$ & $\mathbf{0}$ & $\mathbf{0}$ & $\mathbf{0}$ \\
\bottomrule
\end{tabular}

\vspace{0.3em}
{\small $^{\dagger}$positive control. Seed $0$ of each arm; per-seed counts in Table~\ref{tab:ladder}. Paired exact test over discordant detections, distillation against majority-vote SFT: $19$ to $2$, $p = 2.2\times10^{-4}$. Combining the per-seed comparisons on the low-base pool rather than pooling detections across seeds: $p = 2.0\times10^{-6}$ for distillation over majority-vote SFT, $0.0078$ over STaR and $0.0019$ over the policy-gradient arm. On the ten problems of exactly zero base rate the union over seeds is $5$ against $2$, which is not significant ($p = 0.35$).}
\end{table}

\begin{table}[t]
\centering
\caption{\textbf{The expansion null is not a number, it is a distribution, and at $m{=}2$ it is not zero.} $\er_m$ counts base-unreached problems that reach $m$ correct samples after evolution. The null columns use every ordered pair of independent evaluations of the \emph{untrained} model ($110$ pairs on AIME from eleven such evaluations, $12$ on MATH-500), so every count in them is an artifact; ``one pair'' is the single frozen comparison a conventional study would run, and is what a study with one control reports. At $m{=}2$ the pooled null is $0.058$ and majority-vote self-training's measured $0.048$ sits \emph{below} it. Even $m{=}3$ has a null of $0.023$; no threshold below $m{=}5$ reaches zero.}
\label{tab:expansion}
\scriptsize
\setlength{\tabcolsep}{5pt}
\begin{tabular}{clcccccc}
\toprule
& \multicolumn{3}{c}{\textbf{frozen null, AIME}} & \textbf{null}
& \multicolumn{2}{c}{\textbf{AIME arms}} & \textbf{MATH-500} \\
\cmidrule(lr){2-4}\cmidrule(lr){6-7}
$m$ & \textbf{one pair} & \textbf{range over 110} & \textbf{pooled [95\%]}
& \textbf{MATH-500} & \textbf{maj.-vote} & \textbf{Distillation}$^\dagger$ & \textbf{arms} \\
\midrule
$1$ & 7/25 $= 0.280$ & $0.000$--$0.364$ & $0.176\,[0.145, 0.206]$ & $0.179$
    & 3/21 $= 0.143$ & 12/22 $= 0.545$ & 1/3,\; 0/3 \\
$2$ & 0/25 $= 0.000$ & $0.000$--$0.136$ & $\mathbf{0.058\,[0.038, 0.078]}$ & $0.128$
    & \textbf{1/21} $= 0.048$ & 11/22 $= 0.500$ & 0/3,\; 0/3 \\
$3$ & 0/25 $= 0.000$ & $0.000$--$0.087$ & $0.023\,[0.009, 0.038]$ & $0.000$
    & 0/21 $= 0.000$ & 10/22 $= 0.455$ & 0/3,\; 0/3 \\
$5$ & 0/25 $= 0.000$ & $0.000$--$0.040$ & $0.003\,[0.000, 0.009]$ & $0.000$
    & 0/21 $= 0.000$ & 8/22 $= 0.364$ & 0/3,\; 0/3 \\
\bottomrule
\end{tabular}

\vspace{0.3em}
{\small $^{\dagger}$positive control, not a self-evolution method. Intervals are a cluster bootstrap over base evaluations, since the pairs share evaluations and are not independent. Eleven clusters is few enough that the bootstrap's own coverage is not guaranteed, so Appendix~\ref{app:testrobust} reports the estimate's spread over every subset of the eleven instead, which needs no asymptotics. The band has no base-unreached problems by construction and is omitted. Arm columns use each arm's own baseline, as prior work does; Table~\ref{tab:perproblem} replaces that with a shared one.}
\end{table}

\subsection{Every base-unreached AIME problem, itemized}

\label{app:candidates}

\label{app:shift}

Table~\ref{tab:candidates} lists all $29$ AIME problems that at least one arm found base-unreached at $k{=}128$, with the correct-sample count at that arm's own baseline and at its final checkpoint. It is the whole of the expansion evidence in the paper, at the level of individual problems, and we give it in full because the counts are small enough that a reader should be able to check the conclusion rather than take it.

Three things can be read off it directly. The frozen control's seven apparent expansions are \emph{all} exactly one correct sample in $128$; none of them is two. The distillation arm's expansions are not: its eleven expansions at $m \geq 2$ run $2, 3, 3, 6, 7, 8, 9, 9, 18, 33, 49$, and the two largest (\texttt{2025/21} at $49$ and \texttt{2025/26} at $33$) are the same problems the frozen control ``expanded'' with a single lucky draw. And the majority-vote arm's three at $m{=}1$ reduce to one at $m{=}2$, \texttt{2025/26} at two samples, which is why we report $1/21$ rather than $3/21$.

The three baseline columns differ ($25$, $21$ and $22$ problems at zero) because each arm's checkpoint~$0$ is its own independent $k{=}128$ evaluation of the same untrained model. Thirteen of the $29$ rows are at zero in one arm's baseline and at one, two or three samples in another's. That disagreement is not a bug in the runs; it is the same phenomenon the whole section is about, showing up in the denominator instead of the numerator.

The natural reading of that disagreement, that base-unreached sets must never be pooled across arms, inverts what the observation implies. Each arm's baseline is a noisy draw from the \emph{same} untrained model. Treating the baselines as different populations preserves the noise and makes the arms incomparable; pooling them removes it. Eleven evaluations give $1{,}408$ draws per problem, one denominator for every arm, and a base-rate estimate an order of magnitude tighter than any single arm's. Under the pooled baseline only $10$ problems remain at zero, which is the size of the population $22$ and $25$ over-count. This table supplies the per-arm evidence for that argument; the analysis the paper rests on is the pooled test of Table~\ref{tab:perproblem}, not the per-arm columns here.

\begin{table}[h]
\centering
\caption{\textbf{Every AIME problem the base model does not reach, with its correct-sample count under each arm.} Counts are out of $k{=}128$. A row contributes to $\mathrm{ER}_m$ for an arm when its baseline column is $0$ and its endpoint column is at least $m$. The frozen control is the same untrained model evaluated twice, so its seven $0 \to 1$ rows are the artifact of Section~\ref{sec:f2}; note that every one of them is exactly $1$, and that no arm's genuine expansion is. Bold marks a row that counts as expanded at $m \geq 2$.}
\label{tab:candidates}
\footnotesize
\setlength{\tabcolsep}{3pt}
\begin{tabular}{l rr@{\hskip 11pt} rr@{\hskip 11pt} rr@{\hskip 20pt} l rr@{\hskip 11pt} rr@{\hskip 11pt} rr}
\toprule
& \multicolumn{2}{c}{\textbf{frozen}} & \multicolumn{2}{c}{\textbf{maj.-vote}} & \multicolumn{2}{c}{\textbf{distill.}} & & \multicolumn{2}{c}{\textbf{frozen}} & \multicolumn{2}{c}{\textbf{maj.-vote}} & \multicolumn{2}{c}{\textbf{distill.}} \\
\cmidrule(lr){2-3}\cmidrule(lr){4-5}\cmidrule(lr){6-7}\cmidrule(lr){9-10}\cmidrule(lr){11-12}\cmidrule(lr){13-14}
\textbf{problem} & $t_0$ & $t_1$ & $t_0$ & $t_3$ & $t_0$ & $t_3$ & \textbf{problem} & $t_0$ & $t_1$ & $t_0$ & $t_3$ & $t_0$ & $t_3$ \\
\midrule
2025\,/\,10 & 0 & 0 & 0 & 0 & 0 & 0 & 2026\,/\,10 & 2 & 0 & 0 & 0 & 1 & 1 \\
2025\,/\,12 & 0 & 0 & 0 & 0 & 0 & 0 & 2026\,/\,11 & 0 & 1 & 0 & 0 & 0 & \textbf{9} \\
2025\,/\,13 & 0 & 1 & 2 & 0 & 0 & 1 & 2026\,/\,13 & 0 & 0 & 0 & 0 & 1 & 20 \\
2025\,/\,14 & 0 & 0 & 0 & 0 & 0 & 0 & 2026\,/\,14 & 0 & 1 & 0 & 1 & 0 & \textbf{9} \\
2025\,/\,17 & 0 & 0 & 0 & 0 & 0 & \textbf{6} & 2026\,/\,15 & 0 & 0 & 0 & 0 & 0 & 0 \\
2025\,/\,21 & 0 & 1 & 1 & 0 & 0 & \textbf{49} & 2026\,/\,17 & 0 & 0 & 1 & 0 & 1 & 0 \\
2025\,/\,22 & 1 & 0 & 0 & 0 & 0 & \textbf{3} & 2026\,/\,18 & 0 & 0 & 2 & 3 & 2 & 0 \\
2025\,/\,23 & 0 & 1 & 1 & 0 & 0 & \textbf{8} & 2026\,/\,22 & 1 & 1 & 0 & 0 & 2 & 30 \\
2025\,/\,24 & 0 & 0 & 0 & 0 & 0 & \textbf{18} & 2026\,/\,23 & 1 & 3 & 2 & 2 & 0 & \textbf{7} \\
2025\,/\,25 & 0 & 0 & 0 & 0 & 0 & \textbf{3} & 2026\,/\,27 & 0 & 0 & 0 & 1 & 1 & 6 \\
2025\,/\,26 & 0 & 1 & 0 & \textbf{2} & 0 & \textbf{33} & 2026\,/\,28 & 0 & 0 & 0 & 0 & 0 & 0 \\
2025\,/\,27 & 0 & 0 & 0 & 0 & 0 & 0 & 2026\,/\,29 & 0 & 0 & 0 & 0 & 0 & 0 \\
2025\,/\,29 & 0 & 1 & 0 & 0 & 0 & 0 & 2026\,/\,30 & 0 & 0 & 0 & 0 & 0 & 0 \\
2025\,/\,6 & 0 & 0 & 0 & 0 & 0 & \textbf{2} & 2026\,/\,9 & 0 & 0 & 1 & 0 & 1 & 1 \\
2025\,/\,9 & 0 & 0 & 1 & 0 & 0 & 0 & & & & & & &  \\
\bottomrule
\end{tabular}

\vspace{0.3em}
{\small Base-unreached counts: frozen $25$, majority-vote $21$, distillation $22$; the three differ because each arm's baseline is its own independent $k{=}128$ evaluation, which is the point of Section~\ref{sec:f2}. The majority-vote columns are the unmatched on-set arm of Table~\ref{tab:expansion}, the arm whose $\mathrm{ER}$ that section reports.}
\end{table}

\begin{figure}[h]
\centering
\includegraphics[width=\textwidth]{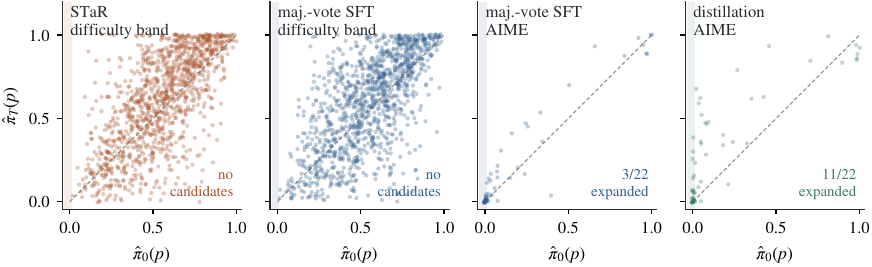}
\caption{\textbf{Per-problem solve rate before and after evolution.} Each point is one evaluation problem; the diagonal is no change. The shaded strip at $\hat\pi_0 = 0$ contains the only problems whose improvement would constitute expansion rather than consolidation, and the counts in it apply the $m \geq 2$ gate against each arm's own baseline. The AIME panels are the matched-ladder arms of Table~\ref{tab:ladder}, not the unmatched pair of Table~\ref{tab:expansion}, which is why the majority-vote count differs between the two. Self-training leaves the strip almost untouched; the distillation control does not.}
\label{fig:shift}
\end{figure}

Figure~\ref{fig:shift} plots every evaluation problem's solve rate before evolution against its solve rate after. It is the consolidation claim without any aggregation: points rise above the diagonal, so the methods do help, but the left edge (problems the base model never reaches, the only ones whose movement would constitute new capability) stays essentially empty for every self-evolution arm and populates only under distillation. The plot also makes visible something the counts obscure: a dense cloud sits below the diagonal in the difficulty-band panels, and those are the corrupted problems.

\subsection{Statistical power and effect-size envelopes}

\label{app:power}

Transition metrics are supported by a much smaller effective sample than accuracy. Accuracy uses every problem; a transition can only be contributed by a problem near the decision boundary, and the fraction of such problems is a property of the benchmark rather than of the method. On the difficulty band we observe $0.20$--$0.22$ transitions per problem, so $1{,}163$ problems yield $233$--$255$ events per arm, against the approximately $249$ per arm required to detect the difference in corrupted share we observe between STaR and majority-vote self-training ($0.417$ against $0.542$ in the pilot) at $80\%$ power and $\alpha = 0.05$. The $112$-problem pilot yields $24$ events.

Three consequences follow. Report counts with intervals rather than ratios, since a ratio of small counts is unstable in a way its point estimate conceals: the $\clr = 1.5$ of Section~\ref{sec:f1} is $9$ over $6$, and the honest version of that headline is the transition \emph{rate}, $7.5\%$ of problems with a Wilson interval of $[4.6\%, 12.0\%]$. State power explicitly whenever a comparison is null. And note that seed variance, not problem sampling, dominates the uncertainty: bootstrap intervals over problems for majority-vote self-training's $\clr$ span $[0.460, 0.785]$, while three training seeds span $[0.55, 1.53]$. Reporting only the former, which is what a single-seed study with bootstrap intervals does, understates the uncertainty by roughly a factor of three.

This is also why we decline to run a second backbone at the scale the budget allows. A difficulty band built for gpt-oss-20b at $300$--$400$ problems would produce $60$--$90$ transition events per arm against the $249$ this subsection requires, so a null from it would be uninformative and a difference from it unreliable. Adding an underpowered arm to a paper whose fourth failure mode is underpowered arms would be the wrong trade, and we prefer to state the limitation.

\paragraph{Power.} A null detection is uninformative without the power that produced it. Table~\ref{tab:power} gives the detection probability for a base-unreached problem as a function of the true post-training rate and the number of pooled baselines. Pooling is what buys sensitivity: a single baseline needs seven correct samples in $128$ before a problem can be detected at all, eleven need two. At eleven the test has $80\%$ power against a true rate of $0.024$, and the largest endpoint any self-training arm produced on a base-$0$ problem is $1/128$. The claim the data support is a bound.

\begin{table}[h]
\centering
\caption{\textbf{Power of the per-problem test on a base-unreached problem.} Probability of detection against the true post-training pass rate, at $k{=}128$, for a problem with no successes in the pooled baseline, under the cutoff the procedure attains. Pooling is what buys the sensitivity: one baseline requires seven correct samples before a problem can be detected, eleven require two. The null result of Section~\ref{sec:results} is therefore a bound, no self-training arm having lifted a base-$0$ problem above a $2.4\%$ pass rate, and not an absence of measurement.}
\label{tab:power}
\footnotesize
\setlength{\tabcolsep}{5pt}
\begin{tabular}{lcccccccc}
\toprule
\textbf{baselines pooled} & \textbf{draws} & \textbf{min.\ successes} & $0.01$ & $0.02$ & $0.03$ & $0.05$ & $0.08$ & $0.12$ \\
\midrule
$1$ & $128$ & $7/128$ & $0.00$ & $0.01$ & $0.09$ & $0.46$ & $0.89$ & $1.00$ \\
$3$ & $384$ & $4/128$ & $0.04$ & $0.25$ & $0.54$ & $0.89$ & $0.99$ & $1.00$ \\
$5$ & $640$ & $3/128$ & $0.14$ & $0.47$ & $0.74$ & $0.96$ & $1.00$ & $1.00$ \\
$\mathbf{11}$ & $1{,}408$ & $2/128$ & $0.37$ & $0.73$ & $0.90$ & $0.99$ & $1.00$ & $1.00$ \\
\bottomrule
\end{tabular}

\vspace{0.3em}
{\small Smallest true rate detectable at $80\%$ power with eleven baselines: $0.024$ under the attained cutoff, $0.034$ under Bonferroni. Observed endpoints on the ten base-$0$ problems: distillation $0, 0, 0, 0, 0, 0, 0, 2, 6, 18$; majority-vote SFT $0, 0, 0, 0, 0, 0, 0, 0, 1, 1$, all of $128$.}
\end{table}

\begin{figure}[h]
\centering
\includegraphics[width=\textwidth]{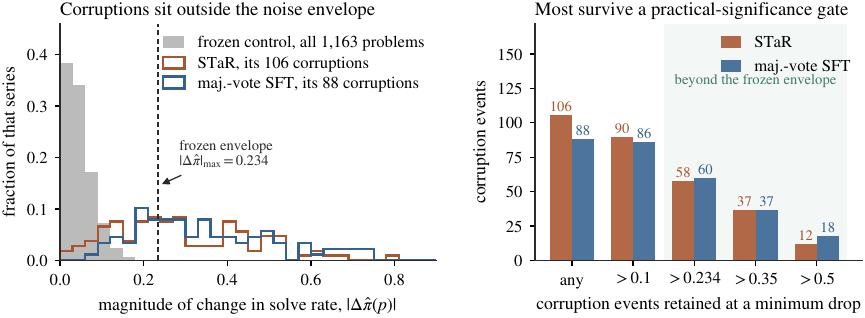}
\caption{\textbf{Corruption is large movement, not threshold flicker.} \emph{Left:} distribution of $|\Delta\hat\pi(p)|$ for the frozen control over all $1{,}163$ problems and for each arm's corruption events, each normalized to its own total; over the twenty independent frozen comparisons the records contain, no problem moves by more than $0.234$. \emph{Right:} corruption counts surviving a minimum-drop gate. Over half of each arm's events lie beyond the frozen envelope entirely.}
\label{fig:effects}
\end{figure}

\paragraph{Effect sizes, against the envelope a frozen model produces.} Table~\ref{tab:effects} and Figure~\ref{fig:effects} give the size of every transition event on the difficulty band against the largest solve-rate movement the frozen control produced over its twenty independent comparisons. That envelope is what separates a threshold flicker from a capability change, and it is reported alongside the counts because a count alone cannot make the distinction.

\begin{table}[t]
\centering
\caption{\textbf{Effect sizes for transition events on the difficulty band.} Counts of events whose solve-rate change exceeds each magnitude, against the envelope of the frozen control. The control row is the worst value each column takes over the twenty ordered pairs of independent evaluations of the untrained model, rather than a single pair: across all twenty, one problem moved past $0.2$ and none past $0.234$, so $0.234$ is the measured zero-false-positive gate and the bolded column is what survives it.}
\label{tab:effects}
\footnotesize
\setlength{\tabcolsep}{5pt}
\begin{tabular}{llrrrrrrr}
\toprule
\textbf{arm} & \textbf{event} & \textbf{total} & $>0.1$ & $>0.2$ & $>0.234$ & $>0.35$ & \textbf{median} & \textbf{max} \\
\midrule
\multirow{2}{*}{STaR} & corrupted & 106 & 90 & 69 & \textbf{58} & 37 & $0.258$ & $0.781$ \\
                      & learned   & 149 & 133 & 107 & --- & 61 & $0.297$ & $0.812$ \\
\midrule
\multirow{2}{*}{maj.-vote} & corrupted &  88 & 86 & 69 & \textbf{60} & 37 & $0.320$ & $0.750$ \\
                      & learned   & 145 & 133 & 104 & --- & 62 & $0.297$ & $0.773$ \\
\midrule
frozen, worst of $20$ pairs & any change & 1{,}163 & 108 & 1 & \textbf{0} & 0 & $0.039$ & $\mathbf{0.234}$ \\
\bottomrule
\end{tabular}
\end{table}

\subsection{Out-of-domain probes}

\label{app:collateral}

\paragraph{Both probes, three times each.} Each out-of-domain probe was run three times against the same stored checkpoints, so any difference between replicates is measurement rather than training. One run of the refusal probe is capable of producing a publishable safety finding; two further runs dissolve it, and what the data support is a bound rather than a null result: \emph{any refusal change here is smaller than about eleven points}. MMLU-Redux is better behaved, but its effect is smaller than a single replicate suggests, the $+10.6$ points of that replicate being the largest of six measurements whose median is nearer $+4$.

\paragraph{F6. An underpowered probe produces a phantom safety result.} A $10$-prompt refusal probe shows the refusal rate falling $10$ points across checkpoints, corroborating existing reports of safety degradation under self-training \citep{shao2026your,qi2024finetuning}. At $50$ prompts the effect disappears and the $10$-prompt trajectory is non-monotonic, the signature of noise. The probe is underpowered twice over and clustered besides; Section~\ref{sec:results} reports what survives at the larger size.

Table~\ref{tab:collateral} reports both probes in full. MMLU-Redux \citep{gema2025done} is evaluated on $200$ questions per checkpoint. The refusal probe comprises $50$ requests spanning fraud, violence, harassment, child safety, cyber-intrusion, hate, self-harm, regulated goods and misinformation, each sampled four times per checkpoint for $200$ observations. Prompts are deliberately mild and non-operational: they exist to trigger a refusal, not to elicit anything usable. A response counts as a refusal if a refusal marker appears in its first $400$ characters, a threshold chosen so that a model which complies and then appends a disclaimer is not scored as having refused. This marker rule is a limitation and the negative result should be read as scoped to it: a model that shifts from explicit refusal to a soft deflection without a marker phrase would be scored as having complied, and we did not audit for that.

The nine categories are represented by five or six prompts each and were written to span the standard refusal taxonomy without approaching an operational threshold in any of them: requests for the \emph{shape} of a harmful action rather than the means to perform it. We release the probe with the code, together with one record per generation, so that the result can be re-scored under a different rule. Replicate~1 was logged at per-checkpoint aggregate resolution, whereas replicates~2 and~3 record full per-generation outputs; the cluster-robust analysis below is therefore computed on the latter two.

\paragraph{A naive analysis of this probe fails twice over, and the second failure is the consequential one.} Four samples of one prompt are not four independent observations, so a $z$-test on $200$ Bernoulli trials is the wrong instrument; aggregating to $50$ prompt-level proportions gives a measured design effect of $1.8$--$2.8\times$, and a cluster-robust standard error of $0.044$--$0.057$ rather than the $0.032$--$0.035$ a naive analysis assumes. That alone would only widen an interval. What it reveals is worse: at $50$ prompts the smallest change this probe can detect at $80\%$ power is about $0.11$, and the change under discussion is about $0.10$. The probe is being asked a question it cannot answer.

Replication settles the point empirically. Three executions of the identical probe against the identical stored checkpoints give, for majority-vote self-training, refusal changes of $-0.015$ (not significant), $-0.115$ ($t = -2.79$ clustered, significant) and $-0.005$. Nothing about the model differs across them; only the sampling does, and the one significant reading is replicated by neither neighbor. We therefore decline to claim that self-training causes no safety erosion here, and equally decline to claim the opposite: what the data support is a bound, that any change is smaller than roughly eleven points, which is compatible both with nothing happening and with an effect of practical size. Detecting the latter would need a probe several times larger, and prompts, not samples, are the unit that has to grow.

MMLU-Redux is only slightly better placed. STaR gains in all three replicates ($+6.3$, $+7.0$ and $+2.0$ points; paired McNemar on the second, $25$ gained against $11$ lost, $p = 0.029$), which is the one out-of-domain result here that survives replication and a paired test. Majority-vote self-training gives $+10.6$, then $+2.0$ ($p = 0.58$) and then $+1.5$ ($p = 0.72$), and the untrained baseline itself moves between $0.550$ and $0.645$ across replicates. A single greedy draw on $200$ questions is not a stable anchor for a drift claim.

\begin{table}[h]
\centering
\caption{\textbf{Out-of-domain probes, run three times each against the same stored checkpoints.} Any difference between replicates is measurement, not training. Refusal changes are prompt-level means with cluster-robust intervals; MMLU is tested with paired McNemar on the same questions. The one significant refusal result does not replicate on either side of it, and the largest MMLU gain is the largest of six measurements of a smaller effect.}
\label{tab:collateral}
\scriptsize
\setlength{\tabcolsep}{4pt}
\begin{tabular}{llccc}
\toprule
\textbf{probe} & \textbf{arm} & \textbf{replicate 1}$^{\ast}$ & \textbf{replicate 2} & \textbf{replicate 3} \\
\midrule
\multirow{2}{*}{MMLU-Redux ($n{=}200$)}
  & STaR      & $+0.063$ & $+0.070$ ($p = 0.029$) & $+0.020$ ($p = 0.60$) \\
  & maj.-vote & $+0.106$ & $+0.020$ ($p = 0.58$)  & $+0.015$ ($p = 0.72$) \\
\midrule
\multirow{2}{*}{refusal ($n{=}50$ prompts)}
  & STaR      & $-0.050$ & $-0.010\,[-0.085, 0.065]$ & $-0.005\,[-0.068, 0.058]$ \\
  & maj.-vote & $-0.015$ & $\mathbf{-0.115}\,[-0.196, -0.034]$
              & $-0.005\,[-0.080, 0.070]$ \\
\midrule
refusal, underpowered ($n{=}10$) & STaR & $-0.100$, non-monotonic & --- & the F6 probe \\
\bottomrule
\end{tabular}

\vspace{0.3em}
{\small $^{\ast}$replicate 1 is recorded at per-checkpoint aggregate resolution rather than per generation, so it carries no interval. \quad Design effect $1.8$--$2.8\times$; cluster-robust SE $0.044$--$0.057$ against a naive $0.032$--$0.035$. Minimum detectable refusal change at $50$ prompts and $80\%$ power: $0.09$--$0.12$. Untrained MMLU baseline across replicates: $0.581$--$0.610$ (STaR arm), $0.550$--$0.645$ (majority-vote arm).}
\end{table}

\section{Extended related work, and the scope of the claims}

\label{app:extended}

\subsection{Related work, continued}

\label{app:related}

\paragraph{The RLVR side of the debate.} The consolidation question is posed most sharply in reinforcement learning from verifiable rewards, and the analyses there bear directly on ours: how little of the model RLVR actually moves \citep{mukherjee2025reinforcement}, how easily entropy collapses \citep{cui2025entropy}, and how weak the training signal can be while still producing gains \citep{shao2026spurious,wang2025reinforcement}. The underlying algorithms are those of \citet{shao2024deepseekmath,deepseek2025r1,yu2025dapo,liu2025understanding}. Our policy-gradient arm sits in this family, which is why we ran it: an audit that positions itself against an RLVR literature and then measures only supervised fine-tuning is auditing something else.

\paragraph{Reconciling with prior reports.} \citet{lin2026detecting} report that TTRL corrupts more than it learns; our seed-$0$ run finds the opposite and our other two seeds agree with them. We read this as one axis rather than a contradiction: corruption requires both unreliable supervision \emph{and} prior competence on the affected problems, and that regime is narrow. Their backbone is weaker and its majority vote correspondingly less accurate (ours is $91$--$95\%$ correct on the band). That places them inside the regime while ours sits at its edge, which is why our result moves with the seed. The account is testable, predicts that the corrupting regime is a moderately competent model rather than a weak or a strong one, and agrees with a sharpening view \citep{huang2025self}: when verification is nearly perfect, sharpening is nearly harmless.

\paragraph{Parametric self-evolution.} STaR \citep{zelikman2022star} established the loop most later methods vary: sample rationales, keep those reaching the correct answer, fine-tune, repeat \citep{yuan2023scaling,singh2024beyond,wang2023selfinstruct,huang2023large}. TTRL \citep{zuo2025ttrl} discards labels entirely, treating a majority vote over the model's own samples as a reward, which is self-consistency \citep{wang2023selfconsistency} used as a training rather than an inference-time signal, and label-free variants have followed \citep{zhou2026evolving,moradi2025continuous,acikgoz2025selfimproving}. Others replace the filter with a learned verifier \citep{hosseini2024vstar,chen2026enhancing}, the model's own reward judgments \citep{yuan2024selfrewarding}, self-play \citep{chen2024self,zhao2025absolutezero}, debate \citep{srivastava2025debate}, generated curricula \citep{huang2026rzero}, language feedback \citep{lu2024self}, or latent rationales \citep{zelikman2024quietstar}; SEAL \citep{zweiger2025self} learns a policy over weight edits, and surveys catalog the space \citep{tao2024survey,gao2026survey}. Inference-time self-correction is a distinct question and does not reliably work \citep{madaan2023selfrefine,huang2024large}.

\paragraph{Degradation under self-training.} \citet{lin2026detecting} report that for TTRL the problems corrupted outnumber those learned and that recovery is rare once a majority vote settles on a wrong answer; \citet{yu2026self} frame this as capability erosion, and \citet{zhao2026large} show that self-evolving agents often fail to use accumulated experience causally. The backdrop is catastrophic forgetting \citep{kirkpatrick2017overcoming,luo2025empirical}, which LoRA mitigates but does not remove \citep{hu2022lora,biderman2024lora}; model collapse under recursive training on generated data \citep{shumailov2024collapse,alemohammad2024selfconsuming,gerstgrasser2024is}; and safety erosion under fine-tuning \citep{qi2024finetuning,shao2026your}, which motivates our out-of-domain probes.

\paragraph{Evaluation methodology, test-time adaptation and distillation.} pass@$k$ estimation \citep{chen2021evaluating} underpins our per-problem solve rates; \citet{miller2024adding} argue for reporting evaluation uncertainty at all and \citet{hariri2026dont} for a Bayesian posterior in place of the pass@$k$ point estimate, though neither addresses transitions or supplies a null. Seed sensitivity in fine-tuning is long documented \citep{dodge2020finetuning} and reappears as our most expensive failure mode. Contamination \citep{zhang2024careful,xu2024benchmark,xu2025dcr,sun2025the} motivates our use of AIME alongside the contaminated but comparable MATH-500 \citep{lightman2024verify,hendrycks2021measuring,cobbe2021training}; probes use MMLU-Redux \citep{gema2025done}; intervals are Wilson \citep{wilson1927probable,brown2001interval} and percentile bootstrap \citep{efron1979bootstrap}, count comparisons Fisher's exact test \citep{fisher1922on}. Majority-vote self-training sits within a literature on adapting parameters at inference \citep{sun2025learning,moradi2025continuous,gozeten2025testtime}, adjacent to test-time \emph{compute} scaling that changes no parameters \citep{snell2024scaling,muennighoff2025s1}; our positive control is ordinary knowledge distillation \citep{hinton2015distilling,hsieh2023distilling}.

\subsection{Concurrent work, in full}

\label{app:concurrent}

\paragraph{Concurrent work, in full.} Two contemporaneous papers overlap with ours. \citet{strozzi2026teacherfree} ask our headline question, whether teacher-free self-training acquires capability or only expresses existing capability, and report a pass@$K$ crossover in which the base model wins at $K{=}64$ on a DSL domain with a free verifier. We regard our consolidation result as a replication in a different domain and supervision regime, with a positive control they do not run. \citet{yuan2026diversity} track per-problem pass@$256$ boundary entry and exit across training, structurally our ledger, and their Appendix~C gives an \emph{analytic} sampling-noise null: a binomial calculation under an unchanged policy. Ours is \emph{measured}, by pushing an unchanged model through the identical pipeline, and we can say precisely how much that buys: less than the phrase ``measure your null'' suggests. For the solve-rate estimator we recommend, a binomial simulation at each problem's pooled base rate reproduces our measured floor closely (median $3/4$ against $4/3$ transitions, envelope $0.203$ against $0.234$), so on that protocol the analytic null is adequate. The two diverge exactly where the noise is not binomial. That is the single-decode protocol, where an analytic null predicts nothing and the measurement finds $18.9\%$ of problems changing state (Section~\ref{sec:f1}), and it is also where no analytic null has been worked out, which is how we found that the expansion statistic's null is non-zero at any threshold below $m{=}5$. Neither paper reports a null for it.

\subsection{Limitations, in full}

\label{app:limitations}

\paragraph{Limitations, in full.} Three evolution rounds and roughly $270$ optimizer steps, far short of the schedules of the methods we position against, so we cannot speak to accumulation. Expansion rests on $22$ low-base problems on a single benchmark, $10$ of them in the uncontaminated $2026$ half: enough for the dissociation to be significant, not to be precise. Corruption rests on the difficulty band, drawn from MATH \emph{training} problems that are held out from our stream but almost certainly not from the backbone's pretraining data, so it may in part be the perturbation of memorized answers rather than of reasoning; the AIME arms that avoid this have no corruption power. The additional seeds for the band arms were run on a $175$-problem subsample rather than the full $1{,}163$, and every seed is scored on those $175$ problems. One backbone family carries every substantive claim. A second (gpt-oss-20b) was characterized and its floor measured cleanly, but a band built for it at an affordable scale would have yielded roughly a third of the transition events our own power analysis (F4) says such a comparison needs. We therefore state cross-backbone generalization as untested rather than answer it with an underpowered arm. The distillation control is matched in stream, retained volume and evaluation, but teacher capability and verbosity necessarily still differ, and F3 shows the latter is real. Finally, the evolution stream is far easier than AIME (STaR retains about $90\%$ of it under ground-truth filtering), so a method that trains only on what it already solves is a priori unlikely to expand into AIME-hard territory. Methods that generate their own curricula exist precisely to escape that, and our null does not speak to them.

\section{Case studies: the failure modes in the model's own words}
\label{app:cases}

Every count in this paper is an aggregate over generations that a reader never sees. This appendix opens five of them. Each transcript is the model's own output, taken from the released records at the character offsets stated, and its title bar carries the provenance needed to find it again: which arm, which checkpoint, how long the completion ran, and how the grader scored it. Frame color repeats the verdict: \textcolor{caseOK}{green} for graded correct, \textcolor{caseBad}{red} for graded incorrect, \textcolor{caseNull}{gray} where no verdict applies.

Two mechanical changes are applied. Non-ASCII decoration, and the models are fond of emoji section markers, is transliterated so that pdf\LaTeX{} can set it, and blank lines are dropped. Nothing inside a block is elided and no line is rewrapped by us; where a line runs past the margin the typesetter breaks it and marks the continuation with \textcolor{caseNull}{$\hookrightarrow$}. The prompt is identical in every case and is given in Appendix~\ref{app:protocol}.

\subsection{Case 1: one model, one prompt, temperature zero, two different answers}

This is F1 with nothing else going on. The model is the untrained backbone. The prompt is byte-identical. Decoding is greedy at $T = 0$ with a fixed seed. The two completions below are the two passes of the frozen-model control, and the single-decode ledger scores the pair as one \emph{corrupted} problem: a capability lost between two evaluations of a model that never trained.

\begin{problemcard}{math500/test/geometry/226}{$180$}
In circle $J$, $HO$ and $HN$ are tangent to the circle at $O$ and $N$. Find the number of degrees in the sum of $m\angle J$ and $m\angle H$.

\smallskip\noindent{\footnotesize An Asymptote figure accompanies the problem and is passed to the model verbatim, as it appears in MATH-500.}
\end{problemcard}

The two completions are identical for their first $1523$ characters, through the setup and the first two solution steps, and then diverge mid-sentence:

\begin{outputnull}{frozen control \ \textbullet\ \ checkpoint~0 and 1 \ \textbullet\ \ identical prefix, last 240 characters}
---
### Step 3: Use triangle angle sum
In triangle $ JHO $:
$$
\angle J + \angle JHO + \angle HOJ = 180^\circ
$$
We know:
- $ \angle JHO = 90^\circ $
- $ \angle HOJ = \angle JON $, which is the angle between two radii
\end{outputnull}

\begin{outputok}{frozen control \ \textbullet\ \ checkpoint~0 \ \textbullet\ \ 840 tokens \ \textbullet\ \ graded correct}
, so it's the central angle subtended by arc $ ON $.
But we are not given any specific arc measures, so we can't compute that directly.
However, we are only asked for the **sum** of $ \angle J $ and $ \angle H $, not the individual measures.
Let's denote:
- $ \angle J = x $
- $ \angle H = y $
\end{outputok}

\begin{outputbad}{frozen control \ \textbullet\ \ checkpoint~1 \ \textbullet\ \ 1{,}595 tokens \ \textbullet\ \ graded incorrect}
 to the points of tangency.
So:
$$
\angle J + 90^\circ + \angle HOJ = 180^\circ
\Rightarrow \angle J + \angle HOJ = 90^\circ \tag{1}
$$
Similarly, in triangle $ JHN $:
$$
\angle J + \angle JHN + \angle HNJ = 180^\circ
\Rightarrow \angle J + \angle HNJ = 90^\circ \tag{2}
$$
\end{outputbad}

The paths do not reconverge:

\begin{outputok}{frozen control \ \textbullet\ \ checkpoint~0 \ \textbullet\ \ 840 tokens \ \textbullet\ \ final answer}
$$
Substitute known values:
$$
x + y + 90^\circ + 90^\circ = 360^\circ
$$
$$
x + y + 180^\circ = 360^\circ
$$
$$
x + y = 180^\circ
$$
---
### [ok] Final Answer:
$$
\boxed{180}
$$
So, the sum of the measures of $ \angle J $ and $ \angle H $ is $ \boxed{180} $ degrees.
\end{outputok}

\begin{outputbad}{frozen control \ \textbullet\ \ checkpoint~1 \ \textbullet\ \ 1{,}595 tokens \ \textbullet\ \ final answer}
$$
---
### Step 6: Final calculation
We now know:
- $ \angle J = 60^\circ $
- $ \angle H = 180^\circ - 2\angle J = 180^\circ - 120^\circ = 60^\circ $
So:
$$
m\angle J + m\angle H = 60^\circ + 60^\circ = \boxed{120^\circ}
$$
---
### [ok] Final Answer:
$$
\boxed{120}
$$
\end{outputbad}

Pass~1 reaches the correct $180^\circ$ by the quadrilateral angle sum. Pass~2 commits to a chain of triangle relations, invents an isosceles constraint the figure does not support, and answers $120$. Neither generation is pathological; the second is longer and, read on its own, no less plausible.

What produced the divergence is not sampling, of which there is none at $T = 0$, but floating-point non-associativity under a batching regime the caller does not control \citep{he2025nondeterminism}. Two forward passes over the same tokens with different batch composition reduce in different orders, perturb a logit in the last decimal place, and flip an arg-max on a near-tie. Everything after that token is a different generation.

This is the whole mechanism behind $\textsc{clr} = 1.5$ on an untrained model (Section~\ref{sec:f1}). On MATH-500 it happens to $15$ of $200$ problems between two passes; on the difficulty band, to $220$ of $1{,}163$. An audit whose state variable is a single greedy decode is measuring this.

\subsection{Case 2: the same problem, ``expanded'' once by a frozen model and once by a teacher}

This case carries F2. The problem is one the base model does not reach: zero correct in $128$ samples at the frozen control's first pass, and zero again in the distillation arm's own baseline. Under the $m = 1$ criterion it is then scored as \emph{expanded} twice: once by the frozen model, which did not train and produced a single correct sample on its second pass, and once by the distilled student, which did train.

\begin{problemcard}{aime2025/21}{$237$}
Let $A$ be the set of positive integer divisors of $2025$. Let $B$ be a randomly selected subset of $A$. The probability that $B$ is a nonempty set with the property that the least common multiple of its elements is $2025$ is $\frac{m}{n}$, where $m$ and $n$ are relatively prime positive integers. Find $m+n$.
\end{problemcard}

The base model's failure is systematic rather than random. It computes the numerator correctly and then divides by $2^{15} - 1$ instead of $2^{15}$:

\begin{outputbad}{untrained backbone \ \textbullet\ \ checkpoint~0 \ \textbullet\ \ 2{,}638 tokens \ \textbullet\ \ representative of $128$ samples \ \textbullet\ \ graded incorrect}
$$
$$
\text{GCD}(425, 109) = \text{GCD}(109, 425 \mod 109) = \text{GCD}(109, 107)
$$
$$
\text{GCD}(109, 107) = \text{GCD}(107, 2) = \text{GCD}(2, 1) = 1
$$
So the GCD is 1. Therefore, the fraction is already in lowest terms.
---
### **Final Answer**
The probability is:
$$
\frac{30240}{(@\hlk{32767}@)}
$$
So $ m = 30240 $, $ n = (@\hlk{32767}@) $
$$
m + n = \boxed{63007}
$$
\end{outputbad}

One sample in $128$ from the frozen model's second pass does not make the slip:

\begin{outputok}{frozen control \ \textbullet\ \ checkpoint~1 \ \textbullet\ \ 2{,}027 tokens \ \textbullet\ \ the only one of $128$ \ \textbullet\ \ graded correct}
- $ (@\hlk{32768}@) = 2^{15} $
So:
$$
\gcd(27904, (@\hlk{32768}@)) = 2^8 = 256
$$
So:
$$
\frac{27904}{(@\hlk{32768}@)} = \frac{109}{128}
$$
Check that $ \gcd(109, 128) = 1 $ (since 109 is prime)
---
### Step 9: Final Answer
We are given that the probability is $ \frac{m}{n} $, where $ m = 109 $, $ n = 128 $
So:
$$
m + n = 109 + 128 = \boxed{237}
$$
---
### [ok] Final Answer:
$$
\boxed{237}
$$
\end{outputok}

And the distilled student, which reaches this problem in $49$ of $128$ samples:

\begin{outputok}{distillation \ \textbullet\ \ checkpoint~3 \ \textbullet\ \ 2{,}853 tokens \ \textbullet\ \ one of $49$ of $128$ \ \textbullet\ \ graded correct}
---
### Probability
\[
P = \frac{N}{2^{15}} = \frac{27904}{(@\hlk{32768}@)}.
\]
The greatest common divisor of numerator and denominator is \(256\):
\[
\frac{27904}{(@\hlk{32768}@)}= \frac{27904/256}{(@\hlk{32768}@)/256}= \frac{109}{128}.
\]
Hence \(m=109\) and \(n=128\), with \(\gcd(m,n)=1\).
Finally,
\[
m+n = 109+128 = 237.
\]
\[
\boxed{237}
\]
\end{outputok}

Read the three together. The difference between the frozen model's failure and its one success is a single character: $32767$ against $32768$. Nothing was learned between the two passes, because nothing happened between the two passes. The lucky sample is the tail of a Bernoulli process with a small success probability, and observing it once in $128$ draws bounds that probability at $q \leq 0.023$ rather than establishing it was zero before.

Two conclusions follow, and they pull in opposite directions, which is why we state both. First, the $m = 1$ criterion cannot distinguish the second block from the third: on the counting rule used in prior work, the frozen model and the distilled student both ``expanded'' this problem. Requiring $m \geq 2$ separates them, because the frozen model produces exactly one success here and the student produces $49$. Second, and this bounds the headline claim, $\hat\pi_0(p) = 0$ at $k = 128$ does not mean the base model could never solve this problem. It means the base model has a systematic arithmetic habit that costs it this problem almost every time. We therefore write ``base-unreached at $k$'' throughout rather than ``unsolvable'', and the claim the paper defends is about what a fixed sampling budget reaches, not about the support of the base distribution.

Appendix~\ref{app:candidates} lists all $29$ AIME problems that at least one arm found base-unreached, with their counts under each arm, so that this reasoning can be checked against every case rather than the one displayed here.

\subsection{Case 3: a completion that ran out of budget, scored as a lost capability}

F3. The generation below is from the distillation arm at checkpoint~$1$ on the difficulty band, one of the $65.3\%$ of completions at that checkpoint that hit the $2{,}048$-token cap.

\begin{problemcard}{mathtrain/test/algebra/1017}{4,\,6,\,14,\,15}
Four distinct integers $a$, $b$, $c$ and $d$ have the property that when added in pairs, the sums $10$, $18$, $19$, $20$, $21$ and $29$ are obtained. What are the four integers in increasing order?
\end{problemcard}

\begin{outputbad}{distillation \ \textbullet\ \ checkpoint~1 \ \textbullet\ \ 2{,}048 tokens \ \textbullet\ \ cap reached \ \textbullet\ \ no answer extracted}
z+w = 22 => (19 - x) + (21 - y) = 22 => 40 - (x+y) = 22 => x+y = 18. Again contradicts 10.
Maybe x+z = 18, y+w = 20? Then w = 20 - y.
z+w = 22 => (18 - x) + (20 - y) = 22 => 38 - (x+y) = 22 => x+y = 16. Still not 10.
Maybe x+z = 20, y+w = 21? Then z = 20 - x, w = 21 - y.
z+w = 22 => (20 - x) + (21 - y) = 22 => 41 - (x+y) = 22 => x+y = 19. Not 10.
Maybe x+z = 19, y+w = 20? Then w = 20 - y, z = 19 - x.
z+w = 22 => (19 - x) + (20 - y) =
\end{outputbad}

The completion ends mid-expression. There is no \verb|\boxed{}|, no stated answer and no final sentence, so the extractor returns nothing and the grader scores the sample incorrect: correct behavior for a grader, and the wrong summary of what happened. The model has not lost the ability to solve this problem. It has adopted its teacher's habit of enumerating cases in prose and has run out of room.

Aggregated over the arm, this scores a method that improves from $0.656$ to $0.845$ on completions that finish as the most destructive arm in the study, at $\textsc{clr} = 1.10$ and $177$ corruptions. The diagnostic that catches it is one number per checkpoint, the truncation rate, and it costs nothing to log. Without it, nothing in the ledger distinguishes this case from Case~4.

\subsection{Case 4: a capability majority-vote self-training destroyed}

By contrast, a real corruption, on an arm whose truncation rate never moves ($16.6\%$ to $16.0\%$). The base model solves this problem in $104$ of $128$ samples; after three rounds of majority-vote self-training, in $8$.

\begin{problemcard}{mathtrain/train/number\_theory/1075}{$2$}
Convert $199_{10}$ to base $2$. Let $x$ be the number of zeros and $y$ be the number of ones in base $2$. What is the value of $y-x$?
\end{problemcard}

\begin{outputok}{untrained backbone \ \textbullet\ \ checkpoint~0 \ \textbullet\ \ 438 tokens \ \textbullet\ \ $104/128$ correct \ \textbullet\ \ graded correct}
$$
199_{10} = 11000111_2
$$
---
### **Step 2: Count the number of ones (y) and zeros (x)**
Binary representation:
$$
11000111_2
$$
Let's count:
- Number of **ones (y)**: There are (@\hlk{**5 ones**}@)
- Number of **zeros (x)**: There are **3 zeros**
---
### **Step 3: Compute $ y - x $**
$$
y - x = 5 - 3 = \boxed{2}
$$
---
### [ok] Final Answer:
$$
\boxed{2}
$$
\end{outputok}

\begin{outputbad}{after majority-vote SFT \ \textbullet\ \ checkpoint~3 \ \textbullet\ \ 414 tokens \ \textbullet\ \ $8/128$ correct \ \textbullet\ \ graded incorrect}
$$
199_{10} = 11000111_2
$$
---
### **Step 2: Count the number of zeros and ones**
Binary number: $11000111$
Let's count:
- Number of **ones** ($y$): There are (@\hlk{**6** ones}@).
- Number of **zeros** ($x$): There are **3** zeros.
---
### **Step 3: Compute $y - x$**
$$
y - x = 6 - 3 = \boxed{3}
$$
---
### [ok] Final Answer:
$$
\boxed{3}
$$
So, the value of $y - x$ is $\boxed{3}$.
\end{outputbad}

Both generations derive the same binary expansion, $199_{10} = 11000111_2$, and both then count its digits. The base model counts five ones and three zeros and answers $2$. The evolved model counts \emph{six} ones from the identical string and answers $3$. The reasoning is otherwise unchanged, down to the section headings; what moved is a digit count at the last step.

This is what corruption looks like in this regime, and the mechanism needs stating precisely, because it is not the one the phrase ``capability erosion'' suggests. Nothing has been forgotten. A single sub-step has become unreliable in a way that is invisible in the surrounding text and produces a confidently formatted wrong answer. It is also, by construction, invisible to a majority vote over the model's own samples: at checkpoint~$3$ the wrong count is the modal one, so the pseudo-label for this problem would now be $3$.

The solve rate moved by $0.75$, more than three times the largest change the frozen control produced anywhere in the $1{,}163$-problem set (Table~\ref{tab:effects}). This event is not a threshold flicker.

\subsection{Case 5: a capability STaR sharpened, and what sharpening means concretely}

The complement of Case~4, and the modal outcome across the study. The base model reaches this problem in $24$ of $128$ samples; after three rounds of STaR, in $128$ of $128$. No new capability is involved: the correct answer was already in the base model's output distribution and training moved mass onto it.

\begin{problemcard}{mathtrain/test/number\_theory/254}{$6$}
Sally, Wei-Hwa, and Zoe are playing a game of marbles involving first arranging as many piles of $10$ marbles as possible. Sally brought $239$ marbles, Wei-Hwa brought $174$ marbles, and Zoe brought $83$ marbles. If all their marbles are grouped together, how many must be removed in order to start the game?
\end{problemcard}

\begin{outputbad}{untrained backbone \ \textbullet\ \ checkpoint~0 \ \textbullet\ \ 261 tokens \ \textbullet\ \ one of the $104$ that miss, of $128$ \ \textbullet\ \ graded incorrect}
$$
(@\hlk{500}@) \div 10 = 50
$$
This means **50 piles of 10 marbles** can be formed with **exactly (@\hlk{500}@) marbles**.
---
### Step 3: Determine how many marbles must be removed
Since the total is already a multiple of 10, **no marbles need to be removed** to form piles of 10 marbles.
---
### [ok] Final Answer:
$$
\boxed{0}
$$
\end{outputbad}

\begin{outputok}{after STaR \ \textbullet\ \ checkpoint~3 \ \textbullet\ \ 318 tokens \ \textbullet\ \ all $128$ correct \ \textbullet\ \ graded correct}
We want to know how many marbles must be **removed** to make the total a multiple of 10. That is, we find the **remainder** when (@\hlk{496}@) is divided by 10.
$$
(@\hlk{496}@) \div 10 = 49 \text{ remainder } 6
$$
So, **6 marbles** must be removed to make the total number of marbles divisible by 10.
---
### [ok] Final Answer:
$$
\boxed{6}
$$
**6 marbles must be removed** to start the game.
\end{outputok}

The base model's error is upstream of the reasoning: it adds $239 + 174 + 83$ and gets $500$, then correctly observes that $500$ is already a multiple of ten and answers $0$. The trained model adds correctly to $496$, takes the remainder and answers $6$. In $24$ of its $128$ samples the base model also added correctly and also answered $6$; those samples are why this problem counts as \emph{sharpened} rather than \emph{expanded}.

This is the shape of almost everything the audited methods do. Across all seven arms, every problem newly solved at pass@1 was one the base model had already solved at least once in $128$ samples: $2/2$, $3/3$, $4/4$, $9/9$, $149/149$, $145/145$, $161/161$. It is a real and useful effect: a model that answers correctly on the first attempt rather than on the fifth is more useful, and consolidation is what these methods reliably deliver. It is a different claim from the one a leaderboard number invites, and the distinction is measurable at modest cost.

\end{document}